\documentclass[]{fairmeta}

\usepackage{amsmath,amsfonts,bm}
\usepackage{xspace}
\usepackage{hyperref}
\usepackage{url}
\usepackage{subcaption}

\usepackage{colortbl}
\usepackage{xcolor}

\usepackage{graphicx}
\usepackage{algorithmic}
\usepackage{algorithm}
\usepackage{float}

\usepackage{booktabs}

\usepackage{enumitem}

\usepackage{amsthm}

\theoremstyle{plain}

\theoremstyle{definition}

\theoremstyle{remark}

\def\eqref#1{equation~\ref{#1}}

\def\1{\bm{1}}

\DeclareMathAlphabet{\mathsfit}{\encodingdefault}{\sfdefault}{m}{sl}
\SetMathAlphabet{\mathsfit}{bold}{\encodingdefault}{\sfdefault}{bx}{n}

\usepackage{subcaption}
\usepackage{wrapfig}
\usepackage[most]{tcolorbox}
\usepackage{xcolor}
\usepackage[table]{xcolor}
\usepackage{hyperref}
\usepackage{url}
\usepackage{multirow}
\usepackage{amsmath}
\usepackage{amsfonts}
\usepackage{graphicx}
\usepackage{listings}
\usepackage{xspace}
\usepackage{enumitem}
\usepackage{seqsplit}
\usepackage{fancyvrb}
\usepackage{fvextra}
\usepackage{caption}
\usepackage{booktabs}
\usepackage{tabularx}
\usepackage{longtable}
\usepackage{etoolbox}
\usepackage{arydshln}
\usepackage{xstring}
\usepackage{pifont}
\usepackage{epigraph}
\usepackage[export]{adjustbox}
\usepackage{placeins}

\AtBeginEnvironment{tabular}{\renewcommand{\arraystretch}{0.92}}
\AtBeginEnvironment{tabular*}{\renewcommand{\arraystretch}{0.92}}
\AtBeginEnvironment{tabularx}{\renewcommand{\arraystretch}{0.92}}
\AtBeginEnvironment{longtable}{\renewcommand{\arraystretch}{0.92}}

\definecolor{softgreen}{RGB}{110, 160, 120}
\definecolor{earlyblue}{HTML}{88A2F1}
\definecolor{midgrey}{HTML}{fadcb4}
\definecolor{latered}{HTML}{EE9C88}
\definecolor{highlightgreen}{HTML}{80c66d}
\definecolor{highlightpurple}{HTML}{9b6d97}
\definecolor{deltaup}{HTML}{DCEAF7}
\definecolor{deltadown}{HTML}{FBE4D5}
\definecolor{deltaflat}{HTML}{ECEFF1}

\newtcolorbox{takeawaybox_basemodel}[1]{
    colback=orange!5!white,
    colframe=black,
    arc=5pt,
    outer arc=5pt,
    boxrule=0.8pt,
    left=5pt,
    right=5pt,
    top=4pt,
    bottom=4pt,
    fontupper=\small,
    enhanced,
    before upper={\textbf{#1 }}
}

\newcounter{promptfigure}[section]
\renewcommand{\thepromptfigure}{\thesection.\arabic{promptfigure}}
\newcommand{\promptfigure}[2]{%
  \refstepcounter{promptfigure}%
  \tcbinputlisting{
    enhanced,
    breakable,
    listing only,
    listing engine=listings,
    listing file={#2},
    title={Prompt \thepromptfigure: #1},
    title after break={Prompt \thepromptfigure: #1 (continued)},
    colback=white,
    colframe=black!55,
    colbacktitle=black!7,
    coltitle=black,
    fonttitle=\bfseries\small,
    boxrule=0.6pt,
    arc=0.8mm,
    left=1.8mm,
    right=1.8mm,
    top=1.2mm,
    bottom=1.2mm,
    toptitle=1.2mm,
    bottomtitle=1.2mm,
    before skip=8pt,
    after skip=10pt,
    listing options={
      basicstyle=\ttfamily\scriptsize,
      breaklines=true,
      breakatwhitespace=false,
      columns=fullflexible,
      keepspaces=true,
      showstringspaces=false
    }
  }%
}

\newcommand{\deltaval}[1]{%
  \IfBeginWith{#1}{+}{%
    {\textcolor{highlightgreen}{\textit{(#1)}}}%
  }{%
    \IfBeginWith{#1}{-}{%
      {\textcolor{highlightpurple}{\textit{(#1)}}}%
    }{%
      {\textit{(#1)}}%
    }%
  }%
}

\definecolor{findingteal}{HTML}{0B6E75}

\tcbset{
  finding style/.style={
    enhanced,
    colback=findingteal!3!white,
    colframe=findingteal!55!black,
    boxrule=0.5pt,
    arc=1mm,
    left=2mm, right=2mm,
    top=1.4mm, bottom=1.4mm,
    before skip=7pt, after skip=7pt,
    fontupper=\normalfont,
    colbacktitle=findingteal,
    coltitle=white,
    fonttitle=\normalfont\bfseries,
    lefttitle=2mm, righttitle=2mm,
    toptitle=0.3mm, bottomtitle=0.3mm,
    titlerule=0pt,
  }
}
\newtcolorbox{findingbox}[1][]{finding style,#1}

\title{{\fontsize{18}{22}\selectfont How Can Rhetoric Reward-Hack AI Reviewers?\\ Dissecting Rhetorical Sensitivity in AI-Based Peer Review}}

\author[*,1]{Ming Li}
\author[*,2]{Chenguang Wang}
\author[1]{Xirui Li}
\author[2]{Xinyue Zeng}
\author[]{Dianqi Li}
\author[4]{Peng Shi}
\author[2]{Dawei Zhou}
\author[3]{Tianyi Zhou}

\renewcommand\affiliation[2][]{%
  \addtolist[#1]{#2}{\affiliationlist}{\affiliationformat}{~~~~}%
}

\affiliation[1]{University of Maryland}
\affiliation[2]{Virginia Tech}
\affiliation[3]{MBZUAI}
\affiliation[4]{University of Waterloo}

\contribution[*]{Co-first Author}

\abstract{
As large language models increasingly participate in scientific evaluation, we investigate a potential form of reward hacking: how rhetorical choices shape AI-review judgments when reported scientific content is preserved and how these effects vary across evaluation conditions.
We construct a controlled corpus of 4,200 full-paper manuscripts derived from 120 anonymized ICLR 2026 submissions. Two LLM rewriters transform six rhetorical dimensions in opposing directions, and five LLM reviewers evaluate the resulting manuscripts under standard and strict protocols. We also test joint, recursive, and reviewer-guided rewriting.
Our results show that rhetorical sensitivity is structured rather than uniform. Evidence framing and novelty stance produce the largest positive-negative contrasts in overall assessment, with scope framing forming a weaker second tier; the remaining dimensions have smaller or less stable effects. This hierarchy persists across human-assessed quality levels, but score movement depends strongly on the AI reviewer's original score: lower scores tend to rise, higher scores tend to fall, and directional contrasts are clearest in the middle ranges. More elaborate workflows do not reliably yield larger gains. Joint rewriting is strongly rewriter-dependent, reviewer guidance does not consistently outperform an unguided second pass, and repeated rewriting yields diminishing, configuration-dependent returns. Across conditions, the rewriter primarily determines the separation between opposing variants, whereas the reviewer determines the magnitude and sign of their score effects. Strict review lowers mean OA by 1.36 points without consistently changing rhetorical sensitivity.
These findings identify when rhetorical presentation influences AI scientific review and motivate evaluation systems robust to content-preserving variation in scientific writing.

}

\date{\today}
\authoremails{\email{minglii@umd.edu}, \email{\{cswang, dzhou\}@vt.edu}, \email{tianyi.zhou@mbzuai.ac.ae}}
\metadata[Project Page]{\url{https://github.com/MingLiiii/Dissecting_AI_Reviews}\\}

\renewcommand{\topfraction}{0.82}
\renewcommand{\bottomfraction}{0.70}
\renewcommand{\textfraction}{0.16}
\renewcommand{\floatpagefraction}{0.86}
\begin{document}

\maketitle

\section{Introduction}

Large language models (LLMs) are increasingly involved on both sides of scientific evaluation: they can revise how authors present their work and serve as scalable evaluators that accelerate review and reduce reviewer burden \citep{wang2020reviewrobot,liang2024can,thakkar2026large,chen2026peercheck,kaneko2026paraphrasing,baumann2026stop}. As these uses meet in the same evaluation pipeline, a central concern is how rhetorical presentation can reward-hack AI reviewers by changing their judgments without corresponding improvements in the underlying science.
Presentation is especially relevant because the same scientific work can be communicated through different rhetorical choices: supported claims may be framed more assertively or cautiously, reported evidence may receive greater or lesser emphasis, and technical content may be expressed with different levels of complexity. Although such variation is routine in scientific writing \citep{hyland1998hedging,hyland2018metadiscourse,james2024rigour}, reviewers are expected to distinguish presentation from scientific merit \citep{iclr2026reviewer,icml2026reviewer,neurips2026reviewer,arr2025reviewer}. The issue is not whether reviewers should ignore presentation, since clarity and exposition legitimately matter, but whether content-preserving rhetorical choices systematically alter merit-related judgments. Understanding AI-based peer review, therefore, requires identifying which choices matter, whether particular directions are favored, and under what evaluation conditions their effects become consequential.
\textit{We neither advocate for nor against the use of AI in manuscript preparation or review; our goal is to better understand how these two uses interact in scientific evaluation.}

Existing work suggests that AI-review judgments can vary for reasons unrelated to changes in scientific evidence. In broader LLM-as-a-judge settings, prior studies have documented systematic evaluation biases, including preferences related to response position and verbosity, as well as sensitivity to evaluation design and inconsistency across repeated judgments~\citep{zheng2023judging,shi2025judging,dubois2024length,stureborg2024large}. Similar concerns arise in scientific peer review, where LLM-generated reviews align only partially with human judgments and may differ in the aspects of a manuscript they emphasize \citep{liang2024can,li2025unveiling,chen2026peercheck}. Although LLMs can provide useful paper-level feedback, they often underperform in identifying substantive weaknesses and show limited sensitivity to paper quality~\citep{liang2024can,li2025unveiling,zhu2025your}. Their evaluations can also be influenced by manuscript-level factors, including hidden prompt injection, adversarial textual perturbations, metadata-related biases, and presentation-related variation~\citep{ye2024we,collu2026misleading,lin2025breaking,vasu2026justice}. More directly, recent studies show that presentation-only revisions to scientific manuscripts can increase AI-review scores without changing the reported evidence~\citep{kaneko2026paraphrasing,li2026gaming,baumann2026stop,yang2026no}. However, \textbf{how this potential reward-hacking vulnerability is structured remains unclear: which dimensions of scientific rhetoric drive score changes, and are their directional effects stable across evaluators?}

Addressing this gap requires separating rhetorical presentation from scientific merit, which comparisons across naturally occurring papers cannot do because stronger papers often also have more polished and confident presentation \citep{kang2018dataset,fytas2021makes,james2024rigour}. We therefore construct matched versions of each manuscript that preserve its underlying scientific content and key methodology elements, while allowing the agent to rewrite the full narrative, including captions, transitions, and the presentation of reported results, rather than enforcing strict sentence-level semantic equivalence. This motivates our central research question: \textbf{\emph{Which rhetorical choices shape AI-review scores when the underlying scientific content is preserved, and how do their effects vary across evaluation conditions?}}

\begin{figure}[t]
  \centering
  \includegraphics[width=\textwidth]{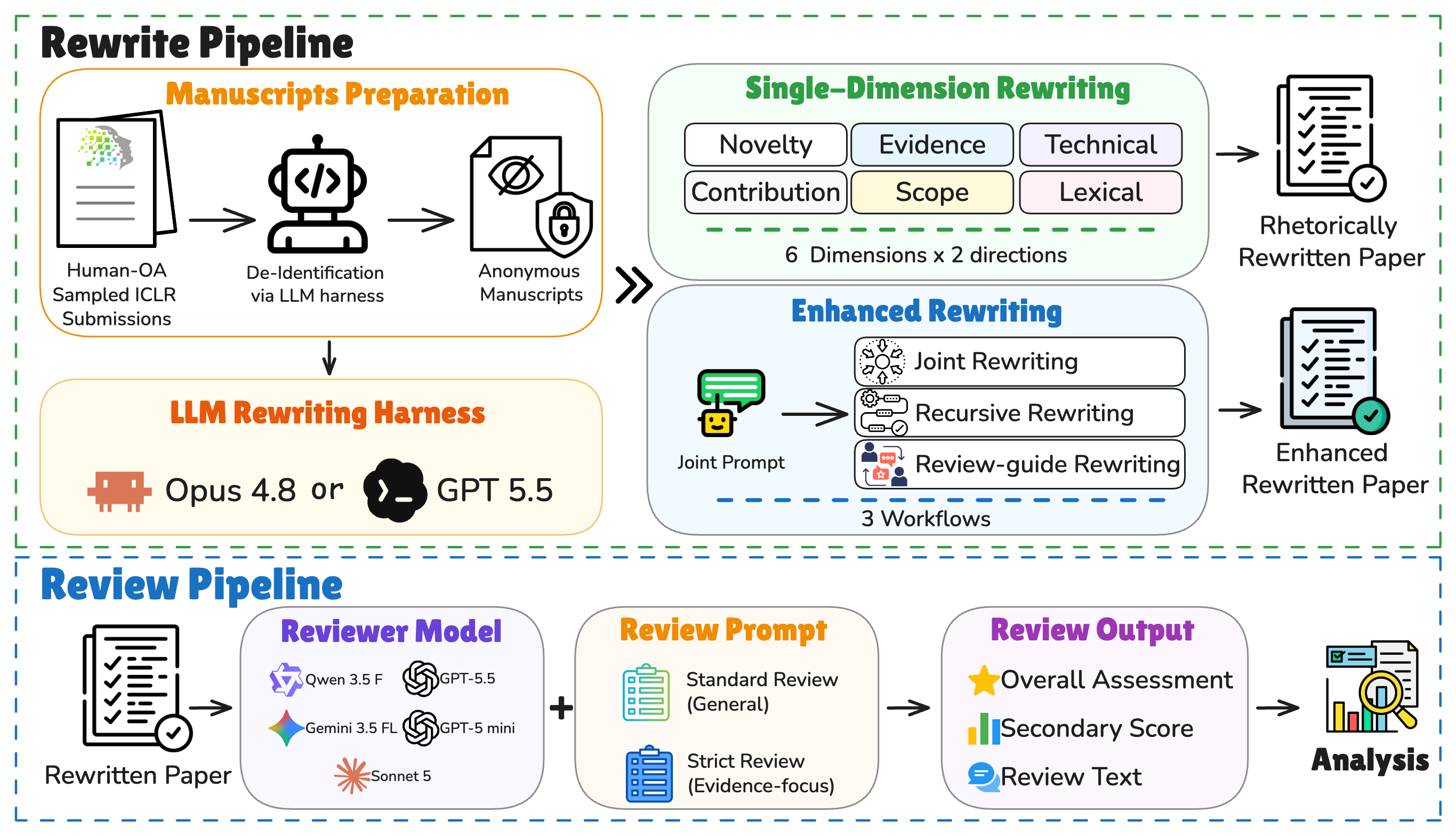}
  \caption{\textbf{Overview of the rewrite and review pipeline.} Anonymized manuscripts are rewritten through single-dimension or enhanced workflows and evaluated by five AI reviewers under standard and strict protocols.}
  \label{fig:rewrite_review_pipeline}
\end{figure}

\begin{figure}[t]
  \centering
  \includegraphics[width=\textwidth]{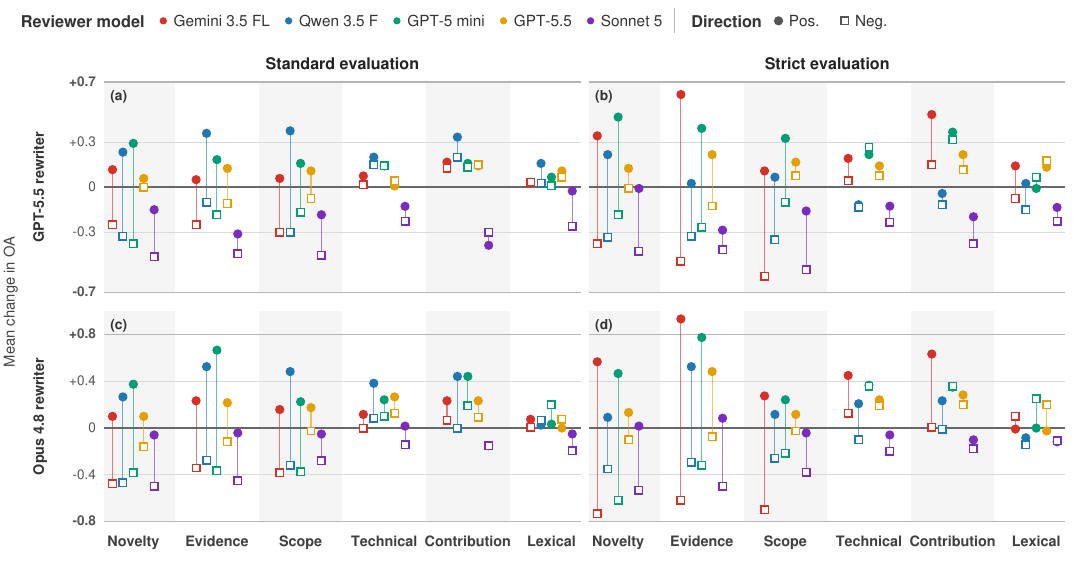}
  \caption{\textbf{Overview of rhetorical effects on AI scientific review.} Evidence framing and novelty stance produce the largest positive-negative contrasts in overall assessment. Rows correspond to the GPT-5.5 and Opus~4.8 rewriters, columns to standard and strict evaluation, and colors to the reviewer models. Filled circles and open squares denote positive and negative rewrites, respectively.}
  \label{fig:dimension_model_pair_rating_changes}
\end{figure}

To answer this question, we construct a controlled corpus of 4,200 full-paper manuscripts, comprising 120 anonymized originals and 4,080 rewrites derived from ICLR 2026 submissions. 
Our framework operationalizes 6 dimensions of scientific rhetoric (Table~\ref{tab:rewrite_dimensions}): (i) claim and novelty stance, (ii) scope and generalization, (iii) quantitative evidence framing, (iv) contribution structure, (v) technical register and formalism, and (vi) lexical and syntactic complexity. An agentic LLM rewriting harness modifies each dimension in opposing directions across the full manuscript and may change prose, captions, tables, and the presentation of some numerical and mathematical expressions. To maximize the scope of presentational change, programmatic checks are limited to structural anchors such as citation and cross-reference keys, labels, graphics paths, supported environments, code and algorithms, and bibliography files. We evaluate each original and rewrite with multiple LLM reviewers under standard and strict protocols, using matched comparisons to measure dimension-specific score changes and their consistency across reviewers and conditions. We additionally examine joint, recursive, and reviewer-guided workflows. Figure~\ref{fig:rewrite_review_pipeline} illustrates the complete pipeline, Figure~\ref{fig:dimension_model_pair_rating_changes} previews the resulting variation across experimental conditions, and Table~\ref{tab:study_design_overview} summarizes the experimental scope.

\begin{table*}[t]
\centering
\caption{\textbf{Overview of the experimental framework and scale.}
The table summarizes the scope of the study.}
\label{tab:study_design_overview}
\small
\setlength{\tabcolsep}{6pt}
\begin{adjustbox}{max width=\textwidth}
\begin{tabular}{@{}llr@{}}
\toprule
\textbf{Component} & \textbf{Specification} & \textbf{Scale} \\
\midrule
\textbf{Corpus} &
ICLR 2026 submissions &
\textbf{120} papers \\

\midrule
\textbf{Intervention space} &
6 dimensions $\times$ 2 directions &
\textbf{12} conditions \\
\textbf{Rewrite models} &
\begin{tabular}[t]{@{}l@{}}
\emph{GPT-5.5} via Codex CLI\\
\emph{Opus 4.8} via Claude Code
\end{tabular} &
\textbf{2} models \\
\textbf{Single-dimension rewrite} &
One dimension at a time &
\textbf{2,880} rewrites \\
\textbf{Joint rewrite} &
All six positive-direction objectives &
\textbf{240} rewrites \\
\textbf{Recursive joint rewrite} &
joint prompt reapplied &
\textbf{480} rewrites \\
\textbf{Reviewer-guided rewrite} &
Review feedback before the final rewrite &
\textbf{480} rewrites \\

\midrule
\textbf{AI reviewers} &
\begin{tabular}[t]{@{}l@{}}
\emph{Gemini 3.5 Flash-Lite} (\emph{Gemini 3.5 FL}),\\
\emph{Qwen 3.5 Flash} (\emph{Qwen 3.5 F}),\\
\emph{GPT-5 mini}, \emph{GPT-5.5}, \emph{Claude Sonnet 5} (\emph{Sonnet 5})
\end{tabular} &
\textbf{5} models \\
\textbf{Review prompts} &
\emph{Standard}, \emph{Strict}&
\textbf{2} primary prompts \\
\textbf{Valid review records} &
 &
\textbf{42,396} records \\

\midrule
\textbf{Full manuscript corpus} &
120 originals $+$ 4,080 rewrites &
\textbf{4,200} manuscripts \\
\textbf{Direct API cost} &
rewrite and review calls &
\textbf{\$29,165.89} \\

\bottomrule
\end{tabular}
\end{adjustbox}
\end{table*}

\textbf{Key Findings.}
\begin{itemize}[leftmargin=*,labelindent=0pt]
  \item \textbf{Finding 1: Rewrite sensitivity concentrates in specific dimensions and varies with the AI reviewer’s initial score.} Evidence framing and novelty stance produce the largest effects, followed by scope framing: positive evidence framing \textit{raises} Overall Assessments (OA) by up to \textit{0.93}, negative novelty stance \textit{lowers} it by up to \textit{0.73}, and evidence framing changes weak-accept probability by \textit{13} percentage points on average. 
  This hierarchy is stable across human-assessed quality levels, but lower initial AI scores tend to rise, higher scores tend to fall, and directional contrasts are strongest in the middle ranges.

  \item \textbf{Finding 2: More elaborate rewriting yields configuration-dependent and diminishing gains.} 
  Joint rewriting produces substantial gains with Opus~4.8 but near-zero gains with GPT-5.5. Reviewer guidance does not consistently outperform an unguided second pass at the same depth, and gains diminish after the second pass. 

  \item \textbf{Finding 3: Rewriters shape rhetorical contrasts, reviewers shape rewrite effects, and protocols shift the score scale.} Rewriter mainly changes the separation between positive and negative versions, whereas the reviewer changes both the magnitude and the sign of the resulting OA effects. Strict review \textit{lowers} absolute OA by \textit{1.36 points} on average but does not consistently strengthen or weaken rewrite effects. The reviewer also determines whether OA changes are reflected primarily in contribution or soundness.
\end{itemize}

Together, these findings show that when AI assists both manuscript revision and review, upstream rhetorical choices can become reviewer-dependent signals in downstream merit scores. Rhetoric, therefore, does not provide a universal reward hack; its influence is selective, configuration-dependent, and sometimes counterproductive.

\textbf{Contributions.}
\begin{itemize}
    \item We develop an end-to-end controlled analysis framework that
    combines controlled rhetorical rewriting, multi-model AI review, and paired
    analysis. The framework supports single-dimension, joint, recursive, and
    reviewer-guided interventions.

    \item We implement an agentic source-level rewriting harness for complete
    \LaTeX{} projects. It uses content-preservation constraints, programmatic
    checks, compilation, and repair to produce reviewable full-paper variants.

    \item We systematically analyze how rhetorical effects vary by dimension and
    direction, how gains change under joint and recursive rewriting, and how
    rewriters, reviewers, and review protocols shape OA and the secondary rubric
    scores.
\end{itemize}

\section{Related Work}
\label{sec:related_work}

In broader LLM evaluations, rubric-based methods can align with human preferences but remain sensitive to prompts, presentation, sampling, and evaluator identity~\citep{zheng2023judging,bavaresco2025llms,panickssery2024llm, li2026rethinking}. In scientific review, LLMs can provide feedback but irreliably identify substantive weaknesses and distinguish paper quality~\citep{liang2024can,li2025unveiling,chen2026peercheck}. Manuscript-side studies also show that hidden instructions, artificial perturbations, and meaning-preserving rewrites alter AI-review outcomes~\citep{ye2024we,lin2025breaking,kaneko2026paraphrasing,baumann2026stop,yang2026no}. However, most rewriting work seeks to alter ratings successfully, without systematically characterizing how rhetorical rewrites influence AI reviewers. \textbf{\emph{Detailed related work is in Appendix~\ref{app:detailed_related_work}.}}

\section{A Controlled Analysis Framework for Rhetorical Rewriting in AI Review}
\label{sec:framework}

We study where AI-review judgments are most responsive when rhetorical presentation varies while scientific content is preserved. The framework has three stages: constructing a verified manuscript baseline and controlled rhetorical variants, evaluating the resulting manuscripts under blinded AI-review conditions, and comparing judgments within the same paper. It covers single-dimension, joint, recursive, and reviewer-guided rewriting. Table~\ref{tab:study_design_overview} summarizes the experimental scope.

\subsection{Controlled Dataset and Rewriting}

\subsubsection{Seed Corpus and Baseline Construction}

The seed corpus is constructed from ICLR 2026 submissions retrieved through the OpenReview API~\footnote{\url{https://docs.openreview.net/reference/api-v2}}. We retain submissions with valid assessments and review opinions while excluding withdrawn and desk-rejected papers. To cover a broad quality range, we sample 20 papers from each of six intervals of mean human overall rating: $[1,3)$, $[3,4)$, $[4,5)$, $[5,6)$, $[6,7)$, and $[7,8.5]$, yielding 120 papers.

Candidate public arXiv \LaTeX{} sources are matched to OpenReview submissions using manuscript metadata. When an arXiv record has multiple versions, we compare normalized source text with the OpenReview PDF using token-based cosine similarity and 5-gram Jaccard similarity, retaining the version with the strongest overall correspondence.

Before rewriting, an agentic anonymization LLM harness removes author names, affiliations, acknowledgments, submission-status statements, and identity-bearing resource links. Source-difference checks restrict edits to identity- and status-related regions. We then manually inspect the edited source and compiled PDF against the submission to confirm that identifying information has been removed while other content remains unchanged. This verified project is the common baseline from which every controlled variant is generated independently.

\begin{table}[t]
\centering
\caption{\textbf{Definition of the rhetorical intervention space.}
Each dimension is rewritten in opposing directions while preserving scientific content. Corresponding rewrite prompts are provided in Appendix~\ref{app:prompts}.}
\label{tab:rewrite_dimensions}
\footnotesize
\setlength{\tabcolsep}{2pt}
\begin{tabularx}{\textwidth}{@{}l
>{\hsize=1.08\hsize\raggedright\arraybackslash}X
>{\hsize=0.92\hsize\raggedright\arraybackslash}X@{}}
\toprule
\textbf{Dimension} & \textbf{Positive direction} & \textbf{Negative direction} \\
\midrule

Novelty stance
& More assertive framing of supported claims and novelty
& More cautious framing of claims and novelty \\
\midrule

Scope framing
& Broader framing within supported boundaries
& Narrower framing tied to evaluated settings \\
\midrule

Evidence framing
& Greater emphasis on reported comparisons and patterns
& Less emphasis on comparative advantage \\
\midrule

Contribution salience
& Contributions explicitly foregrounded and signposted
& Contributions integrated into the narrative \\
\midrule

Technical register
& More formal and explicit technical expression
& Plainer expression of technical content \\
\midrule

Linguistic complexity
& More sophisticated vocabulary and syntax
& Simpler vocabulary and syntax \\

\bottomrule
\end{tabularx}
\end{table}

\subsubsection{Rhetorical Intervention Design}

We define six prespecified dimensions of rhetorical presentation from reviewer guidelines used by major AI venues: claim and novelty stance, scope and generalization, quantitative evidence framing, contribution structure, technical register and formalism, and lexical and syntactic complexity~\citep{iclr2026reviewer,icml2026reviewer,neurips2026reviewer,arr2025reviewer,larson2012systematic,fytas2021makes,james2024rigour,li2025developing}. These dimensions represent distinct, reviewer-relevant ways in which presentation may shape the interpretation of the same scientific contribution.

The two variants for each dimension are generated independently from the same baseline. ``Positive'' and ``negative'' identify the intended rhetorical intervention direction. This bidirectional design separates directional sensitivity from score movement caused by rewriting itself. Every rewriting combines a dimension-specific instruction with the same project-level preservation constraint.

\subsubsection{Source-Level Rewrite Harness}

The agentic rewriting harness operates on complete \LaTeX{} projects. GPT-5.5 through the Codex CLI~\citep{openai_gpt55} and Opus~4.8 through Claude Code~\citep{anthropic_claude_opus48} inspect each project, identify editable manuscript prose, and propagate the requested intervention through the full narrative. The rewrite scope deliberately maximizes coverage across the manuscript: the agent may revise wording, organization, emphasis, captions, tables, and the presentation and interpretation of reported results across the full manuscript, under soft constraints to preserve the underlying methods, experimental settings, reported values and comparisons, evidence boundaries, substantive findings, and scientific meaning without requiring sentence-level semantic equivalence.

After editing, a programmatic checker compares protected structural anchors with the anonymized baseline. These anchors include citation and cross-reference keys, labels, URLs, graphics paths, supported mathematical environments, code and algorithm environments, and bibliography files. These structural checks complement the prompt-level preservation constraint while allowing broad document-level changes in scientific presentation. Detected violations are returned to the rewriting agent for repair, while unresolved violations or compilation failures cause the output to be discarded.

\subsection{Single-Dimension Rewriting}

For every paper, each rewrite model independently generates both directions of all six dimensions from the same anonymized baseline. This produces 12 variants per paper and rewriter, or 2,880 rewritten manuscripts in total. No single-dimension intervention is applied on top of another.

\begin{table*}[t]
  \centering
  \caption{\textbf{Overview of rhetorical presentation effects in AI review.}
Evidence framing and novelty stance produce the largest changes in overall
assessment, followed by scope framing, with effects varying across rewriters,
reviewers, and review protocols. Entries report paper-level mean changes from
the prompt-matched original for positive (Pos.) and negative (Neg.)
interventions. The two rightmost columns average across dimensions; summary
rows average across reviewers, rewriters, and protocols at the indicated
levels. Green and yellow denote increases and decreases, with intensity
proportional to magnitude. Additional statistics and 95\% bootstrap confidence
intervals are reported in Appendix~\ref{app:dimension_rating_details}.}
  \label{tab:dimension_model_pair_rating_changes}
  \scriptsize
  \renewcommand{\arraystretch}{1.03}
  \setlength{\tabcolsep}{1.8pt}
  \newcommand{\increasecell}[2]{\cellcolor{green!#1}#2}
  \newcommand{\decreasecell}[2]{\cellcolor{yellow!#1}#2}
  \begin{adjustbox}{max width=\textwidth}
  \begin{tabular}{@{}ll*{14}{r}@{}}
    \toprule
    \multirow{2}{*}{\textbf{Rewriter}}
      & \multirow{2}{*}{\textbf{Reviewer}}
      & \multicolumn{2}{c}{\textbf{Novelty}}
      & \multicolumn{2}{c}{\textbf{Evidence}}
      & \multicolumn{2}{c}{\textbf{Scope}}
      & \multicolumn{2}{c}{\textbf{Technical}}
      & \multicolumn{2}{c}{\textbf{Contribution}}
      & \multicolumn{2}{c}{\textbf{Lexical}}
      & \multicolumn{2}{c}{\textbf{Across dimensions}} \\
    \cmidrule(lr){3-4}\cmidrule(lr){5-6}\cmidrule(lr){7-8}
    \cmidrule(lr){9-10}\cmidrule(lr){11-12}\cmidrule(lr){13-14}
    \cmidrule(l){15-16}
      & & \multicolumn{1}{c}{\textbf{\textsf{Pos.}}}
      & \multicolumn{1}{c}{\textbf{\textsf{Neg.}}}
      & \multicolumn{1}{c}{\textbf{\textsf{Pos.}}}
      & \multicolumn{1}{c}{\textbf{\textsf{Neg.}}}
      & \multicolumn{1}{c}{\textbf{\textsf{Pos.}}}
      & \multicolumn{1}{c}{\textbf{\textsf{Neg.}}}
      & \multicolumn{1}{c}{\textbf{\textsf{Pos.}}}
      & \multicolumn{1}{c}{\textbf{\textsf{Neg.}}}
      & \multicolumn{1}{c}{\textbf{\textsf{Pos.}}}
      & \multicolumn{1}{c}{\textbf{\textsf{Neg.}}}
      & \multicolumn{1}{c}{\textbf{\textsf{Pos.}}}
      & \multicolumn{1}{c}{\textbf{\textsf{Neg.}}}
      & \multicolumn{1}{c}{\textbf{\textsf{Pos.}}}
      & \multicolumn{1}{c}{\textbf{\textsf{Neg.}}} \\
    \midrule
    \multicolumn{16}{l}{\textit{Standard review prompt}} \\
    \cmidrule(lr){1-16}
    GPT-5.5 & Gemini 3.5 FL & \increasecell{17}{+.117} & \decreasecell{34}{-.250} & \increasecell{11}{+.050} & \decreasecell{34}{-.250} & \increasecell{12}{+.058} & \decreasecell{37}{-.300} & \increasecell{14}{+.075} & \increasecell{6}{+.017} & \increasecell{21}{+.167} & \increasecell{18}{+.125} & \increasecell{9}{+.033} & \increasecell{9}{+.033} & +.083 & -.104 \\
     & Qwen 3.5 F & \increasecell{24}{+.233} & \decreasecell{38}{-.325} & \increasecell{30}{+.358} & \decreasecell{21}{-.100} & \increasecell{31}{+.375} & \decreasecell{37}{-.300} & \increasecell{22}{+.200} & \increasecell{19}{+.150} & \increasecell{29}{+.333} & \increasecell{22}{+.200} & \increasecell{20}{+.158} & \increasecell{8}{+.025} & +.276 & -.058 \\
     & GPT-5 mini & \increasecell{27}{+.292} & \decreasecell{41}{-.375} & \increasecell{22}{+.183} & \decreasecell{29}{-.183} & \increasecell{20}{+.158} & \decreasecell{27}{-.167} & \increasecell{19}{+.142} & \increasecell{19}{+.142} & \increasecell{20}{+.158} & \increasecell{18}{+.133} & \increasecell{13}{+.067} & \increasecell{5}{+.008} & +.167 & -.074 \\
     & GPT-5.5 & \increasecell{12}{+.058} & 0.000 & \increasecell{18}{+.125} & \decreasecell{22}{-.108} & \increasecell{17}{+.108} & \decreasecell{18}{-.075} & \increasecell{5}{+.008} & \increasecell{10}{+.042} & \increasecell{19}{+.142} & \increasecell{19}{+.150} & \increasecell{17}{+.108} & \increasecell{13}{+.067} & +.092 & +.012 \\
     & Sonnet 5 & \decreasecell{26}{-.150} & \decreasecell{46}{-.462} & \decreasecell{37}{-.311} & \decreasecell{45}{-.442} & \decreasecell{29}{-.183} & \decreasecell{45}{-.454} & \decreasecell{24}{-.127} & \decreasecell{32}{-.225} & \decreasecell{42}{-.387} & \decreasecell{37}{-.299} & \decreasecell{11}{-.025} & \decreasecell{34}{-.258} & -.197 & -.357 \\
    \cmidrule(lr){2-16}
    & \textit{GPT-5.5 mean} & +.110 & -.282 & +.081 & -.217 & +.103 & -.259 & +.060 & +.025 & +.083 & +.062 & +.068 & -.025 & +.084 & -.116 \\
    \cmidrule(lr){1-16}
    Opus 4.8 & Gemini 3.5 FL & \increasecell{16}{+.100} & \decreasecell{46}{-.475} & \increasecell{24}{+.233} & \decreasecell{39}{-.342} & \increasecell{20}{+.158} & \decreasecell{42}{-.383} & \increasecell{17}{+.117} & 0.000 & \increasecell{24}{+.233} & \increasecell{13}{+.067} & \increasecell{14}{+.075} & \increasecell{5}{+.008} & +.153 & -.188 \\
     & Qwen 3.5 F & \increasecell{26}{+.267} & \decreasecell{46}{-.467} & \increasecell{36}{+.525} & \decreasecell{35}{-.275} & \increasecell{35}{+.483} & \decreasecell{38}{-.317} & \increasecell{31}{+.383} & \increasecell{15}{+.083} & \increasecell{33}{+.442} & 0.000 & \increasecell{8}{+.025} & \increasecell{13}{+.067} & +.354 & -.151 \\
     & GPT-5 mini & \increasecell{31}{+.375} & \decreasecell{42}{-.383} & \increasecell{41}{+.667} & \decreasecell{41}{-.367} & \increasecell{24}{+.225} & \decreasecell{41}{-.375} & \increasecell{25}{+.242} & \increasecell{16}{+.100} & \increasecell{33}{+.442} & \increasecell{22}{+.192} & \increasecell{9}{+.033} & \increasecell{22}{+.200} & +.331 & -.106 \\
     & GPT-5.5 & \increasecell{16}{+.100} & \decreasecell{27}{-.158} & \increasecell{23}{+.217} & \decreasecell{23}{-.117} & \increasecell{21}{+.175} & \decreasecell{11}{-.025} & \increasecell{26}{+.267} & \increasecell{18}{+.125} & \increasecell{24}{+.233} & \increasecell{15}{+.092} & 0.000 & \increasecell{14}{+.075} & +.165 & -.001 \\
     & Sonnet 5 & \decreasecell{16}{-.059} & \decreasecell{47}{-.496} & \decreasecell{14}{-.042} & \decreasecell{45}{-.449} & \decreasecell{15}{-.051} & \decreasecell{35}{-.280} & \increasecell{7}{+.017} & \decreasecell{25}{-.143} & \decreasecell{26}{-.151} & \decreasecell{26}{-.153} & \decreasecell{15}{-.050} & \decreasecell{30}{-.195} & -.056 & -.286 \\
    \cmidrule(lr){2-16}
    & \textit{Opus 4.8 mean} & +.156 & -.396 & +.320 & -.310 & +.198 & -.276 & +.205 & +.033 & +.240 & +.039 & +.017 & +.031 & +.189 & -.146 \\
    \cmidrule(lr){1-16}
    \multicolumn{2}{l}{\textit{Standard prompt mean}} & +.133 & -.339 & +.201 & -.263 & +.151 & -.268 & +.132 & +.029 & +.161 & +.051 & +.042 & +.003 & +.137 & -.131 \\
    \midrule
    \multicolumn{16}{l}{\textit{Strict review prompt}} \\
    \cmidrule(lr){1-16}
    GPT-5.5 & Gemini 3.5 FL & \increasecell{29}{+.342} & \decreasecell{41}{-.375} & \increasecell{40}{+.617} & \decreasecell{47}{-.492} & \increasecell{17}{+.108} & \decreasecell{52}{-.592} & \increasecell{22}{+.192} & \increasecell{10}{+.042} & \increasecell{35}{+.483} & \increasecell{19}{+.150} & \increasecell{19}{+.142} & \decreasecell{18}{-.075} & +.314 & -.224 \\
     & Qwen 3.5 F & \increasecell{23}{+.217} & \decreasecell{39}{-.333} & \increasecell{8}{+.025} & \decreasecell{38}{-.325} & \increasecell{13}{+.067} & \decreasecell{40}{-.350} & \decreasecell{23}{-.117} & \decreasecell{24}{-.133} & \decreasecell{14}{-.042} & \decreasecell{23}{-.117} & \increasecell{8}{+.025} & \decreasecell{26}{-.150} & +.029 & -.235 \\
     & GPT-5 mini & \increasecell{34}{+.467} & \decreasecell{29}{-.183} & \increasecell{31}{+.392} & \decreasecell{35}{-.267} & \increasecell{29}{+.325} & \decreasecell{21}{-.100} & \increasecell{23}{+.217} & \increasecell{26}{+.267} & \increasecell{30}{+.367} & \increasecell{28}{+.317} & \decreasecell{6}{-.008} & \increasecell{13}{+.067} & +.293 & +.017 \\
     & GPT-5.5 & \increasecell{18}{+.125} & \decreasecell{6}{-.008} & \increasecell{23}{+.217} & \decreasecell{24}{-.125} & \increasecell{21}{+.167} & \increasecell{14}{+.075} & \increasecell{19}{+.142} & \increasecell{14}{+.075} & \increasecell{23}{+.217} & \increasecell{17}{+.117} & \increasecell{18}{+.133} & \increasecell{21}{+.175} & +.167 & +.051 \\
     & Sonnet 5 & \decreasecell{6}{-.008} & \decreasecell{44}{-.427} & \decreasecell{36}{-.286} & \decreasecell{43}{-.417} & \decreasecell{27}{-.158} & \decreasecell{50}{-.550} & \decreasecell{24}{-.126} & \decreasecell{32}{-.233} & \decreasecell{30}{-.197} & \decreasecell{41}{-.376} & \decreasecell{25}{-.134} & \decreasecell{32}{-.225} & -.152 & -.371 \\
    \cmidrule(lr){2-16}
    & \textit{GPT-5.5 mean} & +.228 & -.265 & +.193 & -.325 & +.102 & -.303 & +.061 & +.003 & +.166 & +.018 & +.031 & -.042 & +.130 & -.152 \\
    \cmidrule(lr){1-16}
    Opus 4.8 & Gemini 3.5 FL & \increasecell{38}{+.567} & \decreasecell{57}{-.733} & \increasecell{45}{+.933} & \decreasecell{53}{-.617} & \increasecell{26}{+.275} & \decreasecell{56}{-.700} & \increasecell{34}{+.450} & \increasecell{18}{+.125} & \increasecell{40}{+.633} & \increasecell{5}{+.008} & \decreasecell{6}{-.008} & \increasecell{16}{+.100} & +.475 & -.303 \\
     & Qwen 3.5 F & \increasecell{15}{+.092} & \decreasecell{40}{-.350} & \increasecell{36}{+.525} & \decreasecell{36}{-.292} & \increasecell{17}{+.117} & \decreasecell{34}{-.258} & \increasecell{23}{+.208} & \decreasecell{21}{-.100} & \increasecell{24}{+.233} & \decreasecell{6}{-.008} & \decreasecell{19}{-.083} & \decreasecell{25}{-.142} & +.182 & -.192 \\
     & GPT-5 mini & \increasecell{34}{+.467} & \decreasecell{53}{-.617} & \increasecell{44}{+.775} & \decreasecell{38}{-.317} & \increasecell{25}{+.242} & \decreasecell{31}{-.217} & \increasecell{30}{+.367} & \increasecell{30}{+.358} & \increasecell{30}{+.350} & \increasecell{30}{+.358} & 0.000 & \increasecell{25}{+.250} & +.367 & -.031 \\
     & GPT-5.5 & \increasecell{18}{+.133} & \decreasecell{21}{-.100} & \increasecell{35}{+.483} & \decreasecell{18}{-.075} & \increasecell{17}{+.117} & \decreasecell{11}{-.025} & \increasecell{25}{+.242} & \increasecell{22}{+.192} & \increasecell{27}{+.283} & \increasecell{22}{+.200} & \decreasecell{11}{-.025} & \increasecell{22}{+.200} & +.206 & +.065 \\
     & Sonnet 5 & \increasecell{7}{+.017} & \decreasecell{49}{-.534} & \increasecell{15}{+.084} & \decreasecell{47}{-.496} & \decreasecell{14}{-.042} & \decreasecell{41}{-.379} & \decreasecell{16}{-.059} & \decreasecell{30}{-.197} & \decreasecell{21}{-.101} & \decreasecell{28}{-.175} & \decreasecell{23}{-.118} & \decreasecell{22}{-.108} & -.036 & -.315 \\
    \cmidrule(lr){2-16}
    & \textit{Opus 4.8 mean} & +.255 & -.467 & +.560 & -.359 & +.142 & -.316 & +.242 & +.076 & +.280 & +.077 & -.047 & +.060 & +.239 & -.155 \\
    \cmidrule(lr){1-16}
    \multicolumn{2}{l}{\textit{Strict prompt mean}} & +.242 & -.366 & +.376 & -.342 & +.122 & -.310 & +.152 & +.040 & +.223 & +.047 & -.008 & +.009 & +.184 & -.154 \\
    \midrule
    \multicolumn{2}{l}{\textbf{Overall mean}} & \textbf{+.187} & \textbf{-.353} & \textbf{+.289} & \textbf{-.303} & \textbf{+.136} & \textbf{-.289} & \textbf{+.142} & \textbf{+.034} & \textbf{+.192} & \textbf{+.049} & \textbf{+.017} & \textbf{+.006} & \textbf{+.161} & \textbf{-.142} \\
    \bottomrule
  \end{tabular}
  \end{adjustbox}
\end{table*}

\subsection{Enhanced Rewriting}
\label{app:multistage_rewriting}

\paragraph{\textbf{Joint rewriting.}}
The joint workflow combines the positive directions of all six rhetorical dimensions in one prompt and applies them simultaneously to the anonymized original \LaTeX{} project. In this coordinated rewriting, the rewrite model strengthens supported claims and novelty statements, broadens scope only where justified, foregrounds existing quantitative evidence and contributions, increases technical precision, and uses more sophisticated but readable academic prose. These objectives are applied jointly at the paragraph level rather than as six sequential edits or a sentence-level checklist. The model rewrites substantial prose throughout the paper, including relevant captions and result discussion, while preserving the methods, settings, reported values, evidence boundaries, protected \LaTeX{} structures, and substantive findings. GPT-5.5 and Opus~4.8 each perform this pass independently for every paper, producing 240 joint manuscripts.

\paragraph{\textbf{Recursive rewriting.}}
The recursive workflow applies the joint rewriting procedure for three rounds. Round~1 rewrites the anonymized original, while each subsequent round rewrites the output of the preceding round using the same prompt and preservation requirements. This produces two additional rewritten manuscripts per paper and rewriter, or 480 manuscripts in total.

\paragraph{\textbf{Reviewer-guided rewriting.}}
The reviewer-guided workflow first applies the joint rewrite to produce an Intermediate manuscript, obtains one standard review from the rewriting backbone model, and then rewrites the manuscript once more using that feedback. Specifically, for each paper and rewrite model, the Intermediate manuscript is produced independently from the anonymized original using the joint prompt and shared preservation requirements. The rewriting backbone model reviews the compiled Intermediate PDF under the standard review protocol, and the structured review is validated. To produce the Final manuscript, the same rewrite model receives the Intermediate project together with a wrapper containing the validated review JSON, the complete 6-dimension prompt, and the shared requirements. It is instructed to preserve strengths, address weaknesses and questions through manuscript revisions, and moderate any rhetorical dimension that the review identifies as excessive, overstated, too broad, overly formal, or unnecessarily complex. This second pass produces one Final manuscript for each Intermediate manuscript, yielding 480 Intermediate and Final rewrites in total.

\begin{table*}[t]
  \centering
  \caption{\textbf{Rhetorical effects across human OA ranges under standard
evaluation.}
Entries report mean changes in the overall assessment relative to the
prompt-matched original, stratified by human OA range. The strict-evaluation
counterpart is reported in
Appendix~\ref{app:strict_human_score_ranges}. \(\dagger\) marks ranges
containing fewer than five papers.}
  \label{tab:human_score_directional_separation}
  \scriptsize
  \setlength{\tabcolsep}{1.3pt}
  \renewcommand{\arraystretch}{0.78}
  \newcommand{\scoreup}[2]{\cellcolor{green!#1}#2}
  \newcommand{\scoredown}[2]{\cellcolor{yellow!#1}#2}
  \newcommand{\scorezero}[1]{#1}
  \begin{adjustbox}{width=0.8\textwidth}
  \begin{tabular}{@{}ll*{14}{r}@{}}
    \toprule
    \textbf{Rewriter} & \textbf{Reviewer}
      & \multicolumn{2}{c}{\textbf{Novelty}}
      & \multicolumn{2}{c}{\textbf{Evidence}}
      & \multicolumn{2}{c}{\textbf{Scope}}
      & \multicolumn{2}{c}{\textbf{Technical}}
      & \multicolumn{2}{c}{\textbf{Contribution}}
      & \multicolumn{2}{c}{\textbf{Lexical}}
      & \multicolumn{2}{c}{\shortstack{\textbf{Across}\\\textbf{dimensions}}} \\
    \cmidrule(lr){3-4}\cmidrule(lr){5-6}\cmidrule(lr){7-8}
    \cmidrule(lr){9-10}\cmidrule(lr){11-12}\cmidrule(lr){13-14}
    \cmidrule(l){15-16}
      & & \multicolumn{1}{c}{\textbf{Pos.}} & \multicolumn{1}{c}{\textbf{Neg.}}
      & \multicolumn{1}{c}{\textbf{Pos.}} & \multicolumn{1}{c}{\textbf{Neg.}}
      & \multicolumn{1}{c}{\textbf{Pos.}} & \multicolumn{1}{c}{\textbf{Neg.}}
      & \multicolumn{1}{c}{\textbf{Pos.}} & \multicolumn{1}{c}{\textbf{Neg.}}
      & \multicolumn{1}{c}{\textbf{Pos.}} & \multicolumn{1}{c}{\textbf{Neg.}}
      & \multicolumn{1}{c}{\textbf{Pos.}} & \multicolumn{1}{c}{\textbf{Neg.}}
      & \multicolumn{1}{c}{\textbf{Pos.}} & \multicolumn{1}{c}{\textbf{Neg.}} \\
    \midrule
    \multicolumn{16}{l}{\textit{human OA [1,3]}} \\
    \cmidrule(lr){1-16}
    GPT-5.5 & Gemini 3.5 FL & \scoreup{27}{+.350} & \scoredown{40}{-.450} & \scoreup{22}{+.250} & \scoredown{42}{-.500} & \scoreup{21}{+.225} & \scoredown{48}{-.650} & \scoreup{16}{+.125} & \scoreup{17}{+.150} & \scoreup{30}{+.450} & \scoreup{26}{+.325} & \scoreup{14}{+.100} & \scoreup{20}{+.200} & +.250 & -.154 \\
     & Qwen 3.5 F & \scoreup{30}{+.450} & \scoredown{13}{-.050} & \scoreup{32}{+.500} & \scoredown{13}{-.050} & \scoreup{25}{+.300} & \scoredown{16}{-.075} & \scoreup{30}{+.450} & \scoreup{17}{+.150} & \scoreup{38}{+.700} & \scoreup{27}{+.350} & \scoreup{22}{+.250} & \scorezero{0.000} & +.442 & +.054 \\
     & GPT-5 mini & \scoreup{17}{+.150} & \scoredown{16}{-.075} & \scoreup{29}{+.425} & \scoredown{23}{-.150} & \scoreup{14}{+.100} & \scoreup{7}{+.025} & \scoredown{9}{-.025} & \scoreup{12}{+.075} & \scoreup{10}{+.050} & \scoreup{21}{+.225} & \scoreup{7}{+.025} & \scoredown{13}{-.050} & +.121 & +.008 \\
     & GPT-5.5 & \scoredown{9}{-.025} & \scoredown{19}{-.100} & \scoreup{19}{+.175} & \scoredown{23}{-.150} & \scoreup{16}{+.125} & \scoredown{28}{-.225} & \scoredown{21}{-.125} & \scorezero{0.000} & \scoreup{25}{+.300} & \scorezero{0.000} & \scorezero{0.000} & \scoredown{9}{-.025} & +.075 & -.083 \\
     & Sonnet 5 & \scoredown{19}{-.100} & \scoredown{43}{-.525} & \scoredown{37}{-.375} & \scoredown{41}{-.475} & \scoredown{19}{-.100} & \scoredown{44}{-.538} & \scoredown{21}{-.128} & \scoredown{35}{-.350} & \scoredown{38}{-.400} & \scoredown{42}{-.500} & \scoreup{14}{+.100} & \scoredown{34}{-.325} & -.167 & -.452 \\
    \cmidrule(lr){1-16}
    Opus 4.8 & Gemini 3.5 FL & \scoreup{25}{+.300} & \scoredown{56}{-.875} & \scoreup{35}{+.600} & \scoredown{41}{-.475} & \scoreup{29}{+.425} & \scoredown{46}{-.600} & \scoreup{25}{+.300} & \scorezero{0.000} & \scoreup{35}{+.600} & \scoreup{17}{+.150} & \scoreup{19}{+.175} & \scoreup{7}{+.025} & +.400 & -.296 \\
     & Qwen 3.5 F & \scoreup{25}{+.300} & \scoredown{44}{-.550} & \scoreup{33}{+.550} & \scoredown{23}{-.150} & \scoreup{36}{+.650} & \scoredown{28}{-.225} & \scoreup{30}{+.450} & \scoreup{10}{+.050} & \scoreup{36}{+.650} & \scoreup{21}{+.225} & \scoreup{21}{+.225} & \scorezero{0.000} & +.471 & -.108 \\
     & GPT-5 mini & \scoreup{16}{+.125} & \scoredown{42}{-.500} & \scoreup{33}{+.525} & \scoredown{27}{-.200} & \scoreup{21}{+.225} & \scoredown{34}{-.325} & \scoreup{19}{+.175} & \scoreup{7}{+.025} & \scoreup{28}{+.400} & \scorezero{0.000} & \scoreup{7}{+.025} & \scoreup{14}{+.100} & +.246 & -.150 \\
     & GPT-5.5 & \scoreup{16}{+.125} & \scoredown{28}{-.225} & \scoreup{10}{+.050} & \scoredown{21}{-.125} & \scoreup{16}{+.125} & \scoredown{16}{-.075} & \scoreup{22}{+.250} & \scoredown{13}{-.050} & \scoreup{20}{+.200} & \scoredown{13}{-.050} & \scoreup{7}{+.025} & \scoredown{16}{-.075} & +.129 & -.100 \\
     & Sonnet 5 & \scoredown{32}{-.282} & \scoredown{49}{-.675} & \scorezero{0.000} & \scoredown{42}{-.500} & \scoreup{13}{+.079} & \scoredown{33}{-.308} & \scoredown{23}{-.150} & \scoredown{33}{-.300} & \scoredown{28}{-.225} & \scoredown{23}{-.150} & \scoredown{27}{-.200} & \scoredown{35}{-.350} & -.130 & -.380 \\
    \midrule
    \multicolumn{16}{l}{\textit{human OA [4,5]}} \\
    \cmidrule(lr){1-16}
    GPT-5.5 & Gemini 3.5 FL & \scorezero{0.000} & \scoredown{33}{-.300} & \scoredown{13}{-.050} & \scoredown{23}{-.150} & \scoredown{13}{-.050} & \scoredown{23}{-.150} & \scoreup{14}{+.100} & \scoredown{19}{-.100} & \scoreup{10}{+.050} & \scoreup{10}{+.050} & \scorezero{0.000} & \scoredown{19}{-.100} & +.008 & -.125 \\
     & Qwen 3.5 F & \scoreup{20}{+.200} & \scoredown{35}{-.350} & \scoreup{37}{+.675} & \scoredown{34}{-.325} & \scoreup{33}{+.550} & \scoredown{44}{-.550} & \scoreup{7}{+.025} & \scoreup{24}{+.275} & \scoreup{26}{+.325} & \scoreup{16}{+.125} & \scoreup{27}{+.350} & \scoredown{13}{-.050} & +.354 & -.146 \\
     & GPT-5 mini & \scoreup{24}{+.275} & \scoredown{34}{-.325} & \scoreup{7}{+.025} & \scoredown{19}{-.100} & \scoreup{20}{+.200} & \scoredown{19}{-.100} & \scoreup{25}{+.300} & \scoreup{21}{+.225} & \scoreup{19}{+.175} & \scoreup{17}{+.150} & \scoredown{13}{-.050} & \scoreup{7}{+.025} & +.154 & -.021 \\
     & GPT-5.5 & \scoreup{14}{+.100} & \scoreup{7}{+.025} & \scoreup{26}{+.325} & \scoredown{16}{-.075} & \scoreup{10}{+.050} & \scoreup{12}{+.075} & \scoreup{7}{+.025} & \scoreup{17}{+.150} & \scoreup{7}{+.025} & \scoreup{22}{+.250} & \scoreup{20}{+.200} & \scoreup{19}{+.175} & +.121 & +.100 \\
     & Sonnet 5 & \scoredown{16}{-.075} & \scoredown{46}{-.590} & \scoredown{33}{-.308} & \scoredown{42}{-.500} & \scoredown{31}{-.275} & \scoredown{35}{-.350} & \scoredown{19}{-.100} & \scoredown{27}{-.200} & \scoredown{38}{-.410} & \scoredown{27}{-.205} & \scoredown{17}{-.077} & \scoredown{30}{-.250} & -.207 & -.349 \\
    \cmidrule(lr){1-16}
    Opus 4.8 & Gemini 3.5 FL & \scorezero{0.000} & \scoredown{38}{-.400} & \scoreup{14}{+.100} & \scoredown{35}{-.350} & \scoreup{14}{+.100} & \scoredown{38}{-.400} & \scoreup{10}{+.050} & \scorezero{0.000} & \scoreup{14}{+.100} & \scoreup{10}{+.050} & \scoreup{10}{+.050} & \scorezero{0.000} & +.067 & -.183 \\
     & Qwen 3.5 F & \scoreup{26}{+.325} & \scoredown{37}{-.375} & \scoreup{35}{+.600} & \scoredown{40}{-.450} & \scoreup{34}{+.575} & \scoredown{41}{-.475} & \scoreup{29}{+.425} & \scoreup{19}{+.175} & \scoreup{37}{+.675} & \scoredown{23}{-.150} & \scoredown{28}{-.225} & \scoreup{21}{+.225} & +.396 & -.175 \\
     & GPT-5 mini & \scoreup{28}{+.375} & \scoredown{37}{-.375} & \scoreup{41}{+.850} & \scoredown{33}{-.300} & \scoreup{14}{+.100} & \scoredown{34}{-.325} & \scoreup{22}{+.250} & \scoreup{16}{+.125} & \scoreup{27}{+.350} & \scoreup{29}{+.425} & \scoredown{9}{-.025} & \scoreup{24}{+.275} & +.317 & -.029 \\
     & GPT-5.5 & \scoreup{16}{+.125} & \scoredown{25}{-.175} & \scoreup{30}{+.450} & \scoredown{23}{-.150} & \scoreup{26}{+.325} & \scoredown{13}{-.050} & \scoreup{20}{+.200} & \scoreup{16}{+.125} & \scoreup{25}{+.300} & \scoreup{21}{+.225} & \scoreup{14}{+.100} & \scoreup{20}{+.200} & +.250 & +.029 \\
     & Sonnet 5 & \scoreup{10}{+.050} & \scoredown{43}{-.513} & \scoredown{14}{-.051} & \scoredown{44}{-.526} & \scorezero{0.000} & \scoredown{30}{-.256} & \scoreup{19}{+.179} & \scoreup{10}{+.051} & \scoredown{10}{-.026} & \scoredown{29}{-.237} & \scoredown{10}{-.026} & \scoredown{24}{-.158} & +.021 & -.273 \\
    \midrule
    \multicolumn{16}{l}{\textit{human OA [6,7]}} \\
    \cmidrule(lr){1-16}
    GPT-5.5 & Gemini 3.5 FL & \scorezero{0.000} & \scorezero{0.000} & \scoredown{14}{-.054} & \scoredown{20}{-.108} & \scorezero{0.000} & \scoredown{20}{-.108} & \scorezero{0.000} & \scorezero{0.000} & \scorezero{0.000} & \scorezero{0.000} & \scorezero{0.000} & \scorezero{0.000} & -.009 & -.036 \\
     & Qwen 3.5 F & \scorezero{0.000} & \scoredown{47}{-.622} & \scoredown{24}{-.162} & \scoreup{13}{+.081} & \scoreup{22}{+.243} & \scoredown{36}{-.351} & \scoreup{13}{+.081} & \scoreup{7}{+.027} & \scoredown{17}{-.081} & \scoreup{13}{+.081} & \scoredown{22}{-.135} & \scoreup{13}{+.081} & -.009 & -.117 \\
     & GPT-5 mini & \scoreup{30}{+.432} & \scoredown{55}{-.838} & \scoreup{15}{+.108} & \scoredown{34}{-.324} & \scoreup{17}{+.135} & \scoredown{43}{-.514} & \scoreup{15}{+.108} & \scoreup{17}{+.135} & \scoreup{21}{+.216} & \scoredown{10}{-.027} & \scoreup{22}{+.243} & \scoreup{10}{+.054} & +.207 & -.252 \\
     & GPT-5.5 & \scoreup{10}{+.054} & \scoreup{13}{+.081} & \scoredown{22}{-.135} & \scoredown{20}{-.108} & \scoreup{18}{+.162} & \scoredown{17}{-.081} & \scoreup{13}{+.081} & \scoredown{17}{-.081} & \scoreup{10}{+.054} & \scoreup{18}{+.162} & \scoreup{7}{+.027} & \scorezero{0.000} & +.041 & -.005 \\
     & Sonnet 5 & \scoredown{30}{-.243} & \scoredown{28}{-.216} & \scoredown{31}{-.270} & \scoredown{34}{-.324} & \scoredown{26}{-.189} & \scoredown{41}{-.459} & \scoredown{24}{-.167} & \scoredown{17}{-.081} & \scoredown{37}{-.378} & \scoredown{28}{-.216} & \scoredown{20}{-.108} & \scoredown{24}{-.162} & -.226 & -.243 \\
    \cmidrule(lr){1-16}
    Opus 4.8 & Gemini 3.5 FL & \scorezero{0.000} & \scoredown{24}{-.162} & \scorezero{0.000} & \scoredown{28}{-.216} & \scoredown{14}{-.054} & \scoredown{24}{-.162} & \scorezero{0.000} & \scorezero{0.000} & \scorezero{0.000} & \scorezero{0.000} & \scorezero{0.000} & \scorezero{0.000} & -.009 & -.090 \\
     & Qwen 3.5 F & \scoreup{17}{+.135} & \scoredown{45}{-.568} & \scoreup{29}{+.405} & \scoredown{33}{-.297} & \scoreup{22}{+.243} & \scoredown{34}{-.324} & \scoreup{22}{+.243} & \scoredown{10}{-.027} & \scoreup{7}{+.027} & \scoredown{22}{-.135} & \scoreup{7}{+.027} & \scoredown{17}{-.081} & +.180 & -.239 \\
     & GPT-5 mini & \scoreup{35}{+.622} & \scoredown{36}{-.351} & \scoreup{35}{+.622} & \scoredown{50}{-.703} & \scoreup{26}{+.324} & \scoredown{43}{-.514} & \scoreup{26}{+.324} & \scoreup{15}{+.108} & \scoreup{34}{+.568} & \scoreup{15}{+.108} & \scoreup{10}{+.054} & \scoreup{20}{+.189} & +.419 & -.194 \\
     & GPT-5.5 & \scoredown{14}{-.054} & \scoredown{22}{-.135} & \scoreup{15}{+.108} & \scoredown{17}{-.081} & \scoreup{7}{+.027} & \scoreup{10}{+.054} & \scoreup{26}{+.324} & \scoreup{23}{+.270} & \scoreup{18}{+.162} & \scoreup{10}{+.054} & \scoredown{26}{-.189} & \scoreup{10}{+.054} & +.063 & +.036 \\
     & Sonnet 5 & \scoreup{15}{+.111} & \scoredown{34}{-.324} & \scoredown{17}{-.081} & \scoredown{33}{-.297} & \scoredown{30}{-.243} & \scoredown{30}{-.243} & \scoreup{7}{+.027} & \scoredown{22}{-.135} & \scoredown{26}{-.189} & \scoredown{22}{-.135} & \scoreup{7}{+.027} & \scoredown{17}{-.081} & -.058 & -.203 \\
    \midrule
    \multicolumn{16}{l}{\textit{human OA [8,10]\textsuperscript{\(\dagger\)}}} \\
    \cmidrule(lr){1-16}
    GPT-5.5 & Gemini 3.5 FL & \scorezero{0.000} & \scorezero{0.000} & \scorezero{0.000} & \scorezero{0.000} & \scorezero{0.000} & \scorezero{0.000} & \scorezero{0.000} & \scorezero{0.000} & \scorezero{0.000} & \scorezero{0.000} & \scorezero{0.000} & \scorezero{0.000} & \scorezero{0.000} & \scorezero{0.000} \\
     & Qwen 3.5 F & \scoreup{37}{+.667} & \scorezero{0.000} & \scoreup{37}{+.667} & \scorezero{0.000} & \scoreup{37}{+.667} & \scoreup{37}{+.667} & \scoreup{37}{+.667} & \scorezero{0.000} & \scoreup{37}{+.667} & \scoreup{37}{+.667} & \scorezero{0.000} & \scoreup{37}{+.667} & +.556 & +.333 \\
     & GPT-5 mini & \scoreup{37}{+.667} & \scoreup{37}{+.667} & \scorezero{0.000} & \scorezero{0.000} & \scoreup{37}{+.667} & \scoreup{37}{+.667} & \scoreup{37}{+.667} & \scorezero{0.000} & \scoreup{37}{+.667} & \scoreup{37}{+.667} & \scorezero{0.000} & \scorezero{0.000} & +.444 & +.333 \\
     & GPT-5.5 & \scoreup{37}{+.667} & \scorezero{0.000} & \scorezero{0.000} & \scorezero{0.000} & \scorezero{0.000} & \scorezero{0.000} & \scoreup{37}{+.667} & \scoreup{37}{+.667} & \scoreup{37}{+.667} & \scoreup{37}{+.667} & \scoreup{45}{+1.333} & \scoreup{37}{+.667} & +.556 & +.333 \\
     & Sonnet 5 & \scoredown{49}{-.667} & \scoredown{60}{-1.000} & \scorezero{0.000} & \scoredown{49}{-.667} & \scorezero{0.000} & \scoredown{49}{-.667} & \scorezero{0.000} & \scoredown{49}{-.667} & \scorezero{0.000} & \scorezero{0.000} & \scorezero{0.000} & \scoredown{49}{-.667} & -.111 & -.611 \\
    \cmidrule(lr){1-16}
    Opus 4.8 & Gemini 3.5 FL & \scorezero{0.000} & \scorezero{0.000} & \scorezero{0.000} & \scorezero{0.000} & \scorezero{0.000} & \scorezero{0.000} & \scorezero{0.000} & \scorezero{0.000} & \scorezero{0.000} & \scorezero{0.000} & \scorezero{0.000} & \scorezero{0.000} & \scorezero{0.000} & \scorezero{0.000} \\
     & Qwen 3.5 F & \scoreup{37}{+.667} & \scoreup{37}{+.667} & \scoreup{37}{+.667} & \scoreup{37}{+.667} & \scorezero{0.000} & \scoreup{37}{+.667} & \scoreup{37}{+.667} & \scoreup{37}{+.667} & \scoredown{35}{-.333} & \scoreup{37}{+.667} & \scoreup{37}{+.667} & \scoreup{37}{+.667} & +.389 & +.667 \\
     & GPT-5 mini & \scoreup{37}{+.667} & \scoreup{37}{+.667} & \scoreup{37}{+.667} & \scoreup{37}{+.667} & \scoreup{37}{+.667} & \scorezero{0.000} & \scorezero{0.000} & \scoreup{37}{+.667} & \scoreup{37}{+.667} & \scoreup{37}{+.667} & \scoreup{37}{+.667} & \scoreup{37}{+.667} & +.556 & +.556 \\
     & GPT-5.5 & \scoreup{45}{+1.333} & \scoreup{37}{+.667} & \scoreup{37}{+.667} & \scorezero{0.000} & \scoreup{37}{+.667} & \scorezero{0.000} & \scoreup{37}{+.667} & \scoreup{37}{+.667} & \scoreup{37}{+.667} & \scoreup{37}{+.667} & \scoreup{37}{+.667} & \scoreup{37}{+.667} & +.778 & +.444 \\
     & Sonnet 5 & \scoredown{49}{-.667} & \scorezero{0.000} & \scorezero{0.000} & \scoredown{49}{-.667} & \scorezero{0.000} & \scoredown{49}{-.667} & \scorezero{0.000} & \scoredown{49}{-.667} & \scoredown{35}{-.333} & \scoreup{37}{+.667} & \scoreup{37}{+.667} & \scorezero{0.000} & -.056 & -.222 \\
    \bottomrule
  \end{tabular}
  \end{adjustbox}
\end{table*}

\subsection{AI Review Design and Outcomes}
\label{app:ai_review_design}

Each anonymized original and rewritten source is compiled into a PDF and independently evaluated by \emph{Gemini 3.5 FL}~\citep{google_gemini35flashlite}, \emph{Qwen 3.5 F}~\citep{qwen3.5flash}, \emph{GPT-5 mini}~\citep{openai_gpt5mini}, GPT-5.5~\citep{openai_gpt55}, or \emph{Sonnet 5}~\citep{anthropic_claude_sonnet5}. Reviewers are not told the rewrite condition or rewrite model. The fixed \emph{standard} prompt follows a conventional conference rubric. The \emph{strict} prompt additionally requires concrete evidence for high ratings and instructs the reviewer not to reward presentation unless it changes the scientific assessment. Both request structured reviews and numeric ratings aligned with ICLR guidelines~\citep{iclr2026reviewer}.

Overall rating is the primary outcome. Soundness, presentation, and contribution are secondary outcomes, and reviewer confidence is an auxiliary diagnostic. We define \emph{weak-accept probability} as the share of evaluations assigning an overall rating of 6 or higher. A rating of 6 denotes an acceptable submission under the review rubric and is also above the mean reviewer rating of 5.39 among accepted ICLR 2026 submissions.\footnote{\url{https://papercopilot.com/statistics/iclr-statistics/iclr-2026-statistics/}} This threshold therefore captures whether an evaluation falls on the weak-accept side of the rating scale, rather than predicting an actual conference decision.

\begin{table}[t]
  \centering
  \caption{\textbf{Rhetorical effects across original AI-review score ranges under
standard evaluation.}
Entries report mean changes in the overall assessment relative to the
prompt-matched original, stratified
by the reviewer’s OA score for the unrevised manuscript. The strict-evaluation counterpart is reported in
Appendix~\ref{app:strict_ai_score_ranges}. \(\dagger\) marks ranges with fewer than five papers. ``-'' indicates that no papers are available for the corresponding cell.}
  \label{tab:score_boundary_directional_separation}
  \scriptsize
  \setlength{\tabcolsep}{1.3pt}
  \renewcommand{\arraystretch}{0.78}
  \newcommand{\scoreup}[2]{\cellcolor{green!#1}#2}
  \newcommand{\scoredown}[2]{\cellcolor{yellow!#1}#2}
  \newcommand{\scorezero}[1]{#1}
  \begin{adjustbox}{width=0.82\textwidth}
  \begin{tabular}{@{}ll*{14}{r}@{}}
    \toprule
    \textbf{Rewriter} & \textbf{Reviewer}
      & \multicolumn{2}{c}{\textbf{Novelty}}
      & \multicolumn{2}{c}{\textbf{Evidence}}
      & \multicolumn{2}{c}{\textbf{Scope}}
      & \multicolumn{2}{c}{\textbf{Technical}}
      & \multicolumn{2}{c}{\textbf{Contribution}}
      & \multicolumn{2}{c}{\textbf{Lexical}}
      & \multicolumn{2}{c}{\shortstack{\textbf{Across}\\\textbf{dimensions}}} \\
    \cmidrule(lr){3-4}\cmidrule(lr){5-6}\cmidrule(lr){7-8}
    \cmidrule(lr){9-10}\cmidrule(lr){11-12}\cmidrule(lr){13-14}
    \cmidrule(l){15-16}
      & & \multicolumn{1}{c}{\textbf{Pos.}} & \multicolumn{1}{c}{\textbf{Neg.}}
      & \multicolumn{1}{c}{\textbf{Pos.}} & \multicolumn{1}{c}{\textbf{Neg.}}
      & \multicolumn{1}{c}{\textbf{Pos.}} & \multicolumn{1}{c}{\textbf{Neg.}}
      & \multicolumn{1}{c}{\textbf{Pos.}} & \multicolumn{1}{c}{\textbf{Neg.}}
      & \multicolumn{1}{c}{\textbf{Pos.}} & \multicolumn{1}{c}{\textbf{Neg.}}
      & \multicolumn{1}{c}{\textbf{Pos.}} & \multicolumn{1}{c}{\textbf{Neg.}}
      & \multicolumn{1}{c}{\textbf{Pos.}} & \multicolumn{1}{c}{\textbf{Neg.}} \\
    \midrule
    \multicolumn{16}{l}{\textit{AI Original-score [1,3]}} \\
    \cmidrule(lr){1-16}
    GPT-5.5 & Gemini 3.5 FL & \multicolumn{1}{c}{--} & \multicolumn{1}{c}{--} & \multicolumn{1}{c}{--} & \multicolumn{1}{c}{--} & \multicolumn{1}{c}{--} & \multicolumn{1}{c}{--} & \multicolumn{1}{c}{--} & \multicolumn{1}{c}{--} & \multicolumn{1}{c}{--} & \multicolumn{1}{c}{--} & \multicolumn{1}{c}{--} & \multicolumn{1}{c}{--} & \multicolumn{1}{c}{--} & \multicolumn{1}{c}{--} \\
     & Qwen 3.5 F & \scoreup{45}{+2.000} & \scoreup{45}{+4.500} & \scoreup{45}{+4.500} & \scoreup{45}{+4.000} & \scoreup{45}{+4.500} & \scoreup{45}{+3.500} & \scoreup{45}{+4.500} & \scoreup{45}{+3.500} & \scoreup{45}{+6.000} & \scoreup{45}{+3.500} & \scoreup{45}{+4.000} & \scoreup{45}{+3.000} & +4.250 & +3.667 \\
     & GPT-5 mini & \scorezero{0.000} & \scoreup{45}{+2.500} & \scoreup{45}{+3.000} & \scoreup{45}{+1.500} & \scoreup{45}{+2.000} & \scoreup{45}{+2.500} & \scoreup{45}{+1.000} & \scoreup{45}{+2.000} & \scoreup{45}{+1.500} & \scoreup{45}{+2.500} & \scoreup{45}{+1.000} & \scoreup{45}{+1.000} & +1.417 & +2.000 \\
     & GPT-5.5 & \scoreup{34}{+.583} & \scoreup{32}{+.500} & \scoreup{37}{+.667} & \scoreup{32}{+.500} & \scoreup{39}{+.750} & \scoreup{29}{+.417} & \scoreup{37}{+.667} & \scoreup{37}{+.667} & \scoreup{41}{+.833} & \scoreup{34}{+.583} & \scoreup{39}{+.750} & \scoreup{41}{+.833} & +.708 & +.583 \\
     & Sonnet 5 & \scoreup{32}{+.500} & \scoreup{16}{+.125} & \scoreup{32}{+.500} & \scoreup{28}{+.375} & \scoreup{28}{+.375} & \scoreup{28}{+.375} & \scoreup{36}{+.625} & \scoreup{22}{+.250} & \scoreup{22}{+.250} & \scoreup{28}{+.375} & \scoreup{32}{+.500} & \scoreup{28}{+.375} & +.458 & +.312 \\
    \cmidrule(lr){1-16}
    Opus 4.8 & Gemini 3.5 FL & \multicolumn{1}{c}{--} & \multicolumn{1}{c}{--} & \multicolumn{1}{c}{--} & \multicolumn{1}{c}{--} & \multicolumn{1}{c}{--} & \multicolumn{1}{c}{--} & \multicolumn{1}{c}{--} & \multicolumn{1}{c}{--} & \multicolumn{1}{c}{--} & \multicolumn{1}{c}{--} & \multicolumn{1}{c}{--} & \multicolumn{1}{c}{--} & \multicolumn{1}{c}{--} & \multicolumn{1}{c}{--} \\
     & Qwen 3.5 F & \scoreup{45}{+3.500} & \scoreup{45}{+5.000} & \scoreup{45}{+5.000} & \scoreup{45}{+4.500} & \scoreup{45}{+5.000} & \scoreup{45}{+3.500} & \scoreup{45}{+4.000} & \scoreup{45}{+4.000} & \scoreup{45}{+5.000} & \scoreup{45}{+6.000} & \scoreup{45}{+2.000} & \scoreup{45}{+4.000} & +4.083 & +4.500 \\
     & GPT-5 mini & \scoreup{45}{+2.500} & \scoreup{45}{+1.000} & \scoreup{45}{+2.500} & \scoreup{45}{+2.000} & \scoreup{45}{+2.500} & \scoreup{45}{+2.500} & \scoreup{45}{+2.500} & \scoreup{45}{+1.000} & \scoreup{45}{+3.000} & \scoreup{45}{+1.000} & \scoreup{45}{+2.500} & \scoreup{45}{+1.500} & +2.583 & +1.500 \\
     & GPT-5.5 & \scoreup{32}{+.500} & \scoreup{29}{+.417} & \scoreup{37}{+.667} & \scoreup{37}{+.667} & \scoreup{41}{+.833} & \scoreup{41}{+.833} & \scoreup{45}{+1.250} & \scoreup{37}{+.667} & \scoreup{37}{+.667} & \scoreup{29}{+.417} & \scoreup{32}{+.500} & \scoreup{37}{+.667} & +.736 & +.611 \\
     & Sonnet 5 & \scoreup{28}{+.375} & \scoreup{28}{+.375} & \scoreup{36}{+.625} & \scoreup{22}{+.250} & \scoreup{37}{+.667} & \scoreup{36}{+.625} & \scoreup{32}{+.500} & \scoreup{22}{+.250} & \scoreup{22}{+.250} & \scoreup{30}{+.438} & \scoreup{32}{+.500} & \scoreup{22}{+.250} & +.486 & +.365 \\
    \midrule
    \multicolumn{16}{l}{\textit{AI Original-score [4,5]}} \\
    \cmidrule(lr){1-16}
    GPT-5.5 & Gemini 3.5 FL & \scoreup{45}{+1.000} & \scoredown{42}{-.500} & \scoredown{42}{-.500} & \scoredown{42}{-.500} & \scoreup{32}{+.500} & \scoredown{42}{-.500} & \scoreup{45}{+1.000} & \scoreup{45}{+1.000} & \scoreup{45}{+1.000} & \scoreup{45}{+1.000} & \scoredown{42}{-.500} & \scoreup{45}{+2.000} & +.417 & +.417 \\
     & Qwen 3.5 F & \scoreup{43}{+.909} & \scoreup{30}{+.455} & \scoreup{45}{+1.091} & \scoreup{36}{+.636} & \scoreup{45}{+1.000} & \scoreup{36}{+.636} & \scoreup{45}{+1.091} & \scoreup{43}{+.909} & \scoreup{45}{+1.545} & \scoreup{45}{+1.182} & \scoreup{30}{+.455} & \scoreup{41}{+.818} & +1.015 & +.773 \\
     & GPT-5 mini & \scoreup{45}{+1.000} & \scoreup{14}{+.091} & \scoreup{43}{+.909} & \scoreup{27}{+.364} & \scoreup{30}{+.455} & \scoreup{36}{+.636} & \scoreup{38}{+.727} & \scoreup{33}{+.545} & \scoreup{33}{+.545} & \scoreup{43}{+.909} & \scoreup{27}{+.364} & \scoreup{27}{+.364} & +.667 & +.485 \\
     & GPT-5.5 & \scoreup{17}{+.146} & \scoreup{13}{+.083} & \scoreup{25}{+.312} & \scoredown{9}{-.021} & \scoreup{19}{+.188} & \scoreup{9}{+.042} & \scoreup{11}{+.062} & \scoreup{17}{+.146} & \scoreup{25}{+.312} & \scoreup{24}{+.292} & \scoreup{18}{+.167} & \scoreup{16}{+.125} & +.198 & +.111 \\
     & Sonnet 5 & \scoredown{25}{-.173} & \scoredown{47}{-.615} & \scoredown{39}{-.431} & \scoredown{44}{-.538} & \scoredown{30}{-.250} & \scoredown{43}{-.510} & \scoredown{24}{-.157} & \scoredown{34}{-.327} & \scoredown{39}{-.431} & \scoredown{36}{-.367} & \scoreup{9}{+.039} & \scoredown{36}{-.365} & -.234 & -.454 \\
    \cmidrule(lr){1-16}
    Opus 4.8 & Gemini 3.5 FL & \scoreup{45}{+1.000} & \scoredown{60}{-1.000} & \scoreup{45}{+2.000} & \scoreup{32}{+.500} & \scoreup{45}{+1.000} & \scoreup{45}{+1.000} & \scoredown{60}{-2.000} & \scoreup{45}{+1.000} & \scoreup{45}{+2.000} & \scoreup{32}{+.500} & \scoreup{32}{+.500} & \scoredown{60}{-2.000} & +.750 & \scorezero{0.000} \\
     & Qwen 3.5 F & \scoreup{45}{+1.182} & \scoredown{26}{-.182} & \scoreup{45}{+1.091} & \scoreup{30}{+.455} & \scoreup{45}{+1.182} & \scoreup{36}{+.636} & \scoreup{45}{+1.273} & \scoreup{36}{+.636} & \scoreup{45}{+1.000} & \scoreup{45}{+1.000} & \scoreup{45}{+1.273} & \scoreup{24}{+.273} & +1.167 & +.470 \\
     & GPT-5 mini & \scoreup{33}{+.545} & \scoreup{19}{+.182} & \scoreup{45}{+1.000} & \scoreup{19}{+.182} & \scoreup{38}{+.727} & \scoreup{14}{+.091} & \scoreup{33}{+.545} & \scoreup{27}{+.364} & \scoreup{36}{+.636} & \scoreup{36}{+.636} & \scoreup{19}{+.182} & \scoreup{36}{+.636} & +.606 & +.348 \\
     & GPT-5.5 & \scoreup{21}{+.208} & \scoredown{15}{-.062} & \scoreup{27}{+.354} & \scoredown{15}{-.062} & \scoreup{22}{+.229} & \scoreup{11}{+.062} & \scoreup{28}{+.375} & \scoreup{22}{+.250} & \scoreup{22}{+.229} & \scoreup{24}{+.292} & \scoreup{15}{+.104} & \scoreup{18}{+.167} & +.250 & +.108 \\
     & Sonnet 5 & \scoredown{22}{-.137} & \scoredown{52}{-.765} & \scoredown{21}{-.118} & \scoredown{46}{-.600} & \scoredown{17}{-.080} & \scoredown{35}{-.340} & \scoreup{9}{+.039} & \scoredown{27}{-.196} & \scoredown{25}{-.173} & \scoredown{32}{-.288} & \scoredown{27}{-.196} & \scoredown{34}{-.320} & -.111 & -.418 \\
    \midrule
    \multicolumn{16}{l}{\textit{AI Original-score [6,7]}} \\
    \cmidrule(lr){1-16}
    GPT-5.5 & Gemini 3.5 FL & \scoreup{45}{+1.143} & \scoreup{12}{+.071} & \scoreup{45}{+1.214} & \scoreup{27}{+.357} & \scoreup{45}{+1.143} & \scoreup{12}{+.071} & \scoreup{45}{+1.214} & \scoreup{45}{+1.000} & \scoreup{45}{+1.429} & \scoreup{45}{+1.214} & \scoreup{43}{+.929} & \scoreup{42}{+.857} & +1.179 & +.595 \\
     & Qwen 3.5 F & \scoreup{41}{+.844} & \scoreup{22}{+.244} & \scoreup{45}{+1.022} & \scoreup{29}{+.422} & \scoreup{45}{+1.111} & \scoreup{28}{+.400} & \scoreup{43}{+.911} & \scoreup{34}{+.578} & \scoreup{39}{+.733} & \scoreup{43}{+.911} & \scoreup{39}{+.733} & \scoreup{38}{+.711} & +.893 & +.544 \\
     & GPT-5 mini & \scoreup{33}{+.543} & \scoredown{13}{-.049} & \scoreup{25}{+.321} & \scoreup{11}{+.062} & \scoreup{27}{+.370} & \scoreup{12}{+.074} & \scoreup{25}{+.309} & \scoreup{27}{+.358} & \scoreup{29}{+.407} & \scoreup{25}{+.309} & \scoreup{22}{+.247} & \scoreup{24}{+.284} & +.366 & +.173 \\
     & GPT-5.5 & \scoredown{8}{-.018} & \scoredown{23}{-.145} & \scorezero{0.000} & \scoredown{28}{-.218} & \scoredown{8}{-.018} & \scoredown{26}{-.182} & \scoredown{16}{-.073} & \scoredown{16}{-.073} & \scoredown{16}{-.073} & \scoreup{6}{+.018} & \scoredown{11}{-.036} & \scoredown{16}{-.073} & -.036 & -.112 \\
     & Sonnet 5 & \scoredown{31}{-.260} & \scoredown{39}{-.429} & \scoredown{39}{-.420} & \scoredown{46}{-.580} & \scoredown{31}{-.260} & \scoredown{46}{-.600} & \scoredown{33}{-.306} & \scoredown{27}{-.200} & \scoredown{45}{-.560} & \scoredown{39}{-.420} & \scoredown{31}{-.260} & \scoredown{32}{-.280} & -.344 & -.418 \\
    \cmidrule(lr){1-16}
    Opus 4.8 & Gemini 3.5 FL & \scoreup{45}{+1.143} & \scoredown{32}{-.286} & \scoreup{45}{+1.714} & \scoreup{17}{+.143} & \scoreup{45}{+1.500} & \scoredown{32}{-.286} & \scoreup{45}{+1.571} & \scoreup{24}{+.286} & \scoreup{45}{+1.714} & \scoreup{43}{+.929} & \scoreup{38}{+.714} & \scoreup{40}{+.786} & +1.393 & +.262 \\
     & Qwen 3.5 F & \scoreup{42}{+.867} & \scoreup{16}{+.133} & \scoreup{45}{+1.133} & \scoreup{27}{+.356} & \scoreup{45}{+1.156} & \scoreup{19}{+.178} & \scoreup{45}{+1.022} & \scoreup{38}{+.711} & \scoreup{45}{+1.111} & \scoreup{29}{+.422} & \scoreup{34}{+.556} & \scoreup{36}{+.644} & +.974 & +.407 \\
     & GPT-5 mini & \scoreup{35}{+.593} & \scoredown{26}{-.185} & \scoreup{43}{+.914} & \scoredown{20}{-.111} & \scoreup{28}{+.395} & \scoredown{23}{-.148} & \scoreup{32}{+.494} & \scoreup{22}{+.247} & \scoreup{35}{+.617} & \scoreup{25}{+.321} & \scoreup{21}{+.210} & \scoreup{28}{+.395} & +.537 & +.086 \\
     & GPT-5.5 & \scoredown{11}{-.036} & \scoredown{29}{-.236} & \scoreup{16}{+.127} & \scoredown{29}{-.236} & \scoreup{9}{+.036} & \scoredown{26}{-.182} & \scoreup{11}{+.055} & \scoredown{8}{-.018} & \scoreup{18}{+.164} & \scoredown{11}{-.036} & \scoredown{21}{-.127} & \scoredown{8}{-.018} & +.036 & -.121 \\
     & Sonnet 5 & \scoredown{12}{-.041} & \scoredown{40}{-.440} & \scoredown{19}{-.100} & \scoredown{41}{-.460} & \scoredown{27}{-.200} & \scoredown{40}{-.440} & \scoredown{21}{-.120} & \scoredown{25}{-.180} & \scoredown{28}{-.224} & \scoredown{27}{-.208} & \scoredown{12}{-.040} & \scoredown{28}{-.220} & -.121 & -.325 \\
    \midrule
    \multicolumn{16}{l}{\textit{AI Original-score [8,10]}} \\
    \cmidrule(lr){1-16}
    GPT-5.5 & Gemini 3.5 FL & \scoredown{12}{-.038} & \scoredown{32}{-.288} & \scoredown{19}{-.096} & \scoredown{34}{-.327} & \scoredown{19}{-.096} & \scoredown{35}{-.346} & \scoredown{19}{-.096} & \scoredown{22}{-.135} & \scoredown{8}{-.019} & \scoredown{12}{-.038} & \scoredown{17}{-.077} & \scoredown{20}{-.115} & -.071 & -.208 \\
     & Qwen 3.5 F & \scoredown{37}{-.387} & \scoredown{60}{-1.032} & \scoredown{37}{-.387} & \scoredown{52}{-.742} & \scoredown{38}{-.403} & \scoredown{60}{-1.097} & \scoredown{47}{-.613} & \scoredown{38}{-.403} & \scoredown{36}{-.355} & \scoredown{46}{-.597} & \scoredown{40}{-.435} & \scoredown{51}{-.710} & -.430 & -.763 \\
     & GPT-5 mini & \scoredown{53}{-.769} & \scoredown{60}{-1.808} & \scoredown{53}{-.769} & \scoredown{60}{-1.308} & \scoredown{53}{-.769} & \scoredown{60}{-1.462} & \scoredown{50}{-.692} & \scoredown{55}{-.846} & \scoredown{56}{-.885} & \scoredown{58}{-.923} & \scoredown{50}{-.692} & \scoredown{60}{-1.077} & -.763 & -1.237 \\
     & GPT-5.5 & \scoredown{60}{-1.200} & \scoredown{38}{-.400} & \scoredown{60}{-1.600} & \scoredown{60}{-1.200} & \scoredown{54}{-.800} & \scoredown{60}{-1.200} & \scoredown{60}{-1.200} & \scoredown{60}{-1.200} & \scoredown{54}{-.800} & \scoredown{54}{-.800} & \scoredown{38}{-.400} & \scoredown{54}{-.800} & -1.000 & -.933 \\
     & Sonnet 5 & \scoredown{60}{-2.000} & \scoredown{60}{-2.000} & \scoredown{60}{-1.000} & \scoredown{60}{-1.000} & \scoredown{60}{-1.000} & \scoredown{60}{-2.000} & \scoredown{60}{-1.000} & \scoredown{60}{-2.000} & \scorezero{0.000} & \scoredown{60}{-1.000} & \scorezero{0.000} & \scoredown{60}{-2.000} & -.833 & -1.667 \\
    \cmidrule(lr){1-16}
    Opus 4.8 & Gemini 3.5 FL & \scoredown{14}{-.058} & \scoredown{42}{-.490} & \scorezero{0.000} & \scoredown{39}{-.423} & \scoredown{12}{-.038} & \scoredown{39}{-.423} & \scoredown{12}{-.038} & \scoredown{14}{-.058} & \scorezero{0.000} & \scoredown{14}{-.058} & \scoredown{8}{-.019} & \scoredown{14}{-.058} & -.026 & -.252 \\
     & Qwen 3.5 F & \scoredown{40}{-.435} & \scoredown{60}{-1.129} & \scoredown{24}{-.161} & \scoredown{60}{-1.016} & \scoredown{31}{-.274} & \scoredown{59}{-.968} & \scoredown{36}{-.355} & \scoredown{46}{-.597} & \scoredown{32}{-.290} & \scoredown{49}{-.677} & \scoredown{48}{-.645} & \scoredown{43}{-.516} & -.360 & -.817 \\
     & GPT-5 mini & \scoredown{44}{-.538} & \scoredown{60}{-1.346} & \scoredown{37}{-.385} & \scoredown{60}{-1.577} & \scoredown{50}{-.692} & \scoredown{60}{-1.500} & \scoredown{55}{-.846} & \scoredown{44}{-.538} & \scoredown{37}{-.385} & \scoredown{41}{-.462} & \scoredown{53}{-.769} & \scoredown{50}{-.692} & -.603 & -1.019 \\
     & GPT-5.5 & \scoredown{38}{-.400} & \scoredown{60}{-1.600} & \scoredown{60}{-1.200} & \scoredown{60}{-1.200} & \scoredown{38}{-.400} & \scoredown{60}{-1.200} & \scoredown{54}{-.800} & \scoredown{54}{-.800} & \scorezero{0.000} & \scoredown{60}{-1.200} & \scoredown{54}{-.800} & \scoredown{60}{-1.200} & -.600 & -1.200 \\
     & Sonnet 5 & \scoredown{60}{-2.000} & \scoredown{60}{-2.000} & \scoredown{60}{-2.000} & \scoredown{60}{-2.000} & \scoredown{60}{-1.000} & \scoredown{60}{-2.000} & \scoredown{60}{-1.000} & \scoredown{60}{-1.000} & \scoredown{60}{-1.000} & \scorezero{0.000} & \scoredown{60}{-1.000} & \scorezero{0.000} & -1.333 & -1.167 \\
    \bottomrule
  \end{tabular}
  \end{adjustbox}
\end{table}

\subsection{Paired Analysis}
\label{app:paired_analysis}

Each rewrite is compared with the anonymized original for the same paper, reviewer model, and prompt. This paired change measures whether a score moves, in which direction, and by how much while holding the evaluation condition fixed. We also compare the positive and negative variants of the same dimension under the same rewrite model. Because both variants share an original, this contrast measures directional separation rather than movement caused by rewriting alone. The same pairing is applied to secondary scores and weak-accept probability.

All reported estimands are constructed from matched comparisons. Rewrite effects compare each rewritten manuscript with the anonymized original for the same paper, reviewer model, and evaluation protocol, while contrasts between the two rhetorical directions additionally hold the rhetorical dimension and rewrite model fixed. For pooled analyses, these matched effects are first averaged within each paper and then aggregated with equal weight across the six prespecified human-score strata. Standard and strict protocols are analyzed separately, while fixed-model analyses also hold the reviewer and rewrite models constant. We report signed mean change, mean absolute movement, and 95\% percentile intervals based on 5,000 paper-level bootstrap resamples within strata. Missing outcomes are retained as missing and are not imputed. Estimates are interpreted by their magnitude, direction, and consistency across reviewer models, rewrite models, and evaluation protocols, without applying a separate multiple-testing threshold to classify effects.

\section{Which Rhetorical Dimensions Matter Most?}
\label{sec:dimension_effects}

\begin{table*}[t]
  \centering
  \caption{\textbf{Effects of joint and reviewer-guided rewriting.}
$\Delta$OA and $\Delta_{\geq6}$ report changes from the prompt-matched original
in mean OA and weak-accept probability in percentage points, respectively.
$\Delta_{\mathrm{S+}}$ compares each endpoint with the mean of the six positive
single-dimension rewrites, while $\Delta_{\mathrm{RN}}$ compares
reviewer-guided rewriting with a two-pass rewrite without review feedback.
Additional statistics are reported in
Appendices~\ref{app:combo_composition_details} and
\ref{app:review_final_endpoints}.}
  \label{tab:joint_guided_summary}
  \label{tab:joint_rewrite_summary}
  \label{tab:review_cross_evaluator_summary}
  \scriptsize
  \setlength{\tabcolsep}{2.6pt}
  \newcommand{\sixincrease}[2]{\cellcolor{green!#1}#2}
  \newcommand{\sixdecrease}[2]{\cellcolor{yellow!#1}#2}
  \begin{adjustbox}{width=0.75\columnwidth}
  \begin{tabular}{@{}llccccccc@{}}
    \toprule
    \textbf{Prompt} & \textbf{Reviewer}
      & \multicolumn{3}{c}{\textbf{Joint Rewrite}}
      & \multicolumn{4}{c}{\textbf{Reviewer-guided}} \\
    \cmidrule(lr){3-5}\cmidrule(lr){6-9}
      & & \textbf{\boldmath$\Delta$OA}
      & \textbf{\boldmath$\Delta_{\geq6}$}
      & \textbf{\boldmath$\Delta_{\mathrm{S+}}$}
      & \textbf{\boldmath$\Delta$OA}
      & \textbf{\boldmath$\Delta_{\geq6}$}
      & \textbf{\boldmath$\Delta_{\mathrm{S+}}$}
      & \textbf{\boldmath$\Delta_{\mathrm{RN}}$} \\
    \midrule
    \multicolumn{9}{l}{\textbf{GPT-5.5 rewriter}} \\
    \cmidrule(lr){1-9}
    Standard & Gemini 3.5 FL & \sixincrease{11}{+.058} & \sixincrease{8}{+.83} & \sixdecrease{9}{-.025} & \sixdecrease{8}{-.017} & \sixincrease{12}{+1.68} & \sixdecrease{19}{-.104} & \sixdecrease{16}{-.067} \\
     & Qwen 3.5 F & \sixincrease{12}{+.075} & \sixincrease{14}{+2.50} & \sixdecrease{27}{-.201} & \sixincrease{11}{+.059} & \sixincrease{18}{+4.20} & \sixdecrease{28}{-.216} & \sixdecrease{19}{-.101} \\
     & GPT-5 mini & \sixincrease{15}{+.117} & \sixincrease{16}{+3.33} & \sixdecrease{13}{-.050} & \sixincrease{21}{+.218} & \sixincrease{20}{+5.04} & \sixincrease{10}{+.050} & \sixincrease{14}{+.092} \\
     & GPT-5.5 & \sixdecrease{5}{-.008} & \sixincrease{8}{+.83} & \sixdecrease{19}{-.100} & \sixdecrease{16}{-.067} & \sixdecrease{31}{-6.72} & \sixdecrease{24}{-.155} & \sixdecrease{10}{-.025} \\
     & Sonnet 5 & \sixdecrease{22}{-.136} & \sixdecrease{27}{-5.08} & \sixincrease{11}{+.056} & \sixdecrease{21}{-.119} & \sixdecrease{27}{-5.08} & \sixincrease{13}{+.084} & \sixdecrease{17}{-.076} \\
    \cmidrule(lr){2-9}
     & \textit{Standard mean} & +.021 & +.48 & -.064 & +.015 & -.18 & -.068 & -.035 \\
    \addlinespace
    Strict & Gemini 3.5 FL & \sixdecrease{5}{-.008} & \sixincrease{18}{+4.17} & \sixdecrease{34}{-.322} & \sixdecrease{19}{-.101} & \sixdecrease{16}{-1.68} & \sixdecrease{38}{-.406} & \sixdecrease{29}{-.235} \\
     & Qwen 3.5 F & \sixincrease{9}{+.042} & \sixincrease{18}{+4.17} & \sixincrease{5}{+.013} & \sixincrease{4}{+.008} & \sixincrease{14}{+2.52} & \sixdecrease{9}{-.021} & \sixincrease{6}{+.017} \\
     & GPT-5 mini & \sixincrease{24}{+.275} & \sixincrease{14}{+2.50} & \sixdecrease{8}{-.018} & \sixincrease{26}{+.336} & 0.00 & \sixincrease{9}{+.041} & \sixincrease{12}{+.076} \\
     & GPT-5.5 & \sixincrease{7}{+.025} & \sixdecrease{19}{-2.50} & \sixdecrease{23}{-.142} & \sixdecrease{10}{-.025} & \sixdecrease{33}{-7.56} & \sixdecrease{26}{-.192} & \sixdecrease{22}{-.134} \\
     & Sonnet 5 & \sixdecrease{20}{-.109} & \sixdecrease{11}{-.84} & \sixincrease{10}{+.046} & \sixdecrease{15}{-.059} & \sixdecrease{19}{-2.54} & \sixincrease{15}{+.107} & \sixincrease{12}{+.068} \\
    \cmidrule(lr){2-9}
     & \textit{Strict mean} & +.045 & +1.50 & -.085 & +.032 & -1.85 & -.094 & -.042 \\
    \cmidrule(lr){1-9}
    \multicolumn{2}{l}{\textit{GPT-5.5 mean}} & +.033 & +.99 & -.074 & +.023 & -1.01 & -.081 & -.039 \\
    \midrule
    \multicolumn{9}{l}{\textbf{Opus 4.8 rewriter}} \\
    \cmidrule(lr){1-9}
    Standard & Gemini 3.5 FL & \sixincrease{22}{+.233} & \sixincrease{12}{+1.67} & \sixincrease{13}{+.081} & \sixincrease{18}{+.167} & \sixincrease{12}{+1.67} & \sixincrease{5}{+.014} & \sixdecrease{16}{-.067} \\
     & Qwen 3.5 F & \sixincrease{30}{+.458} & \sixincrease{22}{+5.83} & \sixincrease{15}{+.104} & \sixincrease{33}{+.525} & \sixincrease{25}{+7.50} & \sixincrease{19}{+.171} & \sixdecrease{21}{-.118} \\
     & GPT-5 mini & \sixincrease{31}{+.467} & \sixincrease{23}{+6.67} & \sixincrease{17}{+.136} & \sixincrease{37}{+.692} & \sixincrease{27}{+9.17} & \sixincrease{27}{+.361} & \sixdecrease{13}{-.050} \\
     & GPT-5.5 & \sixincrease{14}{+.092} & \sixincrease{30}{+10.83} & \sixdecrease{16}{-.074} & \sixincrease{28}{+.375} & \sixincrease{41}{+20.83} & \sixincrease{21}{+.210} & \sixincrease{14}{+.101} \\
     & Sonnet 5 & \sixincrease{20}{+.193} & \sixincrease{25}{+7.56} & \sixincrease{22}{+.243} & \sixincrease{21}{+.222} & \sixincrease{33}{+13.68} & \sixincrease{25}{+.305} & \sixincrease{14}{+.095} \\
    \cmidrule(lr){2-9}
     & \textit{Standard mean} & +.289 & +6.51 & +.098 & +.396 & +10.57 & +.212 & -.008 \\
    \addlinespace
    Strict & Gemini 3.5 FL & \sixincrease{38}{+.700} & \sixincrease{26}{+8.33} & \sixincrease{21}{+.225} & \sixincrease{35}{+.600} & \sixincrease{26}{+8.33} & \sixincrease{16}{+.125} & \sixdecrease{32}{-.277} \\
     & Qwen 3.5 F & \sixincrease{26}{+.336} & \sixincrease{33}{+13.45} & \sixincrease{17}{+.147} & \sixincrease{31}{+.483} & \sixincrease{42}{+21.67} & \sixincrease{25}{+.301} & \sixdecrease{16}{-.067} \\
     & GPT-5 mini & \sixincrease{41}{+.833} & \sixincrease{40}{+20.00} & \sixincrease{31}{+.467} & \sixincrease{43}{+.917} & \sixincrease{43}{+23.33} & \sixincrease{33}{+.550} & \sixdecrease{22}{-.134} \\
     & GPT-5.5 & \sixincrease{23}{+.258} & \sixincrease{25}{+7.50} & \sixincrease{10}{+.053} & \sixincrease{31}{+.483} & \sixincrease{35}{+15.00} & \sixincrease{24}{+.278} & \sixincrease{11}{+.059} \\
     & Sonnet 5 & \sixincrease{19}{+.186} & \sixincrease{23}{+6.78} & \sixincrease{21}{+.220} & \sixincrease{25}{+.303} & \sixincrease{23}{+6.72} & \sixincrease{26}{+.339} & \sixincrease{13}{+.085} \\
    \cmidrule(lr){2-9}
     & \textit{Strict mean} & +.463 & +11.21 & +.222 & +.557 & +15.01 & +.319 & -.067 \\
    \cmidrule(lr){1-9}
    \multicolumn{2}{l}{\textit{Opus 4.8 mean}} & +.376 & +8.86 & +.160 & +.477 & +12.79 & +.265 & -.038 \\
    \midrule
    \multicolumn{2}{l}{\textbf{Overall mean}} & \textbf{+.204} & \textbf{+4.93} & \textbf{+.043} & \textbf{+.250} & \textbf{+5.89} & \textbf{+.092} & \textbf{-.038} \\
    \bottomrule
  \end{tabular}
  \end{adjustbox}
\end{table*}

We examine how rhetorical dimensions affect overall assessment (OA) and weak-accept probability, and whether these effects vary with human-assessed paper quality or the AI reviewer's starting score. Sensitivity concentrates in evidence framing, novelty stance, and scope framing, while the starting score chiefly conditions the direction of movement.

\subsection{Evidence Framing and Novelty Stance Most Readily Shape AI-Review Outcomes}
\label{subsec:dimension_rating_changes}

Figure~\ref{fig:dimension_model_pair_rating_changes} and Table~\ref{tab:dimension_model_pair_rating_changes} show that rhetorical effects concentrate in a few dimensions. \textbf{Evidence framing} and \textbf{novelty stance} produce the \textbf{largest and most consistent changes} in overall rating, with \textbf{scope framing} forming a weaker \textbf{second tier}. \textbf{Evidence framing produces the largest overall contrast, followed closely by novelty stance.} Positive evidence and novelty variants are favored across all displayed configurations; scope follows the same direction, but its contrast is driven more by losses under narrower framing than by gains under broader framing. The remaining dimensions have smaller or less stable effects; full estimates appear in Appendix~\ref{app:dimension_rating_details}.

The weak-accept results reinforce this hierarchy: Appendix Table~\ref{tab:dimension_threshold_summary_main} reports positive-negative contrasts of 13.0 percentage points for evidence framing, 12.0 for novelty stance, and 9.0 for scope framing, with novelty slightly stronger under standard evaluation and evidence under strict evaluation. \textbf{The central pattern is not a general reward for stronger rhetoric, but selective sensitivity to particular ways of framing scientific merit.} Novelty and scope effects are mainly penalty-driven, whereas evidence framing moves scores substantially in both directions. Secondary-score results likewise show larger novelty and evidence contrasts in contribution and soundness than in presentation, with unfavorable scope mainly lowering contribution. Thus, AI reviewers appear to treat rhetorical confidence and evidential emphasis as signals of scientific merit, potentially penalizing appropriately cautious writing even when scientific content is preserved; supporting results appear in Appendices~\ref{app:threshold_flips} and~\ref{app:secondary_scores}.

\begin{table}[t]
  \centering
  \caption{\textbf{Joint rewrite effects across AI-original and human OA ranges.}
Score ranges follow the definitions in
Tables~\ref{tab:human_score_directional_separation} and
\ref{tab:score_boundary_directional_separation}.
Cells report mean paired OA changes, and the mean rows average the available
combinations of reviewer and prompt within each range. \(\dagger\) marks cells with fewer
than five matched papers, and ``--'' indicate unavailable estimates.}
  \label{tab:joint_rewrite_score_ranges}
  \scriptsize
  \setlength{\tabcolsep}{2.6pt}
  \newcommand{\sixincrease}[2]{\cellcolor{green!#1}#2}
  \newcommand{\sixdecrease}[2]{\cellcolor{yellow!#1}#2}
  \begin{adjustbox}{max width=\columnwidth}
  \begin{tabular}{llrrrr@{\hspace{8pt}}rrrr}
    \toprule
    \textbf{Review prompt} & \textbf{Reviewer}
      & \multicolumn{4}{c}{\textbf{AI Original OA}}
      & \multicolumn{4}{c}{\textbf{Human OA}} \\
    \cmidrule(lr){3-6}\cmidrule(lr){7-10}
      & & \multicolumn{1}{c}{\textbf{[1,3]}}
      & \multicolumn{1}{c}{\textbf{[4,5]}}
      & \multicolumn{1}{c}{\textbf{[6,7]}}
      & \multicolumn{1}{c}{\textbf{[8,10]}}
      & \multicolumn{1}{c}{\textbf{[1,3]}}
      & \multicolumn{1}{c}{\textbf{[4,5]}}
      & \multicolumn{1}{c}{\textbf{[6,7]}}
      & \multicolumn{1}{c}{\textbf{[8,10]}} \\
    \midrule
    \multicolumn{10}{l}{\textbf{GPT-5.5 rewriter}} \\
    \cmidrule(lr){1-10}
    Standard & Gemini 3.5 FL & -- & \sixincrease{40}{+2.00\textsuperscript{\(\dagger\)}} & \sixincrease{27}{+0.93} & \sixdecrease{12}{-0.10} & \sixincrease{15}{+0.28} & \sixdecrease{12}{-0.10} & 0.00 & 0.00\textsuperscript{\(\dagger\)} \\
     & Qwen 3.5 F & \sixincrease{45}{+2.50\textsuperscript{\(\dagger\)}} & \sixincrease{27}{+0.91} & \sixincrease{28}{+0.96} & \sixdecrease{34}{-0.79} & \sixincrease{14}{+0.23} & \sixincrease{12}{+0.17} & \sixdecrease{19}{-0.24} & \sixincrease{23}{+0.67\textsuperscript{\(\dagger\)}} \\
     & GPT-5 mini & \sixincrease{28}{+1.00\textsuperscript{\(\dagger\)}} & \sixincrease{23}{+0.64} & \sixincrease{16}{+0.31} & \sixdecrease{33}{-0.77} & \sixincrease{9}{+0.10} & \sixincrease{9}{+0.10} & \sixincrease{9}{+0.11} & \sixincrease{23}{+0.67\textsuperscript{\(\dagger\)}} \\
     & GPT-5.5 & \sixincrease{16}{+0.33} & \sixincrease{8}{+0.08} & \sixdecrease{9}{-0.05} & \sixdecrease{42}{-1.20\textsuperscript{\(\dagger\)}} & \sixdecrease{6}{-0.03} & 0.00 & \sixdecrease{9}{-0.05} & \sixincrease{23}{+0.67\textsuperscript{\(\dagger\)}} \\
     & Sonnet 5 & \sixincrease{20}{+0.50} & \sixdecrease{18}{-0.22} & \sixdecrease{16}{-0.18} & \sixdecrease{54}{-2.00\textsuperscript{\(\dagger\)}} & \sixdecrease{16}{-0.18} & \sixdecrease{15}{-0.15} & \sixdecrease{6}{-0.03} & \sixdecrease{31}{-0.67\textsuperscript{\(\dagger\)}} \\
    \cmidrule(lr){2-10}
     & \textit{Standard mean} & +1.08 & +0.68 & +0.39 & -0.97 & +0.08 & 0.00 & -0.04 & +0.27 \\
    Strict & Gemini 3.5 FL & \sixincrease{32}{+1.27} & \sixincrease{20}{+0.50} & \sixincrease{13}{+0.20} & \sixdecrease{30}{-0.61} & \sixdecrease{6}{-0.03} & \sixincrease{4}{+0.03} & \sixincrease{5}{+0.03} & \sixdecrease{31}{-0.67\textsuperscript{\(\dagger\)}} \\
     & Qwen 3.5 F & \sixincrease{31}{+1.21} & \sixdecrease{6}{-0.03} & \sixdecrease{38}{-1.00} & \sixdecrease{56}{-2.17} & \sixdecrease{6}{-0.03} & \sixincrease{14}{+0.23} & \sixdecrease{11}{-0.08} & 0.00\textsuperscript{\(\dagger\)} \\
     & GPT-5 mini & \sixincrease{27}{+0.89} & \sixdecrease{14}{-0.13} & \sixdecrease{25}{-0.45} & N/A & \sixincrease{22}{+0.60} & \sixdecrease{10}{-0.07} & \sixincrease{14}{+0.24} & \sixincrease{28}{+1.00\textsuperscript{\(\dagger\)}} \\
     & GPT-5.5 & \sixincrease{24}{+0.71} & \sixdecrease{24}{-0.39} & \sixdecrease{20}{-0.27} & 0.00\textsuperscript{\(\dagger\)} & \sixincrease{8}{+0.07} & \sixincrease{11}{+0.15} & \sixdecrease{15}{-0.16} & 0.00\textsuperscript{\(\dagger\)} \\
     & Sonnet 5 & \sixincrease{15}{+0.29} & \sixdecrease{20}{-0.27} & \sixdecrease{25}{-0.43} & \sixdecrease{54}{-2.00\textsuperscript{\(\dagger\)}} & \sixdecrease{16}{-0.17} & \sixdecrease{12}{-0.10} & \sixdecrease{9}{-0.05} & 0.00\textsuperscript{\(\dagger\)} \\
    \cmidrule(lr){2-10}
     & \textit{Strict mean} & +0.87 & -0.06 & -0.39 & -1.19 & +0.09 & +0.04 & -0.01 & +0.07 \\
    \cmidrule(lr){1-10}
    \multicolumn{2}{l}{\textit{GPT-5.5 mean}} & +0.97 & +0.31 & 0.00 & -1.07 & +0.08 & +0.02 & -0.02 & +0.17 \\
    \midrule
    \multicolumn{10}{l}{\textbf{Opus 4.8 rewriter}} \\
    \cmidrule(lr){1-10}
    Standard & Gemini 3.5 FL & -- & \sixincrease{40}{+2.00\textsuperscript{\(\dagger\)}} & \sixincrease{37}{+1.71} & 0.00 & \sixincrease{22}{+0.60} & \sixincrease{9}{+0.10} & 0.00 & 0.00\textsuperscript{\(\dagger\)} \\
     & Qwen 3.5 F & \sixincrease{45}{+6.00\textsuperscript{\(\dagger\)}} & \sixincrease{35}{+1.55} & \sixincrease{30}{+1.11} & \sixdecrease{24}{-0.39} & \sixincrease{23}{+0.68} & \sixincrease{20}{+0.47} & \sixincrease{12}{+0.19} & \sixincrease{23}{+0.67\textsuperscript{\(\dagger\)}} \\
     & GPT-5 mini & \sixincrease{45}{+2.50\textsuperscript{\(\dagger\)}} & \sixincrease{28}{+1.00} & \sixincrease{22}{+0.59} & \sixdecrease{21}{-0.31} & \sixincrease{16}{+0.33} & \sixincrease{20}{+0.50} & \sixincrease{22}{+0.62} & 0.00\textsuperscript{\(\dagger\)} \\
     & GPT-5.5 & \sixincrease{26}{+0.83} & \sixincrease{14}{+0.25} & \sixdecrease{11}{-0.09} & \sixdecrease{42}{-1.20\textsuperscript{\(\dagger\)}} & \sixincrease{10}{+0.12} & \sixincrease{14}{+0.25} & \sixdecrease{15}{-0.16} & \sixincrease{23}{+0.67\textsuperscript{\(\dagger\)}} \\
     & Sonnet 5 & \sixincrease{22}{+0.62} & \sixincrease{13}{+0.22} & \sixincrease{6}{+0.04} & 0.00\textsuperscript{\(\dagger\)} & \sixincrease{11}{+0.15} & \sixincrease{11}{+0.15} & \sixincrease{14}{+0.24} & \sixincrease{23}{+0.67\textsuperscript{\(\dagger\)}} \\
    \cmidrule(lr){2-10}
     & \textit{Standard mean} & +2.49 & +1.00 & +0.67 & -0.38 & +0.38 & +0.30 & +0.18 & +0.40 \\
    Strict & Gemini 3.5 FL & \sixincrease{44}{+2.36} & \sixincrease{36}{+1.60} & \sixincrease{29}{+1.04} & \sixdecrease{17}{-0.20} & \sixincrease{27}{+0.90} & \sixincrease{26}{+0.85} & \sixincrease{18}{+0.38} & 0.00\textsuperscript{\(\dagger\)} \\
     & Qwen 3.5 F & \sixincrease{36}{+1.61} & \sixincrease{11}{+0.15} & \sixdecrease{18}{-0.23} & \sixdecrease{56}{-2.17} & \sixincrease{14}{+0.25} & \sixincrease{22}{+0.60} & \sixincrease{13}{+0.22} & \sixdecrease{31}{-0.67\textsuperscript{\(\dagger\)}} \\
     & GPT-5 mini & \sixincrease{36}{+1.61} & \sixincrease{16}{+0.33} & \sixdecrease{12}{-0.10} & -- & \sixincrease{27}{+0.90} & \sixincrease{27}{+0.90} & \sixincrease{24}{+0.73} & \sixincrease{16}{+0.33\textsuperscript{\(\dagger\)}} \\
     & GPT-5.5 & \sixincrease{24}{+0.71} & \sixincrease{6}{+0.05} & 0.00 & 0.00\textsuperscript{\(\dagger\)} & \sixincrease{17}{+0.35} & \sixincrease{17}{+0.38} & 0.00 & \sixincrease{23}{+0.67\textsuperscript{\(\dagger\)}} \\
     & Sonnet 5 & \sixincrease{23}{+0.67} & 0.00 & \sixdecrease{20}{-0.29} & \sixdecrease{54}{-2.00\textsuperscript{\(\dagger\)}} & \sixincrease{19}{+0.45} & \sixincrease{8}{+0.08} & \sixincrease{5}{+0.03} & 0.00\textsuperscript{\(\dagger\)} \\
    \cmidrule(lr){2-10}
     & \textit{Strict mean} & +1.39 & +0.42 & +0.08 & -1.09 & +0.57 & +0.56 & +0.27 & +0.07 \\
    \cmidrule(lr){1-10}
    \multicolumn{2}{l}{\textit{Opus 4.8 mean}} & +1.88 & +0.71 & +0.38 & -0.70 & +0.47 & +0.43 & +0.22 & +0.23 \\
    \midrule
    \multicolumn{2}{l}{\textbf{Overall mean}} & \textbf{+1.42} & \textbf{+0.51} & \textbf{+0.19} & \textbf{-0.88} & \textbf{+0.28} & \textbf{+0.23} & \textbf{+0.10} & \textbf{+0.20} \\
    \bottomrule
  \end{tabular}
  \end{adjustbox}
\end{table}

\subsection{The Dimension Hierarchy Is Broadly Preserved across Human-Assessed Quality Levels}
\label{subsec:human_score_conditioning}

\textbf{The same dimensional hierarchy persists across human-assessed quality levels, but human-rated quality does not produce a consistent sensitivity gradient.} Across sufficiently populated rating ranges, evidence framing, novelty stance, and scope framing retain the clearest positive-negative contrasts (Table~\ref{tab:human_score_directional_separation}); each shows the intended ordering for more than 100 of 120 manuscripts (Appendix~\ref{app:paper_consistency}). Rewrite effects neither systematically increase nor decrease with human rating; stronger papers are not consistently more resistant, nor weaker papers more responsive. Strict-prompt results preserve the same pattern (Appendix Table~\ref{tab:human_score_directional_separation_strict}).

\subsection{The AI Reviewer's Starting Score Strongly Conditions the Direction of Change}
\label{subsec:reviewer_score_conditioning}

\textbf{The AI reviewer’s starting score strongly predicts the direction of movement: low initial scores tend to rise, whereas high initial scores tend to fall.}
Table~\ref{tab:score_boundary_directional_separation} shows this pattern for both rewriters. Pooling both rewriters and all reviewers within paper under standard evaluation, positive evidence variants average $+1.05$ OA for original scores in $[1,3]$ but $-0.19$ for scores in $[8,10]$ (Appendix Table~\ref{tab:score_band_rating_inference_combined}). Near the weak-accept cutoff, rewrites produce upward crossings from $[4,5]$ and downward crossings from $[6,7]$, while the middle ranges retain positive-negative contrasts for evidence, novelty, and scope. Strict-prompt results appear in Appendix Table~\ref{tab:score_boundary_directional_separation_strict}.

Unlike human-assessed quality, the AI reviewer’s starting score is strongly associated with the direction of movement. This descriptive pattern may partly reflect scale bounds and regression to the mean; it can outweigh the intended rewrite direction at the extremes but varies across reviewers (Table~\ref{tab:dimension_model_pair_rating_changes}). Section~\ref{sec:reviewer_conditions} examines that heterogeneity.

\begin{findingbox}[title={Finding 1}]
Evidence framing and novelty stance produce the largest and most consistent contrasts, with scope framing forming a weaker second tier. This hierarchy persists across human-assessed quality levels, whereas score movement is more strongly associated with the AI reviewer's initial score.
\end{findingbox}

\section{Do More Complex Rewrites Yield Reliable Gains?}
\label{sec:targeted_adaptation}

\begin{figure}[t]
  \centering
  \includegraphics[width=0.98\textwidth]{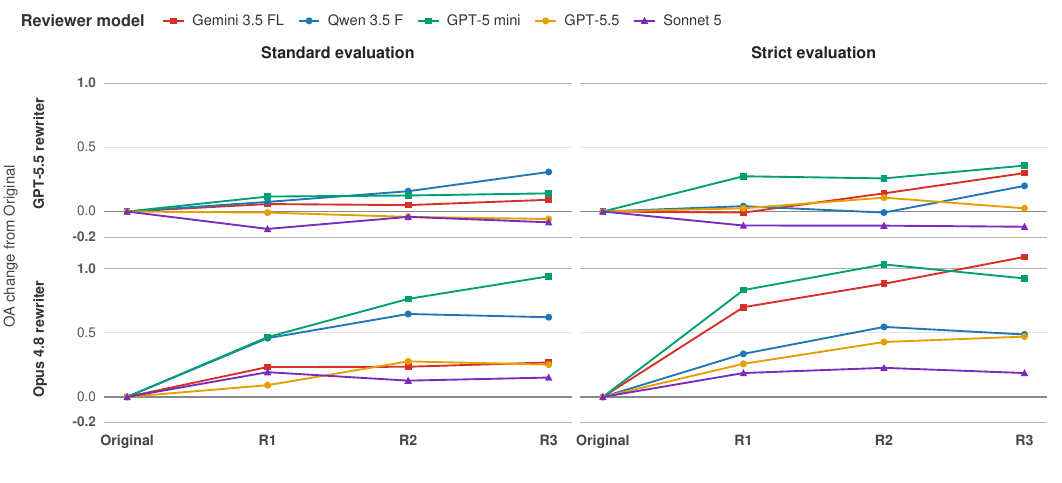}
  \caption{\textbf{Reviewer-specific OA trajectories under recursive joint
rewriting.}
Each round jointly applies all six positive-direction interventions to the
manuscript produced in the preceding round. Rows indicate the rewriter,
columns the evaluation protocol, and colored lines the reviewer model.
Original is normalized to zero, and R1, R2, and R3 denote the three rewrite
rounds showing the cumulative mean OA change after each round. Additional
statistics are in Appendix~\ref{app:combo_trajectory_details}.}
  \label{fig:combo_trajectory}
\end{figure}

Across the multi-stage conditions defined in Section~\ref{app:multistage_rewriting}, score gains depend on the rewriter-reviewer configuration and are generally concentrated by the second round.

\subsection{Joint Rewriting Produces Rewriter-Dependent Gains}
\label{subsec:combo_composition}

\textbf{Only Opus~4.8 joint rewrites yield positive mean changes for every reviewer under both prompts.}
Table~\ref{tab:joint_guided_summary} shows reviewer-averaged gains of $+0.289$ under standard evaluation and $+0.463$ under strict evaluation for Opus~4.8, versus only $+0.021$ and $+0.045$ for GPT-5.5. The latter near-zero means reflect heterogeneous reviewer responses rather than invariance to the rewrite.

\textbf{Combining favorable rhetorical objectives does not necessarily outperform applying them individually.}
Relative to the paper-matched mean of the six positive single-dimension rewrites, the Opus~4.8 joint rewrite gains $+0.160$ OA, whereas the GPT-5.5 rewrite falls $-0.074$ below it. This cross-batch comparison is descriptive rather than a synergy estimate. \textbf{Joint-rewrite effects also track the AI reviewer’s starting score more closely than human-assessed paper quality}: mean OA change declines from $+1.42$ for initial scores in $[1,3]$ to $-0.88$ for scores in $[8,10]$, while effects remain comparatively stable across populated human OA ranges (Table~\ref{tab:joint_rewrite_score_ranges}). As in Section~\ref{subsec:reviewer_score_conditioning}, this score-range pattern is descriptive and may partly reflect scale bounds and regression to the mean.

\begin{figure}[t]
  \centering
  \includegraphics[width=0.88\textwidth]{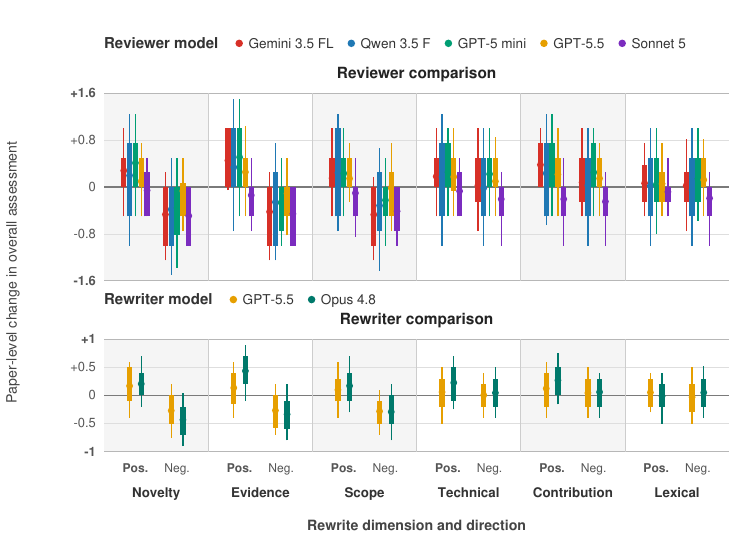}
  \caption{\textbf{Reviewer and rewriter heterogeneity in effects.} Paper-level $\Delta$OA distributions are shown across rhetorical dimensions and directions. The upper panel compares reviewers after averaging over rewriters and protocols, while the lower panel compares rewriters after averaging over reviewers and protocols. Thin lines show the interval from the 10th to the 90th percentile, thick lines the interquartile range, and dots the mean.}
  \vspace{-3mm}
  \label{fig:dimension_response_distributions}
\end{figure}

\subsection{Reviewer Guidance Can Improve on One Pass but Adds Little beyond Repetition}
\label{subsec:review_guidance}

Relative to a single joint rewrite, \textbf{reviewer-averaged scores improve only when Opus~4.8 is used as the rewriter.} Its OA gain increases from $+0.289$ to $+0.396$ under standard evaluation and from $+0.463$ to $+0.557$ under strict evaluation. GPT-5.5 shows no corresponding improvement: its mean gain changes from $+0.021$ to $+0.015$ under standard evaluation and from $+0.045$ to $+0.032$ under strict evaluation. The overall increase from $+0.204$ to $+0.250$ is therefore driven by Opus~4.8.

\textbf{Guided rewriting performs slightly worse than an unguided second pass in all four rewriter-prompt averages}: the guided-minus-unguided contrasts are $-0.035$ and $-0.042$ for GPT-5.5 under standard and strict evaluation, and $-0.008$ and $-0.067$ for Opus~4.8. Because the workflows begin from independently generated first-pass rewrites, this comparison is descriptive rather than a causal estimate of feedback. Weak-accept effects are less consistent and are reported with reviewer-specific intervals in Appendix~\ref{app:review_final_endpoints}.

\subsection{Additional Rewrite Rounds Produce Uneven Gains across Configurations}
\label{subsec:combo_repetition}

Figure~\ref{fig:combo_trajectory} reports the recursive joint-rewrite trajectories defined in Section~\ref{app:multistage_rewriting}. Additional rounds can further increase OA, but the gains depend strongly on the rewriter and reviewer configuration. \textbf{For Opus~4.8, most of the improvement occurs in the second round}: the reviewer-averaged gain rises from $+0.289$ to $+0.410$ under standard and from $+0.463$ to $+0.624$ under strict. The third round adds little further improvement. \textbf{GPT-5.5 remains much less responsive}, reaching only $+0.080$ under standard and $+0.153$ under strict by round~3.

\textbf{Reviewer-specific trajectories are also uneven.} Some continue to increase across rounds, while others flatten or partially reverse earlier gains. More elaborate rewriting is therefore not a model-agnostic route to better evaluation; it can amplify compatibility between a rewriter and reviewer rather than improve assessment of the underlying science. Most additional gains remain concentrated in specific rewriter-reviewer configurations and diminish after the second pass. Weak-accept trajectories show a similarly mixed pattern and are reported in Appendix~\ref{app:combo_trajectory_details}.

\begin{findingbox}[title={Finding 2}]
More elaborate rewriting yields configuration-dependent and diminishing gains: joint-rewrite gains are substantial with Opus~4.8 but near zero with GPT-5.5. Reviewer guidance does not outperform an unguided second pass on average, and later rounds add little.
\end{findingbox}

\section{How Do Rewriters, Reviewers, and Protocols Shape Rhetorical Effects?}
\label{sec:reviewer_conditions}

\begin{table*}
  \centering
  \captionsetup{hypcap=false}
  \scriptsize
  \setlength{\tabcolsep}{1.3pt}
  \begin{adjustbox}{width=0.7\textwidth}
  \begin{tabular}{@{}ll*{10}{c}@{}}
    \toprule
    \textbf{Reviewer} & \textbf{Rewriter}
      & \multicolumn{5}{c}{\textbf{Absolute rewritten OA}}
      & \multicolumn{5}{c}{\textbf{Baseline-adjusted rewrite effect}} \\
    \cmidrule(lr){3-7}\cmidrule(lr){8-12}
      & & \textbf{Std.} & \textbf{Strict}
      & \boldmath$\Delta_{\mathrm{P}}$
      & \boldmath$\rho_{\mathrm{P}}$
      & \textbf{Lower}
      & \textbf{Std.} & \textbf{Strict}
      & \boldmath$\Delta_{\mathrm{P}}$
      & \boldmath$\rho_{\mathrm{P}}$
      & \textbf{Lower} \\
    \midrule
    Gemini 3.5 FL & GPT-5.5 & 7.706 & 6.503 & -1.203 & .791 & 92.5\% & -.010 & +.045 & +.056 & .021 & 44.2\% \\
     & Opus 4.8 & 7.699 & 6.544 & -1.155 & .786 & 90.8\% & -.017 & +.086 & +.103 & -.058 & 42.5\% \\
    \cmidrule(lr){2-12}
     & \textit{Mean} & 7.703 & 6.524 & -1.179 & .788 & 91.7\% & -.014 & +.066 & +.080 & -.019 & 43.3\% \\
    \midrule
    Qwen 3.5 F & GPT-5.5 & 6.984 & 4.697 & -2.287 & .746 & 100.0\% & +.109 & -.103 & -.212 & .087 & 49.2\% \\
     & Opus 4.8 & 6.976 & 4.795 & -2.181 & .790 & 100.0\% & +.101 & -.005 & -.106 & .059 & 49.2\% \\
    \cmidrule(lr){2-12}
     & \textit{Mean} & 6.980 & 4.746 & -2.234 & .768 & 100.0\% & +.105 & -.054 & -.159 & .073 & 49.2\% \\
    \midrule
    GPT-5 mini & GPT-5.5 & 6.338 & 4.422 & -1.917 & .863 & 100.0\% & +.047 & +.155 & +.108 & .109 & 46.7\% \\
     & Opus 4.8 & 6.404 & 4.435 & -1.969 & .876 & 100.0\% & +.112 & +.168 & +.056 & .082 & 47.5\% \\
    \cmidrule(lr){2-12}
     & \textit{Mean} & 6.371 & 4.428 & -1.943 & .869 & 100.0\% & +.080 & +.161 & +.082 & .095 & 47.1\% \\
    \midrule
    GPT-5.5 & GPT-5.5 & 5.435 & 4.751 & -.685 & .949 & 90.8\% & +.052 & +.109 & +.057 & .225 & 45.8\% \\
     & Opus 4.8 & 5.465 & 4.777 & -.688 & .943 & 92.5\% & +.082 & +.135 & +.053 & .243 & 47.5\% \\
    \cmidrule(lr){2-12}
     & \textit{Mean} & 5.450 & 4.764 & -.686 & .946 & 91.7\% & +.067 & +.122 & +.055 & .234 & 46.7\% \\
    \midrule
    Sonnet 5 & GPT-5.5 & 4.893 & 4.150 & -.743 & .932 & 89.7\% & -.275 & -.252 & +.023 & .041 & 40.2\% \\
     & Opus 4.8 & 5.055 & 4.270 & -.785 & .945 & 93.3\% & -.166 & -.143 & +.022 & .000 & 42.3\% \\
    \cmidrule(lr){2-12}
     & \textit{Mean} & 4.974 & 4.210 & -.764 & .938 & 91.5\% & -.220 & -.197 & +.023 & .021 & 41.2\% \\
    \midrule
    All reviewers & \textit{GPT-5.5} & 6.271 & 4.905 & -1.367 & .856 & 94.6\% & -.016 & -.009 & +.006 & .096 & 45.2\% \\
     & \textit{Opus 4.8} & 6.320 & 4.964 & -1.356 & .868 & 95.3\% & +.023 & +.048 & +.026 & .065 & 45.8\% \\
    \midrule
    \multicolumn{2}{l}{\textbf{Overall mean}} & \textbf{6.296} & \textbf{4.934} & \textbf{-1.361} & \textbf{.862} & \textbf{95.0\%} & \textbf{+.003} & \textbf{+.020} & \textbf{+.016} & \textbf{.081} & \textbf{45.5\%} \\
    \bottomrule
  \end{tabular}
  \end{adjustbox}
 \captionof{table}{\textbf{Effects of review protocol on absolute scores and rewrite
sensitivity.}
Rewritten OA is averaged within paper across the 12 single-dimension rewrites,
and baseline-adjusted effects are measured relative to the prompt-matched
original. For each outcome, $\Delta_{\mathrm{P}}$ is strict minus standard,
$\rho_{\mathrm{P}}$ is the paper-level Spearman correlation across protocols,
and Lower is the percentage of papers with a lower value under strict
evaluation. }
  \label{tab:prompt_score_comparison}
  \label{tab:prompt_absolute_comparison}
  \label{tab:prompt_response_comparison}
\end{table*}

\begin{table}[t]
  \centering
  \caption{\textbf{Changes in secondary review scores and their association with
OA.}
For each outcome, Mean $\Delta$ (SD) reports the paper-level mean paired change,
with the standard deviation (SD) in parentheses, while
$\rho_{\Delta\mathrm{OA}}$ reports its Spearman correlation with the change in
overall assessment.}
  \label{tab:secondary_score_oa_association}
  \scriptsize
  \begin{tabular}{llcccccc}
    \toprule
    \textbf{Review prompt} & \textbf{Reviewer}
      & \multicolumn{2}{c}{\textbf{Soundness}}
      & \multicolumn{2}{c}{\textbf{Presentation}}
      & \multicolumn{2}{c}{\textbf{Contribution}} \\
    \cmidrule(lr){3-4}\cmidrule(lr){5-6}\cmidrule(l){7-8}
      & & \textbf{Mean \boldmath$\Delta$ (SD)}
      & \boldmath$\rho_{\Delta\mathrm{OA}}$
      & \textbf{Mean \boldmath$\Delta$ (SD)}
      & \boldmath$\rho_{\Delta\mathrm{OA}}$
      & \textbf{Mean \boldmath$\Delta$ (SD)}
      & \boldmath$\rho_{\Delta\mathrm{OA}}$ \\
    \midrule
    \multicolumn{8}{l}{\textbf{GPT-5.5 rewriter}} \\
    \cmidrule(lr){1-8}
    Standard & Gemini 3.5 FL & +.027 (.282) & .416 & +.019 (.131) & .135 & -.010 (.268) & .651 \\
     & Qwen 3.5 F & +.042 (.449) & .500 & +.006 (.406) & .275 & -.020 (.425) & .747 \\
     & GPT-5 mini & +.048 (.232) & .261 & +.087 (.198) & -.045 & +.007 (.298) & .500 \\
     & GPT-5.5 & +.067 (.282) & .295 & +.037 (.190) & .196 & +.024 (.280) & .298 \\
     & Sonnet 5 & -.041 (.300) & .124 & +.040 (.307) & .018 & -.101 (.303) & -.025 \\
    \addlinespace
    Strict & Gemini 3.5 FL & +.064 (.370) & .631 & +.076 (.295) & .164 & +.063 (.374) & .795 \\
     & Qwen 3.5 F & +.008 (.483) & .388 & +.024 (.525) & .167 & -.039 (.419) & .715 \\
     & GPT-5 mini & +.135 (.390) & .460 & +.007 (.118) & -.052 & +.069 (.365) & .446 \\
     & GPT-5.5 & +.104 (.310) & .373 & +.010 (.072) & .038 & -.005 (.331) & .439 \\
     & Sonnet 5 & -.064 (.288) & .349 & -.008 (.219) & .061 & -.072 (.352) & .187 \\
    \midrule
    \multicolumn{8}{l}{\textbf{Opus 4.8 rewriter}} \\
    \cmidrule(lr){1-8}
    Standard & Gemini 3.5 FL & +.010 (.286) & .504 & +.009 (.134) & .101 & +.006 (.251) & .633 \\
     & Qwen 3.5 F & +.038 (.427) & .458 & -.060 (.402) & .295 & +.003 (.429) & .733 \\
     & GPT-5 mini & +.031 (.230) & .214 & +.097 (.216) & -.056 & +.035 (.278) & .635 \\
     & GPT-5.5 & +.049 (.271) & .349 & +.041 (.194) & .180 & +.045 (.290) & .388 \\
     & Sonnet 5 & -.017 (.307) & -.011 & +.025 (.295) & .074 & -.060 (.291) & -.108 \\
    \addlinespace
    Strict & Gemini 3.5 FL & +.087 (.383) & .587 & +.071 (.319) & .178 & +.089 (.369) & .791 \\
     & Qwen 3.5 F & +.017 (.449) & .361 & +.019 (.492) & .107 & +.007 (.421) & .680 \\
     & GPT-5 mini & +.087 (.384) & .479 & +.013 (.109) & .003 & +.067 (.347) & .477 \\
     & GPT-5.5 & +.094 (.322) & .346 & +.007 (.088) & .087 & +.024 (.335) & .308 \\
     & Sonnet 5 & -.046 (.268) & .363 & -.017 (.230) & .063 & -.038 (.313) & .159 \\
    \bottomrule
  \end{tabular}
\end{table}

We separate how rewriter choice, reviewer choice, and evaluation protocol shape rewrite effects. Rewriters mainly shape rhetorical contrasts, reviewers differ in how they translate into scores and rubric dimensions, and strict evaluation primarily shifts the scoring scale.

\subsection{Reviewer and Rewriter Models Produce Different Patterns of Rewrite Effects}
\label{subsec:planned_reviewer_heterogeneity}

\textbf{Reviewers generally prefer the same rhetorical direction while disagreeing on whether it raises or lowers scores relative to the original.}
Figure~\ref{fig:dimension_response_distributions} shows a shared preference for positive novelty, evidence, and scope variants, but substantial differences in the magnitude and overall level of score changes. Qwen~3.5 F and GPT-5 mini show the widest response distributions, whereas GPT-5.5 moves less and Sonnet~5 shifts effects downward. Opus~4.8 generally produces wider positive-negative separation than GPT-5.5, rather than a uniform upward shift. Thus, reviewer choice shapes how interventions translate into scores, whereas rewriter choice mainly shapes the strength of the rhetorical contrast. Reviewers also agree more on paper rankings than on which papers benefit from rewriting; \textbf{agreement about paper quality therefore does not imply agreement about intervention effects} (Appendix Tables~\ref{tab:reviewer_response_comparison} and~\ref{tab:secondary_cross_reviewer_consistency}).

\subsection{Strict Review Lowers Scores but Does Not Consistently Change Rewrite Effects}
\label{subsec:review_prompt_conditioning}

\textbf{Strict review primarily recalibrates the scoring scale downward.}
Table~\ref{tab:prompt_score_comparison} shows that strict review substantially lowers absolute scores. Mean OA decreases from $6.296$ to $4.934$, with $95.0\%$ of papers receiving a lower score on average across configurations. Paper rankings remain broadly stable, however, with a mean standard-strict Spearman correlation of $.862$ in absolute OA.

\textbf{This downward shift does not produce a consistent change in rewrite effects.} After subtracting the prompt-matched original score, the overall mean rewrite effect changes only from $+.003$ under standard to $+.020$ under strict, and the direction varies across reviewers: Qwen~3.5 F becomes less positive, GPT-5 mini becomes more positive, and Sonnet~5 remains negative. Consistent with this heterogeneity, the mean cross-protocol Spearman correlation in paper-level rewrite effects is only $.081$. Stricter prompting, therefore, changes reviewer severity without making the review more invariant to content-preserving rhetoric.

\vspace{-2.2mm}
\subsection{Associations with Secondary Scores Vary across Reviewers}
\label{subsec:secondary_oa_coupling}

\textbf{Similar OA changes co-move with different rubric dimensions across reviewers.}
Table~\ref{tab:secondary_score_oa_association} shows that $\Delta$OA is only weakly associated with presentation changes and more strongly associated with contribution and soundness across most configurations. Qwen~3.5 F aligns OA changes mainly with contribution, whereas strict Gemini~3.5 FL reviews show strong associations with both contribution and soundness; Sonnet~5 is weaker and less stable. The same rhetorical changes are mapped onto scientific-merit dimensions differently across reviewers, limiting the interpretability of rubric-level explanations. These associations indicate co-movement, not causal mediation or genuine changes in merit; full estimates appear in Appendix~\ref{app:secondary_score_association}.

\begin{findingbox}[title={Finding 3}]
Rewriters shape the separation between rhetorical variants, whereas reviewers differ in the magnitude, sign, and rubric associations of score changes. Strict prompts lower mean OA by 1.36 points but do not consistently strengthen or weaken rewrite sensitivity.
\end{findingbox}

\section{Conclusion}
\label{sec:conclusion}

This paper shows that AI scientific review is systematically sensitive to rhetorical presentation even when reported scientific content is preserved. This sensitivity is not uniform. It depends on the rhetorical dimension, the rewriting process, the rewriter and reviewer models, and the review protocol. The resulting variation cannot be reduced to a single average effect or treated as a stable property of one model configuration. These findings suggest that AI-assisted review should be evaluated for rhetorical robustness across multiple models and conditions, rather than judged solely by aggregate agreement or average scoring behavior.

\section*{Limitations}

This study has several limitations. First, the benchmark is restricted to ICLR 2026 submissions with recoverable full-paper source and public review metadata, so the results may not generalize to other venues or scientific fields. Second, full-paper rewriting and multi-model evaluation are computationally expensive. Some planned reviews did not produce valid structured records, and this missingness may not be random because some failures appear related to paper topics triggering model safeguards. Third, the six rhetorical dimensions are not orthogonal. Each rewrite is a controlled full-paper intervention, so its effect should not be interpreted as an independent coefficient for a single linguistic feature. Finally, most configurations use one AI review per manuscript, with repeated reviews examined only in an auxiliary audit. The results, therefore, characterize the tested models and prompts rather than human review or all possible variations from repeated model sampling.

\section*{Ethical Considerations}

This study uses publicly available review metadata and public arXiv sources. Before manuscripts were sent to model-provider APIs, we removed author names, affiliations, acknowledgments, submission-status statements, and identity-bearing links. We report aggregate results and do not assess individual authors. The framework is dual use: controlled rewriting can diagnose rhetorical sensitivity, but similar procedures could also be used to tailor manuscripts to AI-reviewer preferences without improving the underlying research. We therefore caution against interpreting score increases as improvements in scientific merit or using the framework to manipulate active review processes. Such optimization could reward rhetorical assertiveness over evidential improvement and distort scientific evaluation.

\newpage
\bibliographystyle{assets/plainnat}
\bibliography{main}

\newpage
\beginappendix

\renewcommand{\topfraction}{0.90}
\renewcommand{\bottomfraction}{0.80}
\renewcommand{\textfraction}{0.08}
\renewcommand{\floatpagefraction}{0.70}
\setcounter{topnumber}{3}
\setcounter{bottomnumber}{2}
\setcounter{totalnumber}{5}

\section{Detailed Related Work}
\label{app:detailed_related_work}

\subsection{LLM-Based Evaluation}
\label{subsec:llm_based_evaluation}

LLMs are increasingly used as reference-free evaluators when the target output has no single verifiable answer. Rubric-conditioned scoring in G-Eval and pairwise comparison in MT-Bench demonstrated that LLM judgments can align substantially with human preferences on open-ended generation tasks~\citep{liu2023g,zheng2023judging}. Dedicated evaluator models such as Prometheus~\citep{kim2024prometheus} further showed that fine-grained rubrics and reference materials can support customized scoring and feedback. These works established the practical case for LLM-as-a-judge while leaving open how LLM evaluators respond to differences in the presentation of otherwise comparable content.

Subsequent work has examined several ways in which LLM judgments vary across presentation and evaluation conditions. Candidate order can reverse pairwise preferences, and response length can substantially alter automatic win rates~\citep{wang2024large,dubois2024length}. Scores also vary with prompt wording, anchors, familiarity, and repeated sampling, while judges may recognize and favor outputs from their own model family~\citep{stureborg2024large,panickssery2024llm}. More targeted counterfactual audits show that epistemic markers can lower evaluations even when answer correctness is unchanged~\citep{lee2025llm}. Large-scale meta-evaluation likewise finds that alignment depends on the judge, task, property, and data source~\citep{bavaresco2025llms}. Semantically equivalent reformulations of evaluator instructions can produce a marked accuracy-robustness gap, with stability depending more on the verifiability of the evaluated attribute than on model scale~\citep{bhat2026all}. Adaptive attacks go further by learning which meaning-preserving stylistic edits inflate a particular judge's scores, including in a case study where generated review texts are themselves evaluated outputs~\citep{yang2026turning}. These findings show that LLM judgments vary with both the presentation of evaluated content and the design of the evaluation process, motivating a more systematic analysis of where these effects arise and how they are structured.

\subsection{AI-Assisted Scientific Evaluation}
\label{subsec:ai_scientific_evaluation}

Traditionally, computational methods have been used to support scientific evaluation. PeerRead made manuscripts, expert reviews, aspect scores, and publication decisions available for tasks such as acceptance and review-score prediction~\citep{kang2018dataset}. ReviewRobot subsequently combined manuscript and background knowledge to generate category-specific scores and evidence-linked comments~\citep{wang2020reviewrobot}, while ReviewAdvisor formalized review desiderata and generated aspect-aware first-pass feedback from full papers~\citep{yuan2022can}. This line of work established that parts of the reviewing workflow can be modeled computationally, but also showed that broad coverage or an accurate summary does not by itself yield factual, constructive, and decision-worthy criticism.

LLMs have made automated feedback considerably more fluent and apparently useful. In a large-scale study, GPT-4 feedback overlapped with human review comments at rates comparable to human-human overlap, and many authors reported finding the feedback useful~\citep{liang2024can}. Direct evaluations of reviewing competence are more qualified: LLMs remain weak at long-paper processing, zero-shot scoring, and consistently correct critical feedback~\citep{zhou2024llm}, and large-scale comparisons find that they capture summaries and stated strengths more readily than substantive weaknesses, discriminating questions, or differences in paper quality~\citep{li2025unveiling}. \textsc{PeerCheck} similarly documents differences between human and LLM-generated reviews, finding that chain-of-thought prompting can improve similarity to human reviews while retrieval effects depend on the reviewer model~\citep{chen2026peercheck}. Reviewer-side guideline design also affects automated review: official conference guidelines produce judgments more consistent with human scores, whereas rigid reviewer-imitating rubrics can reduce that alignment~\citep{li2026evaluating}. This reviewer-side comparison is distinct from our standard and strict protocols, which condition the manuscript-side counterfactual analysis. At the same time, a randomized deployment shows that presenting AI feedback to human reviewers can improve review specificity and actionability~\citep{thakkar2026large}. These studies show that LLMs can support scientific evaluation in multiple roles, but that the quality and character of their judgments depend on both the reviewing task and the conditions under which evaluation is conducted.

\subsection{Rhetorical Rewriting and Presentation Effects}
\label{subsec:paper_laundering}

Scientific evaluation necessarily responds to how evidence and contributions are presented. Academic discourse research has long studied hedging, stance, metadiscourse, and contribution positioning as mechanisms for calibrating claims and organizing scientific arguments~\citep{hyland1998hedging,hyland2018metadiscourse}. Computational evidence likewise links linguistic certainty to perceived scientific rigour~\citep{james2024rigour}. Sensitivity to presentation is therefore not inherently problematic, since clarity and exposition are legitimate review criteria. The question is whether rhetorical changes also affect judgments of scientific merit when the underlying methods and reported evidence are preserved while captions, transitions, and interpretations of results are allowed to vary. Even limited interventions can produce such effects: stylistic variants of a paper title can change AI-review scores assigned to the same abstract~\citep{du2025trap}.

Prior work on manuscript-side interventions has examined how changes to a paper can influence automated reviewers. Hidden instructions embedded in manuscripts can inflate ratings, suppress criticism, or redirect generated reviews~\citep{ye2024we,collu2026misleading}, while character-, word-, and sentence-level perturbations can distort review judgments~\citep{lin2025breaking}. These interventions rely on concealed instructions or artificial perturbations rather than ordinary academic rewriting. More recent work examines visible revisions that preserve some or all of the underlying content. At the abstract level, the Paraphrasing Adversarial Attack framework searches over meaning-preserving rewrites to improve evaluator scores~\citep{kaneko2026paraphrasing}, while repeated optimization of paraphrasing, rewriting, and overclaiming can substantially increase AI-review scores~\citep{li2026gaming}. At the full-paper level, \citet{baumann2026stop} uses \emph{paper laundering} to describe automated rewriting intended to raise AI-review scores without new experiments. \citet{yang2026no} similarly introduces \emph{adversarial repackaging}, a closed-loop procedure that searches for favorable presentation strategies while preserving methods, experiments, figures, equations, proofs, and numerical results. Together, these studies show that manuscript presentation can be systematically optimized to influence AI-based evaluation.

Related work also examines a distinct question. \citet{dycke2026automatic} deliberately disrupts the relationships among results, interpretations, and claims, using meaning-preserving edits as controls, to test whether automatic reviewers detect the resulting substantive defects. We instead map how controlled full-manuscript rhetorical interventions affect review outcomes when the underlying methods, experimental settings, and reported results are preserved.

Prior rewriting studies are largely designed to identify revisions that increase evaluator scores and therefore emphasize favorable candidates or aggregate acceptance gains. We instead conduct a controlled analysis across six prespecified rhetorical dimensions, each applied in opposing directions from the same manuscript baseline. We retain all rewritten outcomes, including those that receive lower scores. This design allows us to identify which rhetorical dimensions affect overall ratings, scientific subscores, and weak-accept decisions, where these effects occur along the score scale, and how they vary across rewriting models, reviewer models, and evaluation protocols.

\newpage
\section{AI-Review Execution Details}
\label{app:protocol}

\subsection{AI-Review Execution}

Each compiled PDF was submitted through OpenRouter’s PDF file-input interface~\footnote{\url{https://openrouter.ai/docs/guides/overview/multimodal/pdfs}} using the default \texttt{auto} processing mode, rather than being manually converted through OCR or page-to-image preprocessing.
No browser, web search, file search, retrieval, or other external tool was enabled. Temperature, top-$p$, and other sampling controls were not overridden, so each serving endpoint used its provider defaults. Reviewers received no metadata about the rhetorical condition or rewrite model.

The reported knowledge cutoff dates are May 31, 2024 for GPT-5 mini \citep{openai_gpt5mini}, December 1, 2025 for GPT-5.5 \citep{openai_gpt55}, January 2026 for Sonnet~5 \citep{anthropic_claude_sonnet5}, and January 2025 for Gemini~3.5 FL \citep{google_gemini35flashlite}.

\subsection{Review Missingness and Potential Safeguard Triggers}
\label{app:review_refusal_example}

Across the primary and multi-stage experiments, 42,480 reviews were planned and 42,396 produced valid structured records, leaving 84 evaluations missing. Missing evaluations resulted from model outputs that remained invalid or unavailable after repeated retries. These evaluations were retained as missing rather than imputed, and all analyses use available matched pairs.

A notable concentration occurred for \emph{JULI: Jailbreak Large Language Models by Self-Introspection}, which studies methods for eliciting harmful responses from safety-aligned language models. Sonnet~5 repeatedly failed to produce valid records for 26 evaluation cells involving this paper despite multiple retry attempts. Because the manuscript contains potentially safeguard-triggering instructions and example responses, this pattern is consistent with the activation of model safety mechanisms, although the available records do not establish the cause of each failure. One additional Qwen strict joint-rewrite evaluation was explicitly rejected by provider content inspection and remained unavailable after repeated attempts.

Prompt~\ref{prompt:potential_triggers} reproduces potentially safeguard-triggering instructions and responses from Appendix~A.15 of \citet{wang2026juli}.

\promptfigure{Potentially Safeguard-Triggering Instructions and Responses}
{prompts/potential_safeguard_triggering_examples.txt}
\label{prompt:potential_triggers}

\subsection{API Cost Accounting}

All rewriting and reviewing were conducted through model-provider APIs.
Table~\ref{tab:api_costs} separates direct API expenditure into rewriting and
reviewing costs for each experimental stage. Single-dimension rewriting cost
\$19,627.77 in total, comprising \$14,413.72 for rewriting and \$5,214.05 for
reviewing. The joint-rewrite stage cost \$1,656.25, the recursive joint-rewrite
stage cost \$5,144.49, and the reviewer-guided stage cost \$2,737.38.
Across all four stages, rewriting accounted for \$20,583.34 and reviewing for
\$8,582.55, yielding a total direct API cost of \$29,165.89.

\begin{table}[ht]
  \centering
  \caption{Direct API cost accounting by experimental stage.}
  \label{tab:api_costs}
  \small
  \begin{tabular}{lrrr}
    \toprule
    Experimental stage & Rewrite & Review & Stage total \\
    \midrule
    Single-dimension rewrite
      & \$14,413.72 & \$5,214.05 & \$19,627.77 \\
    Joint rewrite
      & \$996.56 & \$659.69 & \$1,656.25 \\
    Recursive joint rewrite
      & \$3,171.06 & \$1,973.43 & \$5,144.49 \\
    Reviewer-guided rewrite
      & \$2,002.00 & \$735.38 & \$2,737.38 \\
    \midrule
    \textbf{Total}
      & \textbf{\$20,583.34}
      & \textbf{\$8,582.55}
      & \textbf{\$29,165.89} \\
    \bottomrule
  \end{tabular}
\end{table}


\subsection{Comparison with Human Overall Assessments}
\label{app:human_ai_comparison}

As an external descriptive comparison, we contrast each AI reviewer's rating of the anonymized original with the paper's mean overall assessment (OA) from the official ICLR reviews. Table~\ref{tab:baseline_human_comparison} reports this comparison separately for every reviewer model and review prompt.

\begin{table}[!htbp]
  \centering
  \caption{\textbf{AI versus human overall assessments on the anonymized original papers.} AI minus Human is the paper-level AI rating minus human OA; MAE is mean absolute error and Spearman's $\rho$ measures rank correspondence. Each row contains the same 120 papers.}
  \label{tab:baseline_human_comparison}
  \small
  \setlength{\tabcolsep}{4pt}
  \begin{tabular}{llrrrrr}
    \toprule
    Reviewer model & Review prompt & Human OA & AI rating & AI minus Human & MAE & Spearman $\rho$ \\
    \midrule
    Gemini 3.5 FL & Standard & 4.782 & 7.717 & +2.934 & 2.934 & 0.448 \\
     & Strict & 4.782 & 6.458 & +1.676 & 1.848 & 0.464 \\
    \midrule
    Qwen 3.5 F & Standard & 4.782 & 6.875 & +2.093 & 2.254 & 0.439 \\
     & Strict & 4.782 & 4.800 & +0.018 & 1.283 & 0.402 \\
    \midrule
    GPT-5 mini & Standard & 4.782 & 6.292 & +1.509 & 1.748 & 0.401 \\
     & Strict & 4.782 & 4.267 & -0.516 & 1.295 & 0.395 \\
    \midrule
    GPT-5.5 & Standard & 4.782 & 5.383 & +0.601 & 1.232 & 0.532 \\
     & Strict & 4.782 & 4.642 & -0.141 & 1.029 & 0.621 \\
    \midrule
    Sonnet 5 & Standard & 4.782 & 5.200 & +0.418 & 1.187 & 0.497 \\
     & Strict & 4.782 & 4.442 & -0.341 & 1.142 & 0.565 \\
    \bottomrule
  \end{tabular}
\end{table}

The comparison shows a clear prompt-dependent difference between AI and human scores. Under the standard prompt, all five AI reviewers assign higher average ratings than the human OA, with mean differences ranging from 0.418 to 2.934 points. Under the strict prompt, GPT-5 mini, GPT-5.5, and Sonnet~5 assign lower average ratings than the human OA, Qwen~3.5 F is nearly aligned with it, and Gemini~3.5 FL remains 1.676 points higher. At the paper level, MAE ranges from 1.029 to 2.934 points, while Spearman correlations range from 0.395 to 0.621.

Human OA is the mean of the official ICLR review scores for each paper, whereas the AI rating is generated by a single reviewer model under a specified prompt. The table, therefore, describes differences in scoring level and paper ranking between the two evaluation sources.

\newpage
\section{Detailed Results for Individual Rhetorical Dimensions}
\label{app:single_dimension_results}
\label{app:complete_results}

This appendix follows the progression of Section~\ref{sec:dimension_effects}:
overall-rating effects and fixed-model heterogeneity, secondary-score and
paper-level consistency checks, weak-accept decisions, and score-range
heterogeneity. It expands Table~\ref{tab:dimension_model_pair_rating_changes},
reports the complete configuration-level weak-accept results in
Table~\ref{tab:dimension_threshold_summary_main}, and then provides the
underlying score-range decompositions. In the decomposition tables, each cell
reports original, rewrite (change); no arrow notation is used. Original is
always the anonymized PDF evaluated by the same reviewer model and under the
same evaluation protocol as its rewrite. To retain the complete configuration
record without obscuring the main comparisons, each block presents the
paper-level or panel-aligned result first and places the full fixed-model
decomposition afterward.

\subsection{Overall-Rating Effects}
\label{app:dimension_rating_details}

Table~\ref{tab:dimension_panel_rating_combined} reports the pooled original
and rewrite means together with paired changes and paper-bootstrap confidence
intervals. It pools both rewrite models and averages reviewer-model
evaluations within paper, complementing the fixed-pair changes in
Table~\ref{tab:dimension_model_pair_rating_changes}.

\begingroup
\scriptsize
\setlength{\tabcolsep}{3.0pt}
\begin{longtable}{llrrlrrl}
\caption{Pooled overall ratings before and after rhetorical rewriting. Original and rewrite are prompt-matched paper-level means; $\Delta$ is rewrite minus original, with 95\% paper-bootstrap confidence intervals. Values pool both rewrite models and average all five reviewer-model evaluations within paper.}\phantomsection\label{tab:dimension_panel_rating_combined}\\
\toprule
& & \multicolumn{3}{c}{Standard} & \multicolumn{3}{c}{Strict} \\
\cmidrule(lr){3-5}\cmidrule(lr){6-8}
Dimension & Direction & Original & Rewrite & $\Delta$ [95\% CI] & Original & Rewrite & $\Delta$ [95\% CI] \\
\midrule
\endfirsthead
\toprule
& & \multicolumn{3}{c}{Standard} & \multicolumn{3}{c}{Strict} \\
\cmidrule(lr){3-5}\cmidrule(lr){6-8}
Dimension & Direction & Original & Rewrite & $\Delta$ [95\% CI] & Original & Rewrite & $\Delta$ [95\% CI] \\
\midrule
\endhead
\midrule
\endfoot
\bottomrule
\endlastfoot
Novelty & Negative & 6.294 & 5.957 & $-0.339\;[-0.418,-0.263]$ & 4.922 & 4.557 & $-0.366\;[-0.454,-0.279]$ \\
Novelty & Positive & 6.293 & 6.429 & $+0.133\;[+0.064,+0.205]$ & 4.922 & 5.163 & $+0.242\;[+0.154,+0.330]$ \\
Evidence & Negative & 6.294 & 6.033 & $-0.263\;[-0.336,-0.184]$ & 4.922 & 4.579 & $-0.342\;[-0.429,-0.256]$ \\
Evidence & Positive & 6.295 & 6.498 & $+0.203\;[+0.131,+0.280]$ & 4.922 & 5.300 & $+0.379\;[+0.291,+0.467]$ \\
Scope & Negative & 6.294 & 6.028 & $-0.268\;[-0.343,-0.191]$ & 4.922 & 4.610 & $-0.311\;[-0.396,-0.226]$ \\
Scope & Positive & 6.294 & 6.448 & $+0.152\;[+0.084,+0.224]$ & 4.922 & 5.043 & $+0.121\;[+0.043,+0.206]$ \\
Technical & Negative & 6.294 & 6.324 & $+0.030\;[-0.034,+0.095]$ & 4.922 & 4.961 & $+0.041\;[-0.052,+0.132]$ \\
Technical & Positive & 6.294 & 6.429 & $+0.133\;[+0.057,+0.212]$ & 4.922 & 5.074 & $+0.152\;[+0.065,+0.239]$ \\
Contribution & Negative & 6.294 & 6.349 & $+0.052\;[-0.021,+0.128]$ & 4.922 & 4.971 & $+0.048\;[-0.030,+0.133]$ \\
Contribution & Positive & 6.293 & 6.457 & $+0.162\;[+0.092,+0.235]$ & 4.922 & 5.144 & $+0.223\;[+0.126,+0.323]$ \\
Lexical & Negative & 6.294 & 6.299 & $+0.003\;[-0.066,+0.072]$ & 4.922 & 4.931 & $+0.009\;[-0.075,+0.096]$ \\
Lexical & Positive & 6.295 & 6.337 & $+0.042\;[-0.024,+0.111]$ & 4.922 & 4.913 & $-0.009\;[-0.084,+0.068]$ \\
\end{longtable}
\endgroup

Table~\ref{tab:dimension_reviewer_rating_matrix} retains the original and
rewrite means while fixing both model axes. It therefore shows whether a
pooled effect reflects a broadly shared response or a large change from one
reviewer model.

\begingroup
\tiny
\setlength{\tabcolsep}{0.8pt}
\begin{longtable}{llccccc}
\caption{Overall ratings by fixed rewrite model and reviewer model. Each cell reports original to rewrite (change); values are paper-level means.}\phantomsection\label{tab:dimension_reviewer_rating_matrix}\\
\toprule
Dimension & Direction & \shortstack{Gemini 3.5\\FL} & \shortstack{Qwen 3.5\\F} & GPT-5 mini & GPT-5.5 & Sonnet 5 \\
\midrule
\endfirsthead
\toprule
Dimension & Direction & \shortstack{Gemini 3.5\\FL} & \shortstack{Qwen 3.5\\F} & GPT-5 mini & GPT-5.5 & Sonnet 5 \\
\midrule
\endhead
\midrule
\endfoot
\bottomrule
\endlastfoot
\rowcolor{black!6}\multicolumn{7}{l}{\textit{Standard protocol; Rewrite model: GPT-5.5}}\\
Novelty & Negative & 7.72$\to$7.47 (-0.25) & 6.88$\to$6.55 (-0.33) & 6.29$\to$5.92 (-0.38) & 5.38$\to$5.38 (0.00) & 5.20$\to$4.73 (-0.47) \\
Novelty & Positive & 7.72$\to$7.83 (+0.12) & 6.88$\to$7.11 (+0.23) & 6.29$\to$6.58 (+0.29) & 5.38$\to$5.44 (+0.06) & 5.20$\to$5.05 (-0.15) \\
Evidence & Negative & 7.72$\to$7.47 (-0.25) & 6.88$\to$6.78 (-0.10) & 6.29$\to$6.11 (-0.18) & 5.38$\to$5.28 (-0.11) & 5.20$\to$4.76 (-0.44) \\
Evidence & Positive & 7.72$\to$7.77 (+0.05) & 6.88$\to$7.23 (+0.36) & 6.29$\to$6.47 (+0.18) & 5.38$\to$5.51 (+0.12) & 5.20$\to$4.89 (-0.31) \\
Scope & Negative & 7.72$\to$7.42 (-0.30) & 6.88$\to$6.58 (-0.30) & 6.29$\to$6.12 (-0.17) & 5.38$\to$5.31 (-0.08) & 5.20$\to$4.75 (-0.45) \\
Scope & Positive & 7.72$\to$7.78 (+0.06) & 6.88$\to$7.25 (+0.38) & 6.29$\to$6.45 (+0.16) & 5.38$\to$5.49 (+0.11) & 5.20$\to$5.02 (-0.18) \\
Technical & Negative & 7.72$\to$7.73 (+0.02) & 6.88$\to$7.03 (+0.15) & 6.29$\to$6.43 (+0.14) & 5.38$\to$5.42 (+0.04) & 5.20$\to$4.97 (-0.23) \\
Technical & Positive & 7.72$\to$7.79 (+0.08) & 6.88$\to$7.08 (+0.20) & 6.29$\to$6.43 (+0.14) & 5.38$\to$5.39 (+0.01) & 5.20$\to$5.07 (-0.13) \\
Contribution & Negative & 7.72$\to$7.84 (+0.12) & 6.88$\to$7.08 (+0.20) & 6.29$\to$6.42 (+0.13) & 5.38$\to$5.53 (+0.15) & 5.20$\to$4.91 (-0.30) \\
Contribution & Positive & 7.72$\to$7.88 (+0.17) & 6.88$\to$7.21 (+0.33) & 6.29$\to$6.45 (+0.16) & 5.38$\to$5.53 (+0.14) & 5.20$\to$4.82 (-0.38) \\
Lexical & Negative & 7.72$\to$7.75 (+0.03) & 6.88$\to$6.90 (+0.03) & 6.29$\to$6.30 (+0.01) & 5.38$\to$5.45 (+0.07) & 5.20$\to$4.94 (-0.26) \\
Lexical & Positive & 7.72$\to$7.75 (+0.03) & 6.88$\to$7.03 (+0.16) & 6.29$\to$6.36 (+0.07) & 5.38$\to$5.49 (+0.11) & 5.20$\to$5.18 (-0.03) \\
\addlinespace
\rowcolor{black!6}\multicolumn{7}{l}{\textit{Standard protocol; Rewrite model: Opus 4.8}}\\
Novelty & Negative & 7.72$\to$7.24 (-0.48) & 6.88$\to$6.41 (-0.47) & 6.29$\to$5.91 (-0.38) & 5.38$\to$5.22 (-0.16) & 5.20$\to$4.71 (-0.50) \\
Novelty & Positive & 7.72$\to$7.82 (+0.10) & 6.88$\to$7.14 (+0.27) & 6.29$\to$6.67 (+0.38) & 5.38$\to$5.48 (+0.10) & 5.20$\to$5.14 (-0.06) \\
Evidence & Negative & 7.72$\to$7.38 (-0.34) & 6.88$\to$6.60 (-0.28) & 6.29$\to$5.92 (-0.37) & 5.38$\to$5.27 (-0.12) & 5.20$\to$4.75 (-0.45) \\
Evidence & Positive & 7.72$\to$7.95 (+0.23) & 6.88$\to$7.40 (+0.53) & 6.29$\to$6.96 (+0.67) & 5.38$\to$5.60 (+0.22) & 5.20$\to$5.16 (-0.04) \\
Scope & Negative & 7.72$\to$7.33 (-0.38) & 6.88$\to$6.56 (-0.32) & 6.29$\to$5.92 (-0.38) & 5.38$\to$5.36 (-0.03) & 5.20$\to$4.92 (-0.28) \\
Scope & Positive & 7.72$\to$7.88 (+0.16) & 6.88$\to$7.36 (+0.48) & 6.29$\to$6.52 (+0.22) & 5.38$\to$5.56 (+0.17) & 5.20$\to$5.17 (-0.03) \\
Technical & Negative & 7.72$\to$7.72 (0.00) & 6.88$\to$6.96 (+0.08) & 6.29$\to$6.39 (+0.10) & 5.38$\to$5.51 (+0.12) & 5.20$\to$5.06 (-0.14) \\
Technical & Positive & 7.72$\to$7.83 (+0.12) & 6.88$\to$7.26 (+0.38) & 6.29$\to$6.53 (+0.24) & 5.38$\to$5.65 (+0.27) & 5.20$\to$5.22 (+0.02) \\
Contribution & Negative & 7.72$\to$7.78 (+0.07) & 6.88$\to$6.88 (0.00) & 6.29$\to$6.48 (+0.19) & 5.38$\to$5.47 (+0.09) & 5.20$\to$5.03 (-0.17) \\
Contribution & Positive & 7.72$\to$7.95 (+0.23) & 6.88$\to$7.32 (+0.44) & 6.29$\to$6.73 (+0.44) & 5.38$\to$5.62 (+0.23) & 5.20$\to$5.04 (-0.16) \\
Lexical & Negative & 7.72$\to$7.72 (+0.01) & 6.88$\to$6.94 (+0.07) & 6.29$\to$6.49 (+0.20) & 5.38$\to$5.46 (+0.07) & 5.20$\to$5.01 (-0.19) \\
Lexical & Positive & 7.72$\to$7.79 (+0.08) & 6.88$\to$6.90 (+0.03) & 6.29$\to$6.33 (+0.03) & 5.38$\to$5.38 (0.00) & 5.20$\to$5.15 (-0.05) \\
\addlinespace
\rowcolor{black!6}\multicolumn{7}{l}{\textit{Strict protocol; Rewrite model: GPT-5.5}}\\
Novelty & Negative & 6.46$\to$6.08 (-0.38) & 4.80$\to$4.47 (-0.33) & 4.27$\to$4.08 (-0.18) & 4.64$\to$4.63 (-0.01) & 4.44$\to$3.99 (-0.45) \\
Novelty & Positive & 6.46$\to$6.80 (+0.34) & 4.80$\to$5.02 (+0.22) & 4.27$\to$4.73 (+0.47) & 4.64$\to$4.77 (+0.12) & 4.44$\to$4.43 (-0.01) \\
Evidence & Negative & 6.46$\to$5.97 (-0.49) & 4.80$\to$4.47 (-0.33) & 4.27$\to$4.00 (-0.27) & 4.64$\to$4.52 (-0.12) & 4.44$\to$4.03 (-0.42) \\
Evidence & Positive & 6.46$\to$7.08 (+0.62) & 4.80$\to$4.83 (+0.03) & 4.27$\to$4.66 (+0.39) & 4.64$\to$4.86 (+0.22) & 4.44$\to$4.15 (-0.29) \\
Scope & Negative & 6.46$\to$5.87 (-0.59) & 4.80$\to$4.45 (-0.35) & 4.27$\to$4.17 (-0.10) & 4.64$\to$4.72 (+0.08) & 4.44$\to$3.89 (-0.55) \\
Scope & Positive & 6.46$\to$6.57 (+0.11) & 4.80$\to$4.87 (+0.07) & 4.27$\to$4.59 (+0.33) & 4.64$\to$4.81 (+0.17) & 4.44$\to$4.28 (-0.16) \\
Technical & Negative & 6.46$\to$6.50 (+0.04) & 4.80$\to$4.67 (-0.13) & 4.27$\to$4.53 (+0.27) & 4.64$\to$4.72 (+0.08) & 4.44$\to$4.21 (-0.23) \\
Technical & Positive & 6.46$\to$6.65 (+0.19) & 4.80$\to$4.68 (-0.12) & 4.27$\to$4.48 (+0.22) & 4.64$\to$4.78 (+0.14) & 4.44$\to$4.31 (-0.13) \\
Contribution & Negative & 6.46$\to$6.61 (+0.15) & 4.80$\to$4.68 (-0.12) & 4.27$\to$4.58 (+0.32) & 4.64$\to$4.76 (+0.12) & 4.44$\to$4.06 (-0.38) \\
Contribution & Positive & 6.46$\to$6.94 (+0.48) & 4.80$\to$4.76 (-0.04) & 4.27$\to$4.63 (+0.37) & 4.64$\to$4.86 (+0.22) & 4.44$\to$4.20 (-0.24) \\
Lexical & Negative & 6.46$\to$6.38 (-0.07) & 4.80$\to$4.65 (-0.15) & 4.27$\to$4.33 (+0.07) & 4.64$\to$4.82 (+0.17) & 4.44$\to$4.22 (-0.22) \\
Lexical & Positive & 6.46$\to$6.60 (+0.14) & 4.80$\to$4.83 (+0.03) & 4.27$\to$4.26 (-0.01) & 4.64$\to$4.78 (+0.13) & 4.44$\to$4.30 (-0.13) \\
\addlinespace
\rowcolor{black!6}\multicolumn{7}{l}{\textit{Strict protocol; Rewrite model: Opus 4.8}}\\
Novelty & Negative & 6.46$\to$5.72 (-0.73) & 4.80$\to$4.45 (-0.35) & 4.27$\to$3.65 (-0.62) & 4.64$\to$4.54 (-0.10) & 4.44$\to$3.89 (-0.55) \\
Novelty & Positive & 6.46$\to$7.03 (+0.57) & 4.80$\to$4.89 (+0.09) & 4.27$\to$4.73 (+0.47) & 4.64$\to$4.78 (+0.13) & 4.44$\to$4.45 (+0.02) \\
Evidence & Negative & 6.46$\to$5.84 (-0.62) & 4.80$\to$4.51 (-0.29) & 4.27$\to$3.95 (-0.32) & 4.64$\to$4.57 (-0.08) & 4.44$\to$3.94 (-0.50) \\
Evidence & Positive & 6.46$\to$7.39 (+0.93) & 4.80$\to$5.33 (+0.53) & 4.27$\to$5.04 (+0.78) & 4.64$\to$5.12 (+0.48) & 4.44$\to$4.52 (+0.08) \\
Scope & Negative & 6.46$\to$5.76 (-0.70) & 4.80$\to$4.54 (-0.26) & 4.27$\to$4.05 (-0.22) & 4.64$\to$4.62 (-0.03) & 4.44$\to$4.06 (-0.38) \\
Scope & Positive & 6.46$\to$6.73 (+0.28) & 4.80$\to$4.92 (+0.12) & 4.27$\to$4.51 (+0.24) & 4.64$\to$4.76 (+0.12) & 4.44$\to$4.39 (-0.04) \\
Technical & Negative & 6.46$\to$6.58 (+0.12) & 4.80$\to$4.70 (-0.10) & 4.27$\to$4.62 (+0.36) & 4.64$\to$4.83 (+0.19) & 4.44$\to$4.23 (-0.21) \\
Technical & Positive & 6.46$\to$6.91 (+0.45) & 4.80$\to$5.01 (+0.21) & 4.27$\to$4.63 (+0.37) & 4.64$\to$4.88 (+0.24) & 4.44$\to$4.38 (-0.06) \\
Contribution & Negative & 6.46$\to$6.47 (+0.01) & 4.80$\to$4.79 (-0.01) & 4.27$\to$4.62 (+0.36) & 4.64$\to$4.84 (+0.20) & 4.44$\to$4.27 (-0.17) \\
Contribution & Positive & 6.46$\to$7.09 (+0.63) & 4.80$\to$5.03 (+0.23) & 4.27$\to$4.62 (+0.35) & 4.64$\to$4.92 (+0.28) & 4.44$\to$4.34 (-0.10) \\
Lexical & Negative & 6.46$\to$6.56 (+0.10) & 4.80$\to$4.66 (-0.14) & 4.27$\to$4.52 (+0.25) & 4.64$\to$4.84 (+0.20) & 4.44$\to$4.33 (-0.11) \\
Lexical & Positive & 6.46$\to$6.45 (-0.01) & 4.80$\to$4.72 (-0.08) & 4.27$\to$4.27 (0.00) & 4.64$\to$4.62 (-0.03) & 4.44$\to$4.32 (-0.12) \\
\addlinespace
\end{longtable}
\endgroup

\subsection{Fixed-Model Directional Heterogeneity}
\label{app:fixed_model_single}

\begin{figure}[!htbp]
  \centering
  \includegraphics[width=0.98\textwidth]{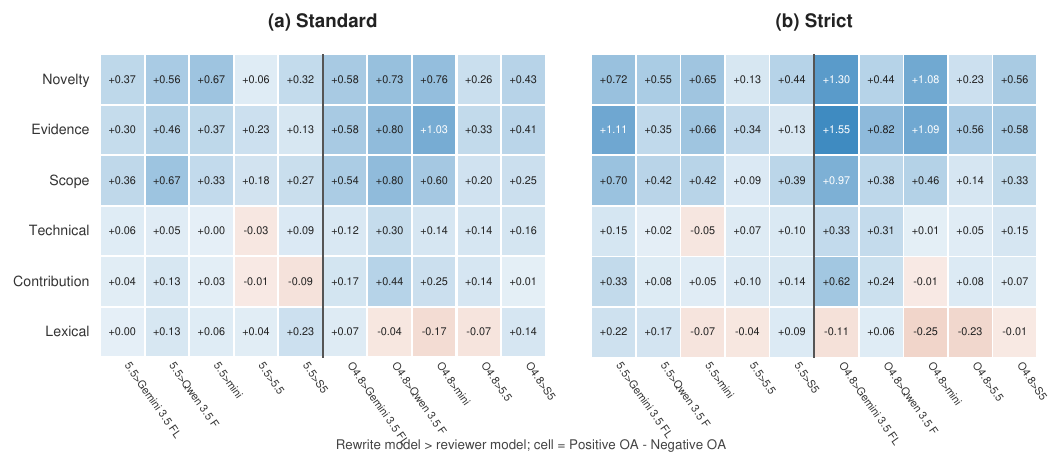}
  \caption{Fixed-model heterogeneity in the directional response to rhetorical
  rewriting. Each cell reports the mean overall-rating difference between the
  positive and negative conditions for one pairing of a rewrite model and a
  reviewer model; blue denotes the intended ordering and orange the reverse.
  Table~\ref{tab:dimension_reviewer_rating_matrix} reports the corresponding
  original and rewrite means. ``Gemini 3.5 FL,'' ``Qwen 3.5 F,'' ``mini,''
  ``5.5,'' and ``S5'' abbreviate Gemini~3.5 FL, Qwen~3.5 F,
  GPT-5 mini, GPT-5.5, and Sonnet~5.}
  \label{fig:producer_evaluator_heatmap}
\end{figure}

Figure~\ref{fig:producer_evaluator_heatmap} summarizes the directional
ordering across the ten fixed pairings of a rewrite model and a reviewer model.
Novelty, evidence, and scope retain the intended ordering most consistently;
technical register, contribution structure, and lexical complexity are
smaller or less stable.

\subsection{Secondary AI-Review Score Spillovers}
\label{app:secondary_scores}

Table~\ref{tab:dimension_secondary_combined} combines the presentation,
contribution, and soundness results referenced in
Section~\ref{subsec:dimension_rating_changes}. Presentation is most proximal
to the intervention. Contribution and soundness are diagnostic judgments
emitted by the AI reviewer. Because the rewrites are constrained to preserve
the underlying methods and reported values while allowing result-presentation prose to change,
these score changes are treated as reviewer-response spillovers rather than
independently verified changes in scientific merit.

\begingroup
\scriptsize
\setlength{\tabcolsep}{2.8pt}
\begin{longtable}{llccc}
\caption{Pooled secondary-score responses across both rewrite models and all five reviewer models. Each cell reports original, rewrite (change).}\phantomsection\label{tab:dimension_secondary_combined}\\
\toprule
Dimension & Direction & Presentation & Contribution & Soundness \\
\midrule
\endfirsthead
\toprule
Dimension & Direction & Presentation & Contribution & Soundness \\
\midrule
\endhead
\midrule
\endfoot
\bottomrule
\endlastfoot
\multicolumn{5}{l}{\textit{Standard evaluation protocol}}\\
Novelty & Negative & 3.301, 3.292 (-0.009) & 3.184, 3.041 (-0.143) & 3.118, 3.017 (-0.101) \\
Novelty & Positive & 3.300, 3.362 (+0.062) & 3.183, 3.261 (+0.077) & 3.118, 3.177 (+0.059) \\
Evidence & Negative & 3.301, 3.322 (+0.021) & 3.184, 3.038 (-0.146) & 3.118, 3.051 (-0.067) \\
Evidence & Positive & 3.302, 3.345 (+0.043) & 3.184, 3.256 (+0.072) & 3.118, 3.206 (+0.088) \\
Scope & Negative & 3.301, 3.341 (+0.040) & 3.184, 3.024 (-0.159) & 3.118, 3.116 (-0.002) \\
Scope & Positive & 3.301, 3.349 (+0.048) & 3.184, 3.230 (+0.047) & 3.118, 3.187 (+0.069) \\
Technical & Negative & 3.301, 3.307 (+0.006) & 3.184, 3.205 (+0.022) & 3.118, 3.142 (+0.024) \\
Technical & Positive & 3.301, 3.370 (+0.069) & 3.184, 3.240 (+0.056) & 3.118, 3.185 (+0.067) \\
Contribution & Negative & 3.301, 3.336 (+0.035) & 3.184, 3.183 (-0.001) & 3.118, 3.177 (+0.059) \\
Contribution & Positive & 3.300, 3.370 (+0.070) & 3.183, 3.245 (+0.062) & 3.118, 3.191 (+0.073) \\
Lexical & Negative & 3.301, 3.320 (+0.019) & 3.184, 3.191 (+0.007) & 3.118, 3.132 (+0.013) \\
Lexical & Positive & 3.302, 3.255 (-0.048) & 3.184, 3.203 (+0.020) & 3.118, 3.153 (+0.035) \\
\addlinespace
\multicolumn{5}{l}{\textit{Strict evaluation protocol}}\\
Novelty & Negative & 3.255, 3.249 (-0.006) & 2.685, 2.539 (-0.146) & 2.810, 2.721 (-0.089) \\
Novelty & Positive & 3.255, 3.310 (+0.055) & 2.684, 2.799 (+0.114) & 2.809, 2.920 (+0.111) \\
Evidence & Negative & 3.255, 3.256 (+0.001) & 2.684, 2.550 (-0.134) & 2.809, 2.738 (-0.071) \\
Evidence & Positive & 3.255, 3.305 (+0.049) & 2.685, 2.842 (+0.158) & 2.810, 2.973 (+0.163) \\
Scope & Negative & 3.255, 3.286 (+0.031) & 2.684, 2.551 (-0.134) & 2.809, 2.814 (+0.005) \\
Scope & Positive & 3.255, 3.282 (+0.027) & 2.684, 2.737 (+0.053) & 2.809, 2.890 (+0.081) \\
Technical & Negative & 3.255, 3.258 (+0.003) & 2.684, 2.709 (+0.025) & 2.809, 2.837 (+0.028) \\
Technical & Positive & 3.255, 3.301 (+0.046) & 2.684, 2.747 (+0.063) & 2.809, 2.901 (+0.092) \\
Contribution & Negative & 3.255, 3.278 (+0.023) & 2.683, 2.706 (+0.022) & 2.808, 2.883 (+0.075) \\
Contribution & Positive & 3.255, 3.308 (+0.052) & 2.685, 2.796 (+0.111) & 2.810, 2.924 (+0.114) \\
Lexical & Negative & 3.255, 3.252 (-0.003) & 2.683, 2.705 (+0.022) & 2.808, 2.852 (+0.043) \\
Lexical & Positive & 3.255, 3.221 (-0.034) & 2.685, 2.728 (+0.043) & 2.810, 2.839 (+0.030) \\
\addlinespace
\end{longtable}
\endgroup

\subsection{Paper-Level Directional Consistency}
\label{app:paper_consistency}

For each paper and dimension, the directional effect compares the paper-level
mean rating under the positive condition with the corresponding mean under
the negative condition after averaging both rewrite models, both evaluation
protocols, and all five reviewer models. A Win is positive, a Tie is zero, and
a Loss is negative. This analysis checks whether the aggregate directional
results are broadly shared or driven by a few extreme manuscripts.

\begin{table}[!htbp]
  \centering
  \caption{Paper-level consistency of the directional rewriting effect after averaging both rewriters, both prompts, and all five reviewers within paper.}
  \label{tab:paper_level_consistency}
  \small
  \begin{tabular}{lrrrr}
    \toprule
    Dimension & Mean difference & Win & Tie & Loss \\
    \midrule
    Novelty & $+0.538$ & 117 & 1 & 2 \\
    Evidence & $+0.593$ & 116 & 1 & 3 \\
    Scope & $+0.425$ & 113 & 2 & 5 \\
    Technical & $+0.108$ & 73 & 7 & 40 \\
    Contribution & $+0.142$ & 79 & 4 & 37 \\
    Lexical & $+0.011$ & 62 & 7 & 51 \\
    \bottomrule
  \end{tabular}
\end{table}

Novelty, evidence, and scope show the intended ordering for more than one
hundred of the 120 papers. Technical register and contribution structure are
less consistent, while lexical complexity is nearly balanced (62 wins, 7
ties, and 51 losses).
The dominant directional findings are therefore distributed across papers
rather than generated by a small set of influential cases.

\subsection{Weak-Accept Decision Effects}
\label{app:threshold_flips}

\subsubsection{Panel and Configuration Summaries}

Table~\ref{tab:dimension_threshold_summary_main} reports the complete
configuration-level percentage-point changes summarized in
the decision-level check in Section~\ref{sec:dimension_effects}.

\newpage
\begin{table}[!t]
  \centering
  \caption{\textbf{Changes in weak-accept probability by rewrite model, reviewer model, and
  rhetorical dimension.} Entries are percentage-point changes relative to
  the prompt-matched original in weak-accept probability.
  Pos. and Neg. denote rewrite directions. The two rightmost columns average
  the six dimensions within each row. Rewriter means average reviewers within
  a prompt; prompt means average both rewriters and all reviewers; the final
  row averages all configurations. Green and yellow denote increases and
  decreases, with color intensity proportional to the absolute change; zero
  changes are unshaded. Complete probabilities,
  transition counts, and confidence intervals are reported in
  Appendix~\ref{app:threshold_flips}.}
  \label{tab:dimension_threshold_summary_main}
  \scriptsize
  \renewcommand{\arraystretch}{1.00}
  \setlength{\tabcolsep}{1.8pt}
  \newcommand{\thresholdincrease}[2]{\cellcolor{green!#1}#2}
  \newcommand{\thresholddecrease}[2]{\cellcolor{yellow!#1}#2}
  \scalebox{0.90}{%
  \begin{tabular*}{\textwidth}{@{\extracolsep{\fill}}ll*{14}{r}@{}}
    \toprule
    \multirow{2}{*}{\textbf{Rewriter}}
      & \multirow{2}{*}{\textbf{Reviewer}}
      & \multicolumn{2}{c}{\textbf{Novelty}}
      & \multicolumn{2}{c}{\textbf{Evidence}}
      & \multicolumn{2}{c}{\textbf{Scope}}
      & \multicolumn{2}{c}{\textbf{Technical}}
      & \multicolumn{2}{c}{\textbf{Contribution}}
      & \multicolumn{2}{c}{\textbf{Lexical}}
      & \multicolumn{2}{c}{\textbf{Across dimensions}} \\
    \cmidrule(lr){3-4}\cmidrule(lr){5-6}\cmidrule(lr){7-8}
    \cmidrule(lr){9-10}\cmidrule(lr){11-12}\cmidrule(lr){13-14}
    \cmidrule(l){15-16}
      & & \multicolumn{1}{c}{\textbf{Pos.}}
      & \multicolumn{1}{c}{\textbf{Neg.}}
      & \multicolumn{1}{c}{\textbf{Pos.}}
      & \multicolumn{1}{c}{\textbf{Neg.}}
      & \multicolumn{1}{c}{\textbf{Pos.}}
      & \multicolumn{1}{c}{\textbf{Neg.}}
      & \multicolumn{1}{c}{\textbf{Pos.}}
      & \multicolumn{1}{c}{\textbf{Neg.}}
      & \multicolumn{1}{c}{\textbf{Pos.}}
      & \multicolumn{1}{c}{\textbf{Neg.}}
      & \multicolumn{1}{c}{\textbf{Pos.}}
      & \multicolumn{1}{c}{\textbf{Neg.}}
      & \multicolumn{1}{c}{\textbf{Pos.}}
      & \multicolumn{1}{c}{\textbf{Neg.}} \\
    \midrule
    \multicolumn{16}{l}{\textit{Standard review prompt}} \\
    \cmidrule(lr){1-16}
    GPT-5.5 & Gemini 3.5 FL & \thresholdincrease{13}{+1.7} & 0.0 & 0.0 & 0.0 & \thresholdincrease{9}{+0.8} & 0.0 & \thresholdincrease{9}{+0.8} & \thresholdincrease{13}{+1.7} & \thresholdincrease{13}{+1.7} & \thresholdincrease{9}{+0.8} & 0.0 & \thresholdincrease{13}{+1.7} & +0.8 & +0.7 \\
     & Qwen 3.5 F & \thresholdincrease{13}{+1.7} & \thresholddecrease{21}{-2.5} & \thresholdincrease{16}{+2.5} & \thresholdincrease{18}{+3.3} & \thresholdincrease{21}{+4.2} & \thresholddecrease{24}{-3.3} & \thresholdincrease{13}{+1.7} & \thresholdincrease{18}{+3.3} & \thresholdincrease{26}{+6.7} & \thresholdincrease{18}{+3.3} & \thresholdincrease{9}{+0.8} & \thresholdincrease{9}{+0.8} & +2.9 & +0.8 \\
     & GPT-5 mini & \thresholdincrease{28}{+7.5} & \thresholddecrease{27}{-4.2} & \thresholdincrease{29}{+8.3} & \thresholdincrease{18}{+3.3} & \thresholdincrease{21}{+4.2} & \thresholdincrease{18}{+3.3} & \thresholdincrease{16}{+2.5} & \thresholdincrease{9}{+0.8} & \thresholdincrease{24}{+5.8} & \thresholdincrease{29}{+8.3} & \thresholdincrease{18}{+3.3} & \thresholdincrease{16}{+2.5} & +5.3 & +2.4 \\
     & GPT-5.5 & \thresholdincrease{33}{+10.8} & 0.0 & \thresholdincrease{33}{+10.8} & \thresholddecrease{21}{-2.5} & \thresholdincrease{28}{+7.5} & \thresholdincrease{16}{+2.5} & \thresholdincrease{24}{+5.8} & \thresholdincrease{24}{+5.8} & \thresholdincrease{30}{+9.2} & \thresholdincrease{34}{+11.7} & \thresholdincrease{21}{+4.2} & \thresholdincrease{26}{+6.7} & +8.1 & +4.0 \\
     & Sonnet 5 & \thresholddecrease{30}{-5.0} & \thresholddecrease{54}{-16.0} & \thresholddecrease{54}{-16.0} & \thresholddecrease{59}{-19.2} & \thresholddecrease{35}{-6.7} & \thresholddecrease{60}{-23.5} & \thresholddecrease{45}{-11.0} & \thresholddecrease{32}{-5.8} & \thresholddecrease{60}{-20.2} & \thresholddecrease{54}{-16.2} & \thresholddecrease{37}{-7.6} & \thresholddecrease{44}{-10.8} & -11.1 & -15.3 \\
    \cmidrule(lr){2-16}
    & \textit{GPT-5.5 mean} & +3.3 & -4.5 & +1.1 & -3.0 & +2.0 & -4.2 & 0.0 & +1.2 & +0.6 & +1.6 & +0.2 & +0.2 & +1.2 & -1.5 \\
    \cmidrule(lr){1-16}
    Opus 4.8 & Gemini 3.5 FL & \thresholdincrease{13}{+1.7} & \thresholddecrease{27}{-4.2} & \thresholdincrease{13}{+1.7} & \thresholddecrease{12}{-0.8} & \thresholdincrease{9}{+0.8} & 0.0 & 0.0 & 0.0 & \thresholdincrease{13}{+1.7} & 0.0 & \thresholdincrease{9}{+0.8} & \thresholddecrease{12}{-0.8} & +1.1 & -1.0 \\
     & Qwen 3.5 F & \thresholdincrease{18}{+3.3} & \thresholddecrease{46}{-11.7} & \thresholdincrease{21}{+4.2} & \thresholddecrease{37}{-7.5} & \thresholdincrease{26}{+6.7} & \thresholddecrease{39}{-8.3} & \thresholdincrease{22}{+5.0} & \thresholdincrease{13}{+1.7} & \thresholdincrease{24}{+5.8} & \thresholdincrease{22}{+5.0} & \thresholddecrease{12}{-0.8} & \thresholddecrease{24}{-3.3} & +4.0 & -4.0 \\
     & GPT-5 mini & \thresholdincrease{16}{+2.5} & \thresholddecrease{46}{-11.7} & \thresholdincrease{26}{+6.7} & \thresholddecrease{35}{-6.7} & \thresholdincrease{24}{+5.8} & \thresholddecrease{27}{-4.2} & \thresholdincrease{24}{+5.8} & \thresholdincrease{18}{+3.3} & \thresholdincrease{24}{+5.8} & \thresholdincrease{16}{+2.5} & 0.0 & \thresholdincrease{18}{+3.3} & +4.4 & -2.2 \\
     & GPT-5.5 & \thresholdincrease{26}{+6.7} & \thresholddecrease{27}{-4.2} & \thresholdincrease{45}{+20.0} & \thresholddecrease{24}{-3.3} & \thresholdincrease{30}{+9.2} & \thresholddecrease{12}{-0.8} & \thresholdincrease{34}{+11.7} & \thresholdincrease{28}{+7.5} & \thresholdincrease{32}{+10.0} & \thresholdincrease{33}{+10.8} & 0.0 & \thresholdincrease{30}{+9.2} & +9.6 & +3.2 \\
     & Sonnet 5 & \thresholdincrease{9}{+0.8} & \thresholddecrease{59}{-19.3} & \thresholddecrease{12}{-0.8} & \thresholddecrease{59}{-19.5} & \thresholddecrease{30}{-5.1} & \thresholddecrease{54}{-16.1} & \thresholdincrease{13}{+1.7} & \thresholddecrease{28}{-4.2} & \thresholddecrease{30}{-5.0} & \thresholddecrease{35}{-6.8} & 0.0 & \thresholddecrease{33}{-5.9} & -1.4 & -12.0 \\
    \cmidrule(lr){2-16}
    & \textit{Opus 4.8 mean} & +3.0 & -10.2 & +6.3 & -7.6 & +3.5 & -5.9 & +4.8 & +1.7 & +3.7 & +2.3 & 0.0 & +0.5 & +3.6 & -3.2 \\
    \cmidrule(lr){1-16}
    \multicolumn{2}{l}{\textit{Standard prompt mean}} & +3.2 & -7.4 & +3.7 & -5.3 & +2.7 & -5.0 & +2.4 & +1.4 & +2.1 & +1.9 & +0.1 & +0.3 & +2.4 & -2.3 \\
    \midrule
    \multicolumn{16}{l}{\textit{Strict review prompt}} \\
    \cmidrule(lr){1-16}
    GPT-5.5 & Gemini 3.5 FL & \thresholdincrease{24}{+5.8} & \thresholddecrease{21}{-2.5} & \thresholdincrease{29}{+8.3} & \thresholddecrease{32}{-5.8} & \thresholdincrease{16}{+2.5} & \thresholddecrease{50}{-14.2} & \thresholdincrease{21}{+4.2} & \thresholdincrease{16}{+2.5} & \thresholdincrease{22}{+5.0} & \thresholdincrease{29}{+8.3} & \thresholdincrease{16}{+2.5} & \thresholddecrease{12}{-0.8} & +4.7 & -2.1 \\
     & Qwen 3.5 F & \thresholdincrease{32}{+10.0} & \thresholddecrease{35}{-6.7} & \thresholdincrease{28}{+7.5} & \thresholddecrease{27}{-4.2} & \thresholdincrease{13}{+1.7} & \thresholddecrease{35}{-6.7} & 0.0 & 0.0 & \thresholdincrease{9}{+0.8} & 0.0 & \thresholdincrease{9}{+0.8} & \thresholddecrease{17}{-1.7} & +3.5 & -3.2 \\
     & GPT-5 mini & \thresholdincrease{32}{+10.0} & \thresholddecrease{35}{-6.7} & \thresholdincrease{33}{+10.8} & \thresholddecrease{39}{-8.3} & \thresholdincrease{21}{+4.2} & \thresholddecrease{35}{-6.7} & \thresholdincrease{22}{+5.0} & \thresholdincrease{18}{+3.3} & \thresholdincrease{26}{+6.7} & \thresholdincrease{22}{+5.0} & \thresholdincrease{9}{+0.8} & \thresholdincrease{18}{+3.3} & +6.2 & -1.7 \\
     & GPT-5.5 & \thresholdincrease{16}{+2.5} & \thresholdincrease{9}{+0.8} & \thresholdincrease{29}{+8.3} & \thresholddecrease{21}{-2.5} & \thresholdincrease{18}{+3.3} & \thresholdincrease{16}{+2.5} & \thresholdincrease{9}{+0.8} & \thresholdincrease{16}{+2.5} & \thresholdincrease{29}{+8.3} & \thresholdincrease{18}{+3.3} & 0.0 & \thresholdincrease{21}{+4.2} & +3.9 & +1.8 \\
     & Sonnet 5 & \thresholddecrease{21}{-2.5} & \thresholddecrease{39}{-8.5} & \thresholddecrease{35}{-6.7} & \thresholddecrease{35}{-6.7} & \thresholddecrease{32}{-5.8} & \thresholddecrease{35}{-6.7} & \thresholddecrease{28}{-4.2} & \thresholddecrease{30}{-5.0} & \thresholddecrease{21}{-2.6} & \thresholddecrease{30}{-5.1} & 0.0 & \thresholddecrease{32}{-5.8} & -3.6 & -6.3 \\
    \cmidrule(lr){2-16}
    & \textit{GPT-5.5 mean} & +5.2 & -4.7 & +5.7 & -5.5 & +1.2 & -6.3 & +1.2 & +0.7 & +3.7 & +2.3 & +0.8 & -0.2 & +2.9 & -2.3 \\
    \cmidrule(lr){1-16}
    Opus 4.8 & Gemini 3.5 FL & \thresholdincrease{26}{+6.7} & \thresholddecrease{49}{-13.3} & \thresholdincrease{34}{+11.7} & \thresholddecrease{49}{-13.3} & \thresholdincrease{21}{+4.2} & \thresholddecrease{57}{-18.3} & \thresholdincrease{29}{+8.3} & \thresholdincrease{21}{+4.2} & \thresholdincrease{34}{+11.7} & \thresholdincrease{9}{+0.8} & \thresholdincrease{9}{+0.8} & \thresholdincrease{18}{+3.3} & +7.2 & -6.1 \\
     & Qwen 3.5 F & \thresholdincrease{28}{+7.5} & \thresholddecrease{35}{-6.7} & \thresholdincrease{45}{+22.5} & \thresholddecrease{32}{-5.8} & \thresholdincrease{29}{+8.3} & \thresholddecrease{27}{-4.2} & \thresholdincrease{36}{+12.5} & \thresholdincrease{13}{+1.7} & \thresholdincrease{26}{+6.7} & \thresholdincrease{16}{+2.5} & 0.0 & \thresholdincrease{16}{+2.5} & +9.6 & -1.7 \\
     & GPT-5 mini & \thresholdincrease{39}{+15.0} & \thresholddecrease{46}{-11.7} & \thresholdincrease{45}{+20.8} & \thresholddecrease{39}{-8.3} & \thresholdincrease{24}{+5.8} & \thresholddecrease{35}{-6.7} & \thresholdincrease{26}{+6.7} & \thresholdincrease{24}{+5.8} & \thresholdincrease{29}{+8.3} & \thresholdincrease{28}{+7.5} & \thresholddecrease{17}{-1.7} & \thresholdincrease{22}{+5.0} & +9.2 & -1.4 \\
     & GPT-5.5 & \thresholdincrease{26}{+6.7} & \thresholddecrease{35}{-6.7} & \thresholdincrease{41}{+16.7} & \thresholddecrease{37}{-7.5} & \thresholdincrease{18}{+3.3} & \thresholddecrease{12}{-0.8} & \thresholdincrease{24}{+5.8} & \thresholdincrease{21}{+4.2} & \thresholdincrease{32}{+10.0} & \thresholdincrease{29}{+8.3} & \thresholdincrease{9}{+0.8} & \thresholdincrease{29}{+8.3} & +7.2 & +1.0 \\
     & Sonnet 5 & \thresholdincrease{18}{+3.4} & \thresholddecrease{37}{-7.6} & 0.0 & \thresholddecrease{37}{-7.6} & \thresholdincrease{21}{+4.2} & \thresholddecrease{35}{-6.9} & \thresholdincrease{9}{+0.8} & \thresholddecrease{21}{-2.6} & \thresholddecrease{17}{-1.7} & \thresholddecrease{21}{-2.5} & 0.0 & \thresholddecrease{21}{-2.5} & +1.1 & -4.9 \\
    \cmidrule(lr){2-16}
    & \textit{Opus 4.8 mean} & +7.8 & -9.2 & +14.3 & -8.5 & +5.2 & -7.4 & +6.8 & +2.7 & +7.0 & +3.3 & 0.0 & +3.3 & +6.9 & -2.6 \\
    \cmidrule(lr){1-16}
    \multicolumn{2}{l}{\textit{Strict prompt mean}} & +6.5 & -7.0 & +10.0 & -7.0 & +3.2 & -6.9 & +4.0 & +1.7 & +5.3 & +2.8 & +0.4 & +1.6 & +4.9 & -2.5 \\
    \midrule
    \multicolumn{2}{l}{\textbf{Overall mean}} & \textbf{+4.8} & \textbf{-7.2} & \textbf{+6.9} & \textbf{-6.1} & \textbf{+3.0} & \textbf{-6.0} & \textbf{+3.2} & \textbf{+1.5} & \textbf{+3.7} & \textbf{+2.4} & \textbf{+0.2} & \textbf{+1.0} & \textbf{+3.6} & \textbf{-2.4} \\
    \bottomrule
  \end{tabular*}
  }
\end{table}

Table~\ref{tab:dimension_panel_threshold_combined} reports the corresponding
paper-level estimates and confidence intervals. As in the rating analysis,
the first result column pools the two rewrite models and the remaining
columns hold the rewrite model fixed.

\begingroup
\scriptsize
\setlength{\tabcolsep}{3.0pt}
\begin{longtable}{llccc}
\caption{Panel-aligned percentage-point changes in weak-accept probability. Entries are mean paper-level changes with 95\% paper-bootstrap confidence intervals; all five reviewer models are averaged within paper.}\phantomsection\label{tab:dimension_panel_threshold_combined}\\
\toprule
Dimension & Direction & All Rewrite models & GPT-5.5 Rewrite & Opus 4.8 Rewrite \\
\midrule
\endfirsthead
\toprule
Dimension & Direction & All Rewrite models & GPT-5.5 Rewrite & Opus 4.8 Rewrite \\
\midrule
\endhead
\midrule
\endfoot
\bottomrule
\endlastfoot
\multicolumn{5}{l}{\textit{Standard evaluation protocol}}\\
Novelty & Negative & $-7.39\;[-9.83,-4.89]$ & $-4.67\;[-7.67,-1.67]$ & $-10.12\;[-13.13,-7.12]$ \\
Novelty & Positive & $+3.08\;[+0.58,+5.67]$ & $+3.33\;[+0.33,+6.33]$ & $+2.83\;[+0.17,+5.50]$ \\
Evidence & Negative & $-5.22\;[-7.58,-2.78]$ & $-3.00\;[-5.83,-0.33]$ & $-7.46\;[-10.29,-4.58]$ \\
Evidence & Positive & $+3.83\;[+1.42,+6.42]$ & $+1.25\;[-1.50,+4.17]$ & $+6.42\;[+3.75,+9.17]$ \\
Scope & Negative & $-4.97\;[-7.58,-2.39]$ & $-4.17\;[-7.00,-1.33]$ & $-5.79\;[-8.96,-2.67]$ \\
Scope & Positive & $+2.77\;[+0.54,+5.10]$ & $+2.00\;[-0.67,+4.67]$ & $+3.54\;[+0.88,+6.29]$ \\
Technical & Negative & $+1.45\;[-0.83,+3.83]$ & $+1.17\;[-1.33,+3.83]$ & $+1.75\;[-1.00,+4.50]$ \\
Technical & Positive & $+2.34\;[-0.08,+5.01]$ & $-0.17\;[-3.17,+2.83]$ & $+4.88\;[+2.33,+7.67]$ \\
Contribution & Negative & $+1.86\;[-0.67,+4.42]$ & $+1.71\;[-1.00,+4.54]$ & $+2.00\;[-0.67,+4.83]$ \\
Contribution & Positive & $+2.08\;[-0.42,+4.75]$ & $+0.67\;[-2.17,+3.67]$ & $+3.50\;[+0.67,+6.50]$ \\
Lexical & Negative & $+0.35\;[-1.92,+2.79]$ & $+0.17\;[-2.50,+2.83]$ & $+0.58\;[-2.08,+3.33]$ \\
Lexical & Positive & $+0.10\;[-2.15,+2.46]$ & $+0.21\;[-2.67,+3.04]$ & $0.00\;[-2.50,+2.67]$ \\
\addlinespace
\multicolumn{5}{l}{\textit{Strict evaluation protocol}}\\
Novelty & Negative & $-7.06\;[-9.92,-4.17]$ & $-4.79\;[-7.96,-1.58]$ & $-9.33\;[-12.33,-6.17]$ \\
Novelty & Positive & $+6.53\;[+3.83,+9.17]$ & $+5.17\;[+2.00,+8.33]$ & $+7.92\;[+4.83,+11.08]$ \\
Evidence & Negative & $-6.98\;[-9.92,-4.06]$ & $-5.50\;[-8.83,-2.17]$ & $-8.46\;[-11.67,-5.29]$ \\
Evidence & Positive & $+10.10\;[+7.14,+13.02]$ & $+5.75\;[+2.25,+9.08]$ & $+14.46\;[+11.12,+17.88]$ \\
Scope & Negative & $-6.81\;[-9.73,-3.94]$ & $-6.33\;[-9.67,-3.17]$ & $-7.29\;[-10.46,-4.04]$ \\
Scope & Positive & $+3.19\;[+0.72,+5.58]$ & $+1.17\;[-1.83,+4.17]$ & $+5.21\;[+2.38,+8.08]$ \\
Technical & Negative & $+1.69\;[-0.95,+4.36]$ & $+0.67\;[-2.00,+3.33]$ & $+2.75\;[-0.75,+6.25]$ \\
Technical & Positive & $+4.04\;[+1.24,+6.91]$ & $+1.17\;[-2.00,+4.33]$ & $+6.92\;[+3.67,+10.33]$ \\
Contribution & Negative & $+2.75\;[+0.08,+5.50]$ & $+2.17\;[-1.17,+5.50]$ & $+3.33\;[+0.17,+6.50]$ \\
Contribution & Positive & $+5.21\;[+2.50,+8.04]$ & $+3.38\;[-0.12,+7.00]$ & $+7.04\;[+4.17,+9.92]$ \\
Lexical & Negative & $+1.58\;[-0.83,+4.08]$ & $-0.17\;[-3.17,+2.83]$ & $+3.33\;[+0.50,+6.33]$ \\
Lexical & Positive & $+0.44\;[-2.06,+2.92]$ & $+0.83\;[-1.67,+3.33]$ & $+0.04\;[-3.17,+3.21]$ \\
\addlinespace
\end{longtable}
\endgroup

\clearpage
\subsubsection{Full Fixed-Model Rates and Transition Counts}

Table~\ref{tab:dimension_reviewer_threshold_matrix} reports the original and
rewrite weak-accept rates for every fixed pairing of a rewrite model and a
reviewer model. Table~\ref{tab:dimension_threshold_counts_combined} then retains the
pooled numbers and rates of upward and downward crossings. Reporting both
directions is important because they can cancel in the signed net change.

\begingroup
\tiny
\setlength{\tabcolsep}{0.8pt}
\begin{longtable}{llccccc}
\caption{Weak-accept probabilities by fixed rewrite model and reviewer model. Each cell reports original to rewrite (change in percentage points).}\phantomsection\label{tab:dimension_reviewer_threshold_matrix}\\
\toprule
Dimension & Direction & \shortstack{Gemini 3.5\\FL} & \shortstack{Qwen 3.5\\F} & GPT-5 mini & GPT-5.5 & Sonnet 5 \\
\midrule
\endfirsthead
\toprule
Dimension & Direction & \shortstack{Gemini 3.5\\FL} & \shortstack{Qwen 3.5\\F} & GPT-5 mini & GPT-5.5 & Sonnet 5 \\
\midrule
\endhead
\midrule
\endfoot
\bottomrule
\endlastfoot
\rowcolor{black!6}\multicolumn{7}{l}{\textit{Standard protocol; Rewrite model: GPT-5.5}}\\
Novelty & Negative & 98$\to$98 (0.0 pp) & 89$\to$87 (-2.5 pp) & 89$\to$85 (-4.2 pp) & 50$\to$50 (0.0 pp) & 43$\to$27 (-16.7 pp) \\
Novelty & Positive & 98$\to$100 (+1.7 pp) & 89$\to$91 (+1.7 pp) & 89$\to$97 (+7.5 pp) & 50$\to$61 (+10.8 pp) & 43$\to$38 (-5.0 pp) \\
Evidence & Negative & 98$\to$98 (0.0 pp) & 89$\to$92 (+3.3 pp) & 89$\to$92 (+3.3 pp) & 50$\to$48 (-2.5 pp) & 43$\to$24 (-19.2 pp) \\
Evidence & Positive & 98$\to$98 (0.0 pp) & 89$\to$92 (+2.5 pp) & 89$\to$98 (+8.3 pp) & 50$\to$61 (+10.8 pp) & 44$\to$28 (-16.0 pp) \\
Scope & Negative & 98$\to$98 (0.0 pp) & 89$\to$86 (-3.3 pp) & 89$\to$92 (+3.3 pp) & 50$\to$52 (+2.5 pp) & 43$\to$20 (-23.3 pp) \\
Scope & Positive & 98$\to$99 (+0.8 pp) & 89$\to$93 (+4.2 pp) & 89$\to$93 (+4.2 pp) & 50$\to$57 (+7.5 pp) & 43$\to$37 (-6.7 pp) \\
Technical & Negative & 98$\to$100 (+1.7 pp) & 89$\to$92 (+3.3 pp) & 89$\to$90 (+0.8 pp) & 50$\to$56 (+5.8 pp) & 43$\to$38 (-5.8 pp) \\
Technical & Positive & 98$\to$99 (+0.8 pp) & 89$\to$91 (+1.7 pp) & 89$\to$92 (+2.5 pp) & 50$\to$56 (+5.8 pp) & 43$\to$32 (-11.7 pp) \\
Contribution & Negative & 98$\to$99 (+0.8 pp) & 89$\to$92 (+3.3 pp) & 89$\to$98 (+8.3 pp) & 50$\to$62 (+11.7 pp) & 44$\to$28 (-16.0 pp) \\
Contribution & Positive & 98$\to$100 (+1.7 pp) & 89$\to$96 (+6.7 pp) & 89$\to$95 (+5.8 pp) & 50$\to$59 (+9.2 pp) & 43$\to$23 (-20.0 pp) \\
Lexical & Negative & 98$\to$100 (+1.7 pp) & 89$\to$90 (+0.8 pp) & 89$\to$92 (+2.5 pp) & 50$\to$57 (+6.7 pp) & 43$\to$32 (-10.8 pp) \\
Lexical & Positive & 98$\to$98 (0.0 pp) & 89$\to$90 (+0.8 pp) & 89$\to$92 (+3.3 pp) & 50$\to$54 (+4.2 pp) & 44$\to$36 (-7.6 pp) \\
\addlinespace
\rowcolor{black!6}\multicolumn{7}{l}{\textit{Standard protocol; Rewrite model: Opus 4.8}}\\
Novelty & Negative & 98$\to$94 (-4.2 pp) & 89$\to$78 (-11.7 pp) & 89$\to$78 (-11.7 pp) & 50$\to$46 (-4.2 pp) & 44$\to$24 (-19.3 pp) \\
Novelty & Positive & 98$\to$100 (+1.7 pp) & 89$\to$92 (+3.3 pp) & 89$\to$92 (+2.5 pp) & 50$\to$57 (+6.7 pp) & 43$\to$43 (0.0 pp) \\
Evidence & Negative & 98$\to$98 (-0.8 pp) & 89$\to$82 (-7.5 pp) & 89$\to$82 (-6.7 pp) & 50$\to$47 (-3.3 pp) & 44$\to$24 (-19.3 pp) \\
Evidence & Positive & 98$\to$100 (+1.7 pp) & 89$\to$93 (+4.2 pp) & 89$\to$96 (+6.7 pp) & 50$\to$70 (+20.0 pp) & 44$\to$43 (-0.8 pp) \\
Scope & Negative & 98$\to$98 (0.0 pp) & 89$\to$81 (-8.3 pp) & 89$\to$85 (-4.2 pp) & 50$\to$49 (-0.8 pp) & 44$\to$28 (-16.0 pp) \\
Scope & Positive & 98$\to$99 (+0.8 pp) & 89$\to$96 (+6.7 pp) & 89$\to$95 (+5.8 pp) & 50$\to$59 (+9.2 pp) & 44$\to$39 (-5.0 pp) \\
Technical & Negative & 98$\to$98 (0.0 pp) & 89$\to$91 (+1.7 pp) & 89$\to$92 (+3.3 pp) & 50$\to$57 (+7.5 pp) & 44$\to$39 (-4.2 pp) \\
Technical & Positive & 98$\to$98 (0.0 pp) & 89$\to$94 (+5.0 pp) & 89$\to$95 (+5.8 pp) & 50$\to$62 (+11.7 pp) & 44$\to$45 (+1.7 pp) \\
Contribution & Negative & 98$\to$98 (0.0 pp) & 89$\to$94 (+5.0 pp) & 89$\to$92 (+2.5 pp) & 50$\to$61 (+10.8 pp) & 43$\to$35 (-8.3 pp) \\
Contribution & Positive & 98$\to$100 (+1.7 pp) & 89$\to$95 (+5.8 pp) & 89$\to$95 (+5.8 pp) & 50$\to$60 (+10.0 pp) & 43$\to$38 (-5.8 pp) \\
Lexical & Negative & 98$\to$98 (-0.8 pp) & 89$\to$86 (-3.3 pp) & 89$\to$92 (+3.3 pp) & 50$\to$59 (+9.2 pp) & 44$\to$38 (-5.9 pp) \\
Lexical & Positive & 98$\to$99 (+0.8 pp) & 89$\to$88 (-0.8 pp) & 89$\to$89 (0.0 pp) & 50$\to$50 (0.0 pp) & 44$\to$44 (0.0 pp) \\
\addlinespace
\rowcolor{black!6}\multicolumn{7}{l}{\textit{Strict protocol; Rewrite model: GPT-5.5}}\\
Novelty & Negative & 82$\to$80 (-2.5 pp) & 17$\to$10 (-6.7 pp) & 17$\to$10 (-6.7 pp) & 29$\to$30 (+0.8 pp) & 13$\to$3 (-9.2 pp) \\
Novelty & Positive & 82$\to$88 (+5.8 pp) & 17$\to$27 (+10.0 pp) & 17$\to$27 (+10.0 pp) & 29$\to$32 (+2.5 pp) & 12$\to$10 (-2.5 pp) \\
Evidence & Negative & 82$\to$77 (-5.8 pp) & 17$\to$12 (-4.2 pp) & 17$\to$8 (-8.3 pp) & 29$\to$27 (-2.5 pp) & 12$\to$6 (-6.7 pp) \\
Evidence & Positive & 82$\to$91 (+8.3 pp) & 17$\to$24 (+7.5 pp) & 17$\to$28 (+10.8 pp) & 29$\to$38 (+8.3 pp) & 13$\to$6 (-6.7 pp) \\
Scope & Negative & 82$\to$68 (-14.2 pp) & 17$\to$10 (-6.7 pp) & 17$\to$10 (-6.7 pp) & 29$\to$32 (+2.5 pp) & 12$\to$6 (-6.7 pp) \\
Scope & Positive & 82$\to$85 (+2.5 pp) & 17$\to$18 (+1.7 pp) & 17$\to$21 (+4.2 pp) & 29$\to$32 (+3.3 pp) & 12$\to$7 (-5.8 pp) \\
Technical & Negative & 82$\to$85 (+2.5 pp) & 17$\to$17 (0.0 pp) & 17$\to$20 (+3.3 pp) & 29$\to$32 (+2.5 pp) & 12$\to$8 (-5.0 pp) \\
Technical & Positive & 82$\to$87 (+4.2 pp) & 17$\to$17 (0.0 pp) & 17$\to$22 (+5.0 pp) & 29$\to$30 (+0.8 pp) & 12$\to$8 (-4.2 pp) \\
Contribution & Negative & 82$\to$91 (+8.3 pp) & 17$\to$17 (0.0 pp) & 17$\to$22 (+5.0 pp) & 29$\to$32 (+3.3 pp) & 12$\to$7 (-5.8 pp) \\
Contribution & Positive & 82$\to$88 (+5.0 pp) & 17$\to$18 (+0.8 pp) & 17$\to$23 (+6.7 pp) & 29$\to$38 (+8.3 pp) & 13$\to$8 (-4.2 pp) \\
Lexical & Negative & 82$\to$82 (-0.8 pp) & 17$\to$15 (-1.7 pp) & 17$\to$20 (+3.3 pp) & 29$\to$33 (+4.2 pp) & 12$\to$7 (-5.8 pp) \\
Lexical & Positive & 82$\to$85 (+2.5 pp) & 17$\to$18 (+0.8 pp) & 17$\to$18 (+0.8 pp) & 29$\to$29 (0.0 pp) & 13$\to$13 (0.0 pp) \\
\addlinespace
\rowcolor{black!6}\multicolumn{7}{l}{\textit{Strict protocol; Rewrite model: Opus 4.8}}\\
Novelty & Negative & 82$\to$69 (-13.3 pp) & 17$\to$10 (-6.7 pp) & 17$\to$5 (-11.7 pp) & 29$\to$22 (-6.7 pp) & 13$\to$4 (-8.4 pp) \\
Novelty & Positive & 82$\to$89 (+6.7 pp) & 17$\to$24 (+7.5 pp) & 17$\to$32 (+15.0 pp) & 29$\to$36 (+6.7 pp) & 13$\to$16 (+3.4 pp) \\
Evidence & Negative & 82$\to$69 (-13.3 pp) & 17$\to$11 (-5.8 pp) & 17$\to$8 (-8.3 pp) & 29$\to$22 (-7.5 pp) & 13$\to$5 (-7.6 pp) \\
Evidence & Positive & 82$\to$94 (+11.7 pp) & 17$\to$39 (+22.5 pp) & 17$\to$38 (+20.8 pp) & 29$\to$46 (+16.7 pp) & 13$\to$13 (0.0 pp) \\
Scope & Negative & 82$\to$64 (-18.3 pp) & 17$\to$12 (-4.2 pp) & 17$\to$10 (-6.7 pp) & 29$\to$28 (-0.8 pp) & 13$\to$6 (-6.7 pp) \\
Scope & Positive & 82$\to$87 (+4.2 pp) & 17$\to$25 (+8.3 pp) & 17$\to$22 (+5.8 pp) & 29$\to$32 (+3.3 pp) & 13$\to$17 (+4.2 pp) \\
Technical & Negative & 82$\to$87 (+4.2 pp) & 17$\to$18 (+1.7 pp) & 17$\to$22 (+5.8 pp) & 29$\to$33 (+4.2 pp) & 13$\to$10 (-2.5 pp) \\
Technical & Positive & 82$\to$91 (+8.3 pp) & 17$\to$29 (+12.5 pp) & 17$\to$23 (+6.7 pp) & 29$\to$35 (+5.8 pp) & 13$\to$13 (+0.8 pp) \\
Contribution & Negative & 82$\to$83 (+0.8 pp) & 17$\to$19 (+2.5 pp) & 17$\to$24 (+7.5 pp) & 29$\to$38 (+8.3 pp) & 12$\to$10 (-2.5 pp) \\
Contribution & Positive & 82$\to$94 (+11.7 pp) & 17$\to$23 (+6.7 pp) & 17$\to$25 (+8.3 pp) & 29$\to$39 (+10.0 pp) & 13$\to$11 (-1.7 pp) \\
Lexical & Negative & 82$\to$86 (+3.3 pp) & 17$\to$19 (+2.5 pp) & 17$\to$22 (+5.0 pp) & 29$\to$38 (+8.3 pp) & 12$\to$10 (-2.5 pp) \\
Lexical & Positive & 82$\to$83 (+0.8 pp) & 17$\to$17 (0.0 pp) & 17$\to$15 (-1.7 pp) & 29$\to$30 (+0.8 pp) & 13$\to$13 (0.0 pp) \\
\addlinespace
\end{longtable}
\endgroup

\begingroup
\scriptsize
\setlength{\tabcolsep}{1.8pt}
\begin{longtable}{llrrrlll}
\caption{Pooled transition counts underlying weak-accept probability across both rewrite models and all five reviewer models. Upward and Downward denote crossings of the weak-accept cutoff.}\phantomsection\label{tab:dimension_threshold_counts_combined}\\
\toprule
Dimension & Direction & Original $\geq6$ & Rewrite $\geq6$ & Change & Upward & Downward & Any flip \\
\midrule
\endfirsthead
\toprule
Dimension & Direction & Original $\geq6$ & Rewrite $\geq6$ & Change & Upward & Downward & Any flip \\
\midrule
\endhead
\midrule
\endfoot
\bottomrule
\endlastfoot
\rowcolor{black!6}\multicolumn{8}{l}{\textit{Standard evaluation protocol}}\\
Novelty & Negative & 887 (74.0\%) & 799 (66.7\%) & -7.3 pp & 46 (3.8\%) & 134 (11.2\%) & 180 (15.0\%) \\
Novelty & Positive & 887 (74.0\%) & 925 (77.2\%) & +3.2 pp & 86 (7.2\%) & 48 (4.0\%) & 134 (11.2\%) \\
Evidence & Negative & 888 (74.1\%) & 825 (68.9\%) & -5.3 pp & 46 (3.8\%) & 109 (9.1\%) & 155 (12.9\%) \\
Evidence & Positive & 888 (74.1\%) & 933 (77.9\%) & +3.8 pp & 94 (7.8\%) & 49 (4.1\%) & 143 (11.9\%) \\
Scope & Negative & 888 (74.2\%) & 828 (69.2\%) & -5.0 pp & 58 (4.8\%) & 118 (9.9\%) & 176 (14.7\%) \\
Scope & Positive & 888 (74.2\%) & 921 (76.9\%) & +2.8 pp & 78 (6.5\%) & 45 (3.8\%) & 123 (10.3\%) \\
Technical & Negative & 888 (74.1\%) & 905 (75.5\%) & +1.4 pp & 70 (5.8\%) & 53 (4.4\%) & 123 (10.3\%) \\
Technical & Positive & 887 (74.1\%) & 916 (76.5\%) & +2.4 pp & 80 (6.7\%) & 51 (4.3\%) & 131 (10.9\%) \\
Contribution & Negative & 886 (74.1\%) & 910 (76.2\%) & +2.0 pp & 84 (7.0\%) & 60 (5.0\%) & 144 (12.1\%) \\
Contribution & Positive & 887 (74.0\%) & 913 (76.2\%) & +2.2 pp & 83 (6.9\%) & 57 (4.8\%) & 140 (11.7\%) \\
Lexical & Negative & 888 (74.1\%) & 892 (74.5\%) & +0.3 pp & 63 (5.3\%) & 59 (4.9\%) & 122 (10.2\%) \\
Lexical & Positive & 888 (74.1\%) & 889 (74.2\%) & +0.1 pp & 62 (5.2\%) & 61 (5.1\%) & 123 (10.3\%) \\
\addlinespace
\rowcolor{black!6}\multicolumn{8}{l}{\textit{Strict evaluation protocol}}\\
Novelty & Negative & 376 (31.5\%) & 293 (24.5\%) & -6.9 pp & 49 (4.1\%) & 132 (11.0\%) & 181 (15.1\%) \\
Novelty & Positive & 378 (31.5\%) & 456 (38.0\%) & +6.5 pp & 137 (11.4\%) & 59 (4.9\%) & 196 (16.3\%) \\
Evidence & Negative & 378 (31.5\%) & 294 (24.5\%) & -7.0 pp & 53 (4.4\%) & 137 (11.4\%) & 190 (15.8\%) \\
Evidence & Positive & 378 (31.6\%) & 498 (41.6\%) & +10.0 pp & 172 (14.4\%) & 52 (4.3\%) & 224 (18.7\%) \\
Scope & Negative & 378 (31.6\%) & 296 (24.7\%) & -6.9 pp & 52 (4.3\%) & 134 (11.2\%) & 186 (15.6\%) \\
Scope & Positive & 378 (31.5\%) & 416 (34.7\%) & +3.2 pp & 97 (8.1\%) & 59 (4.9\%) & 156 (13.0\%) \\
Technical & Negative & 378 (31.6\%) & 398 (33.2\%) & +1.7 pp & 97 (8.1\%) & 77 (6.4\%) & 174 (14.5\%) \\
Technical & Positive & 378 (31.6\%) & 426 (35.6\%) & +4.0 pp & 118 (9.8\%) & 70 (5.8\%) & 188 (15.7\%) \\
Contribution & Negative & 377 (31.5\%) & 411 (34.3\%) & +2.8 pp & 103 (8.6\%) & 69 (5.8\%) & 172 (14.4\%) \\
Contribution & Positive & 376 (31.4\%) & 440 (36.8\%) & +5.4 pp & 126 (10.5\%) & 62 (5.2\%) & 188 (15.7\%) \\
Lexical & Negative & 378 (31.5\%) & 397 (33.1\%) & +1.6 pp & 95 (7.9\%) & 76 (6.3\%) & 171 (14.2\%) \\
Lexical & Positive & 378 (31.6\%) & 383 (32.0\%) & +0.4 pp & 83 (6.9\%) & 78 (6.5\%) & 161 (13.4\%) \\
\addlinespace
\end{longtable}
\endgroup

\subsection{Strict-Prompt Human-OA Range Analysis}
\label{app:strict_human_score_ranges}

Table~\ref{tab:human_score_directional_separation_strict} is the
strict-prompt counterpart to main-text
Table~\ref{tab:human_score_directional_separation}. It retains every fixed
row for each combination of rewriter and reviewer across the four human-OA ranges. As in the main-text
table, the only averages are the two rightmost within-row summaries across
the six rhetorical dimensions.

\begin{table}[!htbp]
  \centering
  \caption{\textbf{Strict-prompt overall-rating changes across human OA ranges.}
  Mean paired rating change relative to original by human OA range
  and fixed pairing of rewriter and reviewer. The two rightmost columns are the only displayed averages and summarize the six dimensions within each reviewer-level row.
  Pos. and Neg. denote rewrite directions. Green and yellow indicate increases
  and decreases, with color intensity proportional to the absolute change;
  zero changes remain white. \(\dagger\) marks [8,10], which contains only 3 papers.}
  \label{tab:human_score_directional_separation_strict}
  \scriptsize
  \setlength{\tabcolsep}{1.3pt}
  \renewcommand{\arraystretch}{0.78}
  \newcommand{\scoreup}[2]{\cellcolor{green!#1}#2}
  \newcommand{\scoredown}[2]{\cellcolor{yellow!#1}#2}
  \newcommand{\scorezero}[1]{#1}
  \begin{adjustbox}{max width=\textwidth}
  \begin{tabular}{@{}ll*{14}{r}@{}}
    \toprule
    \textbf{Rewriter} & \textbf{Reviewer}
      & \multicolumn{2}{c}{\textbf{Novelty}}
      & \multicolumn{2}{c}{\textbf{Evidence}}
      & \multicolumn{2}{c}{\textbf{Scope}}
      & \multicolumn{2}{c}{\textbf{Technical}}
      & \multicolumn{2}{c}{\textbf{Contribution}}
      & \multicolumn{2}{c}{\textbf{Lexical}}
      & \multicolumn{2}{c}{\shortstack{\textbf{Across}\\\textbf{dimensions}}} \\
    \cmidrule(lr){3-4}\cmidrule(lr){5-6}\cmidrule(lr){7-8}
    \cmidrule(lr){9-10}\cmidrule(lr){11-12}\cmidrule(lr){13-14}
    \cmidrule(l){15-16}
      & & \multicolumn{1}{c}{\textbf{Pos.}} & \multicolumn{1}{c}{\textbf{Neg.}}
      & \multicolumn{1}{c}{\textbf{Pos.}} & \multicolumn{1}{c}{\textbf{Neg.}}
      & \multicolumn{1}{c}{\textbf{Pos.}} & \multicolumn{1}{c}{\textbf{Neg.}}
      & \multicolumn{1}{c}{\textbf{Pos.}} & \multicolumn{1}{c}{\textbf{Neg.}}
      & \multicolumn{1}{c}{\textbf{Pos.}} & \multicolumn{1}{c}{\textbf{Neg.}}
      & \multicolumn{1}{c}{\textbf{Pos.}} & \multicolumn{1}{c}{\textbf{Neg.}}
      & \multicolumn{1}{c}{\textbf{Pos.}} & \multicolumn{1}{c}{\textbf{Neg.}} \\
    \midrule
    \multicolumn{16}{l}{\textit{human OA [1,3]}} \\
    \cmidrule(lr){1-16}
    GPT-5.5 & Gemini 3.5 FL & \scoreup{30}{+.450} & \scoredown{41}{-.475} & \scoreup{34}{+.575} & \scoredown{45}{-.575} & \scoreup{22}{+.250} & \scoredown{51}{-.725} & \scoreup{16}{+.125} & \scoredown{16}{-.075} & \scoreup{34}{+.575} & \scoreup{20}{+.200} & \scoreup{22}{+.250} & \scoredown{25}{-.175} & +.371 & -.304 \\
     & Qwen 3.5 F & \scoreup{24}{+.275} & \scoredown{25}{-.175} & \scoreup{24}{+.275} & \scoredown{9}{-.025} & \scoreup{17}{+.150} & \scoredown{41}{-.475} & \scoredown{16}{-.075} & \scoredown{27}{-.200} & \scoreup{16}{+.125} & \scoredown{21}{-.125} & \scoreup{17}{+.150} & \scoredown{27}{-.200} & +.150 & -.200 \\
     & GPT-5 mini & \scoreup{26}{+.325} & \scorezero{0.000} & \scoreup{24}{+.275} & \scoredown{16}{-.075} & \scoreup{28}{+.375} & \scorezero{0.000} & \scoreup{16}{+.125} & \scoreup{21}{+.225} & \scoreup{24}{+.275} & \scoreup{27}{+.350} & \scoredown{19}{-.100} & \scoreup{14}{+.100} & +.212 & +.100 \\
     & GPT-5.5 & \scoreup{21}{+.225} & \scoreup{12}{+.075} & \scoreup{26}{+.325} & \scoredown{16}{-.075} & \scoreup{20}{+.200} & \scoreup{16}{+.125} & \scoreup{21}{+.225} & \scoreup{20}{+.200} & \scoreup{25}{+.300} & \scoreup{19}{+.175} & \scoreup{16}{+.125} & \scoreup{20}{+.200} & +.233 & +.117 \\
     & Sonnet 5 & \scoreup{12}{+.075} & \scoredown{23}{-.150} & \scoredown{23}{-.150} & \scoredown{27}{-.200} & \scoreup{14}{+.100} & \scoredown{38}{-.400} & \scoredown{23}{-.150} & \scoredown{19}{-.100} & \scoredown{13}{-.050} & \scoredown{28}{-.211} & \scoredown{16}{-.075} & \scoredown{13}{-.050} & -.042 & -.185 \\
    \cmidrule(lr){1-16}
    Opus 4.8 & Gemini 3.5 FL & \scoreup{31}{+.475} & \scoredown{56}{-.875} & \scoreup{45}{+1.275} & \scoredown{49}{-.675} & \scoreup{24}{+.275} & \scoredown{55}{-.850} & \scoreup{36}{+.650} & \scoreup{12}{+.075} & \scoreup{38}{+.725} & \scoredown{33}{-.300} & \scoreup{14}{+.100} & \scoreup{19}{+.175} & +.583 & -.408 \\
     & Qwen 3.5 F & \scoreup{17}{+.150} & \scoredown{34}{-.325} & \scoreup{27}{+.350} & \scoredown{39}{-.425} & \scoreup{26}{+.325} & \scoredown{27}{-.200} & \scoreup{19}{+.175} & \scoredown{21}{-.125} & \scoreup{28}{+.375} & \scoredown{21}{-.125} & \scorezero{0.000} & \scorezero{0.000} & +.229 & -.200 \\
     & GPT-5 mini & \scoreup{33}{+.550} & \scoredown{43}{-.525} & \scoreup{40}{+.775} & \scoredown{23}{-.150} & \scoreup{26}{+.325} & \scoredown{13}{-.050} & \scoreup{25}{+.300} & \scoreup{24}{+.275} & \scoreup{28}{+.375} & \scoreup{20}{+.200} & \scoreup{10}{+.050} & \scoreup{28}{+.375} & +.396 & +.021 \\
     & GPT-5.5 & \scoreup{16}{+.125} & \scoreup{10}{+.050} & \scoreup{31}{+.475} & \scoredown{9}{-.025} & \scoreup{22}{+.250} & \scoreup{19}{+.175} & \scoreup{22}{+.250} & \scoreup{17}{+.150} & \scoreup{22}{+.250} & \scoreup{16}{+.125} & \scoredown{23}{-.150} & \scoreup{20}{+.200} & +.200 & +.113 \\
     & Sonnet 5 & \scorezero{0.000} & \scoredown{38}{-.400} & \scoreup{20}{+.200} & \scoredown{39}{-.425} & \scoredown{13}{-.050} & \scoredown{34}{-.316} & \scoredown{16}{-.075} & \scoredown{19}{-.103} & \scoredown{13}{-.050} & \scoredown{16}{-.075} & \scoreup{7}{+.025} & \scoredown{9}{-.025} & +.008 & -.224 \\
    \midrule
    \multicolumn{16}{l}{\textit{human OA [4,5]}} \\
    \cmidrule(lr){1-16}
    GPT-5.5 & Gemini 3.5 FL & \scoreup{32}{+.500} & \scoredown{34}{-.325} & \scoreup{41}{+.825} & \scoredown{41}{-.475} & \scoreup{27}{+.350} & \scoredown{34}{-.325} & \scoreup{22}{+.250} & \scoreup{22}{+.250} & \scoreup{35}{+.600} & \scoreup{28}{+.375} & \scoreup{19}{+.175} & \scoreup{14}{+.100} & +.450 & -.067 \\
     & Qwen 3.5 F & \scoreup{29}{+.425} & \scoredown{31}{-.275} & \scoreup{22}{+.250} & \scoredown{41}{-.475} & \scoreup{25}{+.300} & \scoredown{13}{-.050} & \scoredown{19}{-.100} & \scoredown{16}{-.075} & \scoreup{19}{+.175} & \scoredown{13}{-.050} & \scoreup{7}{+.025} & \scoreup{14}{+.100} & +.179 & -.138 \\
     & GPT-5 mini & \scoreup{33}{+.525} & \scoredown{38}{-.400} & \scoreup{35}{+.600} & \scoredown{37}{-.375} & \scoreup{28}{+.375} & \scoredown{19}{-.100} & \scoreup{25}{+.300} & \scoreup{21}{+.225} & \scoreup{31}{+.475} & \scoreup{22}{+.250} & \scorezero{0.000} & \scoredown{9}{-.025} & +.379 & -.071 \\
     & GPT-5.5 & \scoreup{12}{+.075} & \scoreup{17}{+.150} & \scoreup{27}{+.350} & \scorezero{0.000} & \scoreup{24}{+.275} & \scoreup{16}{+.125} & \scoreup{21}{+.225} & \scoredown{19}{-.100} & \scoreup{30}{+.450} & \scoreup{28}{+.375} & \scoreup{21}{+.225} & \scoreup{20}{+.200} & +.267 & +.125 \\
     & Sonnet 5 & \scoredown{19}{-.100} & \scoredown{48}{-.632} & \scoredown{41}{-.462} & \scoredown{44}{-.550} & \scoredown{35}{-.350} & \scoredown{50}{-.700} & \scoredown{30}{-.256} & \scoredown{35}{-.350} & \scoredown{38}{-.410} & \scoredown{43}{-.525} & \scoredown{27}{-.205} & \scoredown{34}{-.325} & -.297 & -.514 \\
    \cmidrule(lr){1-16}
    Opus 4.8 & Gemini 3.5 FL & \scoreup{39}{+.750} & \scoredown{46}{-.600} & \scoreup{45}{+1.200} & \scoredown{40}{-.450} & \scoreup{28}{+.400} & \scoredown{48}{-.650} & \scoreup{33}{+.525} & \scoreup{27}{+.350} & \scoreup{41}{+.825} & \scoreup{22}{+.250} & \scorezero{0.000} & \scoreup{14}{+.100} & +.617 & -.167 \\
     & Qwen 3.5 F & \scoreup{27}{+.350} & \scoredown{28}{-.225} & \scoreup{36}{+.650} & \scoreup{10}{+.050} & \scoreup{10}{+.050} & \scoredown{34}{-.325} & \scoreup{30}{+.450} & \scoredown{19}{-.100} & \scoreup{16}{+.125} & \scoreup{26}{+.325} & \scoreup{16}{+.125} & \scoredown{13}{-.050} & +.292 & -.054 \\
     & GPT-5 mini & \scoreup{29}{+.425} & \scoredown{50}{-.700} & \scoreup{41}{+.850} & \scoredown{45}{-.575} & \scoreup{7}{+.025} & \scoredown{37}{-.375} & \scoreup{30}{+.450} & \scoreup{19}{+.175} & \scoreup{26}{+.325} & \scoreup{28}{+.375} & \scoredown{25}{-.175} & \scorezero{0.000} & +.317 & -.183 \\
     & GPT-5.5 & \scoreup{28}{+.375} & \scoredown{16}{-.075} & \scoreup{36}{+.625} & \scoredown{9}{-.025} & \scoreup{21}{+.225} & \scoredown{21}{-.125} & \scoreup{24}{+.275} & \scoreup{26}{+.325} & \scoreup{29}{+.425} & \scoreup{29}{+.425} & \scoreup{19}{+.175} & \scoreup{27}{+.350} & +.350 & +.146 \\
     & Sonnet 5 & \scoreup{14}{+.103} & \scoredown{48}{-.641} & \scoredown{14}{-.051} & \scoredown{43}{-.513} & \scoredown{17}{-.077} & \scoredown{38}{-.395} & \scoreup{7}{+.026} & \scoredown{34}{-.316} & \scoredown{21}{-.128} & \scoredown{31}{-.275} & \scoredown{32}{-.282} & \scoredown{23}{-.150} & -.068 & -.382 \\
    \midrule
    \multicolumn{16}{l}{\textit{human OA [6,7]}} \\
    \cmidrule(lr){1-16}
    GPT-5.5 & Gemini 3.5 FL & \scoreup{13}{+.081} & \scoredown{33}{-.297} & \scoreup{31}{+.486} & \scoredown{41}{-.459} & \scoredown{33}{-.297} & \scoredown{51}{-.730} & \scoreup{23}{+.270} & \scoredown{14}{-.054} & \scoreup{25}{+.297} & \scoredown{22}{-.135} & \scorezero{0.000} & \scoredown{24}{-.162} & +.140 & -.306 \\
     & Qwen 3.5 F & \scoredown{10}{-.027} & \scoredown{43}{-.514} & \scoredown{43}{-.514} & \scoredown{39}{-.432} & \scoredown{31}{-.270} & \scoredown{42}{-.486} & \scoredown{24}{-.162} & \scoredown{20}{-.108} & \scoredown{36}{-.351} & \scoredown{22}{-.135} & \scoredown{24}{-.162} & \scoredown{31}{-.270} & -.248 & -.324 \\
     & GPT-5 mini & \scoreup{31}{+.486} & \scoredown{28}{-.216} & \scoreup{25}{+.297} & \scoredown{36}{-.351} & \scoreup{20}{+.189} & \scoredown{26}{-.189} & \scoreup{20}{+.189} & \scoreup{25}{+.297} & \scoreup{26}{+.324} & \scoreup{26}{+.324} & \scoreup{17}{+.135} & \scoreup{17}{+.135} & +.270 & \scorezero{0.000} \\
     & GPT-5.5 & \scoreup{7}{+.027} & \scoredown{31}{-.270} & \scoredown{10}{-.027} & \scoredown{31}{-.270} & \scoreup{7}{+.027} & \scorezero{0.000} & \scoredown{17}{-.081} & \scoreup{13}{+.081} & \scoredown{20}{-.108} & \scoredown{28}{-.216} & \scorezero{0.000} & \scoreup{13}{+.081} & -.027 & -.099 \\
     & Sonnet 5 & \scoreup{7}{+.027} & \scoredown{42}{-.500} & \scoredown{30}{-.243} & \scoredown{43}{-.514} & \scoredown{33}{-.297} & \scoredown{45}{-.568} & \scorezero{0.000} & \scoredown{33}{-.297} & \scoredown{25}{-.171} & \scoredown{37}{-.389} & \scoredown{20}{-.108} & \scoredown{34}{-.324} & -.132 & -.432 \\
    \cmidrule(lr){1-16}
    Opus 4.8 & Gemini 3.5 FL & \scoreup{32}{+.514} & \scoredown{53}{-.784} & \scoreup{27}{+.351} & \scoredown{48}{-.649} & \scoreup{18}{+.162} & \scoredown{46}{-.595} & \scoreup{20}{+.189} & \scoredown{14}{-.054} & \scoreup{28}{+.378} & \scoreup{13}{+.081} & \scoredown{22}{-.135} & \scoreup{7}{+.027} & +.243 & -.329 \\
     & Qwen 3.5 F & \scoredown{22}{-.135} & \scoredown{41}{-.459} & \scoreup{34}{+.568} & \scoredown{44}{-.541} & \scoreup{10}{+.054} & \scoredown{31}{-.270} & \scorezero{0.000} & \scoredown{14}{-.054} & \scoreup{20}{+.189} & \scoredown{20}{-.108} & \scoredown{37}{-.378} & \scoredown{38}{-.405} & +.050 & -.306 \\
     & GPT-5 mini & \scoreup{29}{+.405} & \scoredown{49}{-.676} & \scoreup{38}{+.730} & \scoredown{30}{-.243} & \scoreup{28}{+.378} & \scoredown{24}{-.162} & \scoreup{25}{+.297} & \scoreup{35}{+.622} & \scoreup{23}{+.270} & \scoreup{31}{+.459} & \scoreup{13}{+.081} & \scoreup{27}{+.351} & +.360 & +.059 \\
     & GPT-5.5 & \scoredown{24}{-.162} & \scoredown{36}{-.351} & \scoreup{28}{+.378} & \scoredown{26}{-.189} & \scoredown{26}{-.189} & \scoredown{17}{-.081} & \scoreup{18}{+.162} & \scoreup{15}{+.108} & \scoreup{17}{+.135} & \scorezero{0.000} & \scoredown{24}{-.162} & \scoreup{10}{+.054} & +.027 & -.077 \\
     & Sonnet 5 & \scoredown{10}{-.027} & \scoredown{45}{-.556} & \scoreup{13}{+.081} & \scoredown{47}{-.622} & \scorezero{0.000} & \scoredown{39}{-.432} & \scoredown{20}{-.108} & \scoredown{24}{-.162} & \scoredown{22}{-.135} & \scoredown{26}{-.189} & \scoredown{20}{-.108} & \scoredown{22}{-.135} & -.050 & -.349 \\
    \midrule
    \multicolumn{16}{l}{\textit{human OA [8,10]\textsuperscript{\(\dagger\)}}} \\
    \cmidrule(lr){1-16}
    GPT-5.5 & Gemini 3.5 FL & \scorezero{0.000} & \scoredown{49}{-.667} & \scorezero{0.000} & \scorezero{0.000} & \scorezero{0.000} & \scoredown{49}{-.667} & \scoredown{49}{-.667} & \scorezero{0.000} & \scorezero{0.000} & \scorezero{0.000} & \scorezero{0.000} & \scorezero{0.000} & -.111 & -.222 \\
     & Qwen 3.5 F & \scoredown{35}{-.333} & \scoredown{60}{-1.000} & \scoreup{26}{+.333} & \scoredown{60}{-1.000} & \scorezero{0.000} & \scoredown{60}{-1.000} & \scoredown{35}{-.333} & \scoredown{35}{-.333} & \scoredown{60}{-1.333} & \scoredown{49}{-.667} & \scoreup{37}{+.667} & \scoredown{60}{-1.333} & -.167 & -.889 \\
     & GPT-5 mini & \scoreup{45}{+1.333} & \scoreup{37}{+.667} & \scoreup{26}{+.333} & \scoredown{35}{-.333} & \scoreup{37}{+.667} & \scoredown{35}{-.333} & \scoreup{37}{+.667} & \scoreup{45}{+1.000} & \scoreup{37}{+.667} & \scoreup{37}{+.667} & \scoredown{49}{-.667} & \scorezero{0.000} & +.500 & +.278 \\
     & GPT-5.5 & \scoreup{37}{+.667} & \scorezero{0.000} & \scorezero{0.000} & \scoredown{49}{-.667} & \scorezero{0.000} & \scoredown{35}{-.333} & \scoreup{37}{+.667} & \scoreup{37}{+.667} & \scorezero{0.000} & \scorezero{0.000} & \scoreup{37}{+.667} & \scoreup{37}{+.667} & +.333 & +.056 \\
     & Sonnet 5 & \scoredown{35}{-.333} & \scoredown{49}{-.667} & \scoredown{35}{-.333} & \scoredown{35}{-.333} & \scoreup{37}{+.667} & \scoredown{35}{-.333} & \scoreup{26}{+.333} & \scoreup{26}{+.333} & \scoreup{26}{+.333} & \scoredown{35}{-.333} & \scoredown{35}{-.333} & \scorezero{0.000} & +.056 & -.222 \\
    \cmidrule(lr){1-16}
    Opus 4.8 & Gemini 3.5 FL & \scorezero{0.000} & \scorezero{0.000} & \scorezero{0.000} & \scoredown{60}{-1.667} & \scorezero{0.000} & \scoredown{49}{-.667} & \scorezero{0.000} & \scorezero{0.000} & \scorezero{0.000} & \scorezero{0.000} & \scorezero{0.000} & \scorezero{0.000} & \scorezero{0.000} & -.389 \\
     & Qwen 3.5 F & \scoredown{60}{-1.333} & \scoredown{60}{-1.000} & \scoreup{37}{+.667} & \scorezero{0.000} & \scoredown{60}{-1.000} & \scorezero{0.000} & \scorezero{0.000} & \scoredown{35}{-.333} & \scoreup{26}{+.333} & \scoredown{60}{-1.667} & \scoredown{35}{-.333} & \scorezero{0.000} & -.278 & -.500 \\
     & GPT-5 mini & \scoreup{37}{+.667} & \scorezero{0.000} & \scoreup{26}{+.333} & \scorezero{0.000} & \scoreup{26}{+.333} & \scoredown{60}{-1.000} & \scoreup{45}{+1.000} & \scoreup{37}{+.667} & \scoreup{45}{+1.333} & \scoreup{45}{+1.000} & \scoreup{37}{+.667} & \scoreup{37}{+.667} & +.722 & +.222 \\
     & GPT-5.5 & \scoreup{37}{+.667} & \scoreup{37}{+.667} & \scorezero{0.000} & \scorezero{0.000} & \scoreup{37}{+.667} & \scoredown{49}{-.667} & \scoreup{37}{+.667} & \scorezero{0.000} & \scoreup{37}{+.667} & \scoreup{37}{+.667} & \scoreup{37}{+.667} & \scorezero{0.000} & +.556 & +.111 \\
     & Sonnet 5 & \scoredown{35}{-.333} & \scoredown{49}{-.667} & \scoreup{26}{+.333} & \scoreup{26}{+.333} & \scorezero{0.000} & \scoredown{35}{-.333} & \scoredown{35}{-.333} & \scoredown{35}{-.333} & \scorezero{0.000} & \scorezero{0.000} & \scorezero{0.000} & \scoredown{35}{-.333} & -.056 & -.222 \\
    \bottomrule
  \end{tabular}
  \end{adjustbox}
\end{table}

\subsection{AI Original-Score Range Analysis}
\label{app:original_score_range_analysis}
\label{app:strict_ai_score_ranges}

Table~\ref{tab:score_boundary_directional_separation_strict} is the
strict-prompt counterpart to main-text
Table~\ref{tab:score_boundary_directional_separation}. It retains every fixed
row for each combination of rewriter and reviewer across the four AI original-score ranges. The only
averages are the two rightmost within-row summaries across the six rhetorical
dimensions.

\begin{table}[!htbp]
  \centering
  \caption{\textbf{Strict-prompt overall-rating changes across AI original-score ranges.}
  Mean paired rating change relative to original by AI original-score range
  and fixed pairing of rewriter and reviewer. The two rightmost columns are the only displayed averages and summarize the six dimensions within each reviewer-level row.
  Pos. and Neg. denote rewrite directions. Green and yellow indicate increases
  and decreases, with color intensity proportional to the absolute change;
  zero changes remain white. \(\dagger\) marks sparse [8,10] reviewer-specific cells. A dash indicates that no papers are available for the corresponding cell.}
  \label{tab:score_boundary_directional_separation_strict}
  \scriptsize
  \setlength{\tabcolsep}{1.3pt}
  \renewcommand{\arraystretch}{0.78}
  \newcommand{\scoreup}[2]{\cellcolor{green!#1}#2}
  \newcommand{\scoredown}[2]{\cellcolor{yellow!#1}#2}
  \newcommand{\scorezero}[1]{#1}
  \begin{adjustbox}{max width=\textwidth}
  \begin{tabular}{@{}ll*{14}{r}@{}}
    \toprule
    \textbf{Rewriter} & \textbf{Reviewer}
      & \multicolumn{2}{c}{\textbf{Novelty}}
      & \multicolumn{2}{c}{\textbf{Evidence}}
      & \multicolumn{2}{c}{\textbf{Scope}}
      & \multicolumn{2}{c}{\textbf{Technical}}
      & \multicolumn{2}{c}{\textbf{Contribution}}
      & \multicolumn{2}{c}{\textbf{Lexical}}
      & \multicolumn{2}{c}{\shortstack{\textbf{Across}\\\textbf{dimensions}}} \\
    \cmidrule(lr){3-4}\cmidrule(lr){5-6}\cmidrule(lr){7-8}
    \cmidrule(lr){9-10}\cmidrule(lr){11-12}\cmidrule(lr){13-14}
    \cmidrule(l){15-16}
      & & \multicolumn{1}{c}{\textbf{Pos.}} & \multicolumn{1}{c}{\textbf{Neg.}}
      & \multicolumn{1}{c}{\textbf{Pos.}} & \multicolumn{1}{c}{\textbf{Neg.}}
      & \multicolumn{1}{c}{\textbf{Pos.}} & \multicolumn{1}{c}{\textbf{Neg.}}
      & \multicolumn{1}{c}{\textbf{Pos.}} & \multicolumn{1}{c}{\textbf{Neg.}}
      & \multicolumn{1}{c}{\textbf{Pos.}} & \multicolumn{1}{c}{\textbf{Neg.}}
      & \multicolumn{1}{c}{\textbf{Pos.}} & \multicolumn{1}{c}{\textbf{Neg.}}
      & \multicolumn{1}{c}{\textbf{Pos.}} & \multicolumn{1}{c}{\textbf{Neg.}} \\
    \midrule
    \multicolumn{16}{l}{\textit{AI Original-score [1,3]}} \\
    \cmidrule(lr){1-16}
    GPT-5.5 & Gemini 3.5 FL & \scoreup{45}{+1.545} & \scoreup{43}{+.909} & \scoreup{45}{+2.091} & \scoreup{45}{+1.000} & \scoreup{45}{+1.364} & \scoreup{45}{+1.182} & \scoreup{45}{+1.455} & \scoreup{45}{+1.455} & \scoreup{45}{+1.909} & \scoreup{45}{+1.727} & \scoreup{45}{+1.636} & \scoreup{38}{+.727} & +1.667 & +1.167 \\
     & Qwen 3.5 F & \scoreup{45}{+1.750} & \scoreup{40}{+.786} & \scoreup{45}{+1.214} & \scoreup{42}{+.857} & \scoreup{45}{+1.393} & \scoreup{40}{+.786} & \scoreup{45}{+1.071} & \scoreup{45}{+1.179} & \scoreup{45}{+1.393} & \scoreup{44}{+.964} & \scoreup{45}{+1.071} & \scoreup{45}{+1.429} & +1.315 & +1.000 \\
     & GPT-5 mini & \scoreup{45}{+1.167} & \scoreup{37}{+.685} & \scoreup{45}{+1.111} & \scoreup{27}{+.370} & \scoreup{45}{+.981} & \scoreup{34}{+.556} & \scoreup{36}{+.648} & \scoreup{42}{+.889} & \scoreup{45}{+1.111} & \scoreup{43}{+.926} & \scoreup{32}{+.500} & \scoreup{36}{+.648} & +.920 & +.679 \\
     & GPT-5.5 & \scoreup{31}{+.463} & \scoreup{31}{+.463} & \scoreup{40}{+.805} & \scoreup{27}{+.366} & \scoreup{34}{+.585} & \scoreup{33}{+.537} & \scoreup{34}{+.561} & \scoreup{28}{+.390} & \scoreup{38}{+.707} & \scoreup{31}{+.488} & \scoreup{31}{+.463} & \scoreup{34}{+.561} & +.598 & +.467 \\
     & Sonnet 5 & \scoreup{29}{+.429} & \scoreup{17}{+.143} & \scoreup{17}{+.143} & \scoreup{17}{+.143} & \scoreup{26}{+.333} & \scoreup{10}{+.048} & \scoreup{26}{+.333} & \scoreup{17}{+.143} & \scoreup{20}{+.190} & \scoreup{14}{+.098} & \scoreup{17}{+.143} & \scoreup{20}{+.190} & +.262 & +.127 \\
    \cmidrule(lr){1-16}
    Opus 4.8 & Gemini 3.5 FL & \scoreup{45}{+1.727} & \scoreup{38}{+.727} & \scoreup{45}{+2.364} & \scoreup{30}{+.455} & \scoreup{45}{+1.455} & \scoreup{19}{+.182} & \scoreup{45}{+2.000} & \scoreup{45}{+1.182} & \scoreup{45}{+2.273} & \scoreup{38}{+.727} & \scoreup{45}{+1.000} & \scoreup{45}{+1.909} & +1.803 & +.864 \\
     & Qwen 3.5 F & \scoreup{45}{+1.036} & \scoreup{43}{+.893} & \scoreup{45}{+1.643} & \scoreup{44}{+.964} & \scoreup{45}{+1.179} & \scoreup{42}{+.857} & \scoreup{45}{+1.500} & \scoreup{45}{+1.214} & \scoreup{45}{+1.714} & \scoreup{45}{+1.107} & \scoreup{45}{+1.250} & \scoreup{43}{+.893} & +1.387 & +.988 \\
     & GPT-5 mini & \scoreup{45}{+1.000} & \scoreup{23}{+.259} & \scoreup{45}{+1.481} & \scoreup{29}{+.426} & \scoreup{40}{+.778} & \scoreup{32}{+.500} & \scoreup{45}{+1.037} & \scoreup{45}{+.981} & \scoreup{42}{+.889} & \scoreup{42}{+.889} & \scoreup{33}{+.537} & \scoreup{40}{+.796} & +.954 & +.642 \\
     & GPT-5.5 & \scoreup{28}{+.390} & \scoreup{33}{+.537} & \scoreup{44}{+.951} & \scoreup{24}{+.293} & \scoreup{32}{+.512} & \scoreup{31}{+.488} & \scoreup{41}{+.829} & \scoreup{34}{+.585} & \scoreup{38}{+.732} & \scoreup{31}{+.488} & \scoreup{25}{+.317} & \scoreup{35}{+.610} & +.622 & +.500 \\
     & Sonnet 5 & \scoreup{28}{+.381} & \scoreup{10}{+.048} & \scoreup{31}{+.476} & \scorezero{0.000} & \scoreup{29}{+.429} & \scoreup{14}{+.098} & \scoreup{24}{+.286} & \scoreup{14}{+.095} & \scoreup{26}{+.333} & \scoreup{20}{+.190} & \scoreup{22}{+.238} & \scoreup{26}{+.333} & +.357 & +.127 \\
    \midrule
    \multicolumn{16}{l}{\textit{AI Original-score [4,5]}} \\
    \cmidrule(lr){1-16}
    GPT-5.5 & Gemini 3.5 FL & \scoreup{35}{+.600} & \scoreup{20}{+.200} & \scoreup{45}{+1.400} & \scoredown{27}{-.200} & \scoreup{14}{+.100} & \scoredown{60}{-1.000} & \scoreup{28}{+.400} & \scoreup{32}{+.500} & \scoreup{45}{+1.000} & \scoreup{38}{+.700} & \scoreup{35}{+.600} & \scoreup{25}{+.300} & +.683 & +.083 \\
     & Qwen 3.5 F & \scoredown{10}{-.028} & \scoredown{36}{-.361} & \scoreup{8}{+.028} & \scoredown{41}{-.472} & \scoredown{23}{-.153} & \scoredown{42}{-.486} & \scoredown{29}{-.236} & \scoredown{35}{-.347} & \scoredown{26}{-.194} & \scoredown{30}{-.250} & \scoredown{17}{-.083} & \scoredown{39}{-.417} & -.111 & -.389 \\
     & GPT-5 mini & \scoredown{25}{-.174} & \scoredown{56}{-.870} & \scoredown{20}{-.109} & \scoredown{50}{-.696} & \scoredown{22}{-.130} & \scoredown{47}{-.609} & \scoredown{18}{-.087} & \scoredown{25}{-.174} & \scoredown{28}{-.217} & \scoredown{29}{-.239} & \scoredown{43}{-.522} & \scoredown{40}{-.435} & -.207 & -.504 \\
     & GPT-5.5 & \scoredown{20}{-.114} & \scoredown{26}{-.182} & \scoredown{18}{-.091} & \scoredown{31}{-.273} & \scorezero{0.000} & \scoredown{9}{-.023} & \scoredown{9}{-.023} & \scoredown{9}{-.023} & \scoreup{14}{+.091} & \scoreup{10}{+.045} & \scoredown{9}{-.023} & \scoredown{13}{-.045} & -.027 & -.083 \\
     & Sonnet 5 & \scoredown{24}{-.159} & \scoredown{51}{-.721} & \scoredown{41}{-.468} & \scoredown{51}{-.714} & \scoredown{38}{-.397} & \scoredown{57}{-.905} & \scoredown{36}{-.355} & \scoredown{39}{-.413} & \scoredown{39}{-.419} & \scoredown{46}{-.597} & \scoredown{29}{-.226} & \scoredown{38}{-.397} & -.337 & -.624 \\
    \cmidrule(lr){1-16}
    Opus 4.8 & Gemini 3.5 FL & \scoreup{45}{+1.000} & \scoredown{60}{-1.000} & \scoreup{45}{+1.700} & \scoreup{20}{+.200} & \scoreup{32}{+.500} & \scoredown{57}{-.900} & \scoreup{45}{+1.100} & \scoreup{20}{+.200} & \scoreup{45}{+1.100} & \scoreup{40}{+.800} & \scoreup{14}{+.100} & \scoreup{20}{+.200} & +.917 & -.083 \\
     & Qwen 3.5 F & \scoreup{5}{+.014} & \scoredown{45}{-.569} & \scoreup{30}{+.431} & \scoredown{39}{-.417} & \scoredown{7}{-.014} & \scoredown{38}{-.403} & \scoreup{14}{+.097} & \scoredown{30}{-.250} & \scoredown{10}{-.028} & \scoredown{26}{-.194} & \scoredown{34}{-.319} & \scoredown{28}{-.222} & +.030 & -.343 \\
     & GPT-5 mini & \scoreup{11}{+.065} & \scoredown{60}{-1.326} & \scoreup{23}{+.261} & \scoredown{57}{-.913} & \scoredown{23}{-.152} & \scoredown{55}{-.826} & \scoredown{27}{-.196} & \scoredown{23}{-.152} & \scorezero{0.000} & \scoredown{18}{-.087} & \scoredown{41}{-.457} & \scoredown{29}{-.239} & -.080 & -.591 \\
     & GPT-5.5 & \scoreup{10}{+.045} & \scoredown{41}{-.477} & \scoreup{27}{+.364} & \scoredown{16}{-.068} & \scoredown{18}{-.091} & \scoredown{24}{-.159} & \scoredown{22}{-.136} & \scoreup{10}{+.045} & \scoreup{10}{+.045} & \scoreup{10}{+.045} & \scoredown{22}{-.136} & \scoreup{17}{+.136} & +.015 & -.080 \\
     & Sonnet 5 & \scoredown{22}{-.129} & \scoredown{55}{-.855} & \scoredown{17}{-.081} & \scoredown{52}{-.742} & \scoredown{34}{-.323} & \scoredown{46}{-.600} & \scoredown{23}{-.145} & \scoredown{31}{-.267} & \scoredown{33}{-.306} & \scoredown{35}{-.333} & \scoredown{31}{-.274} & \scoredown{32}{-.286} & -.210 & -.514 \\
    \midrule
    \multicolumn{16}{l}{\textit{AI Original-score [6,7]}} \\
    \cmidrule(lr){1-16}
    GPT-5.5 & Gemini 3.5 FL & \scoreup{38}{+.700} & \scoredown{27}{-.200} & \scoreup{45}{+.980} & \scoredown{32}{-.280} & \scoreup{32}{+.500} & \scoredown{31}{-.260} & \scoreup{31}{+.460} & \scoreup{24}{+.280} & \scoreup{40}{+.800} & \scoreup{25}{+.320} & \scoreup{23}{+.260} & \scoreup{16}{+.120} & +.617 & -.003 \\
     & Qwen 3.5 F & \scoredown{36}{-.357} & \scoredown{60}{-1.214} & \scoredown{60}{-1.071} & \scoredown{56}{-.857} & \scoredown{45}{-.571} & \scoredown{60}{-1.000} & \scoredown{56}{-.857} & \scoredown{48}{-.643} & \scoredown{60}{-1.071} & \scoredown{58}{-.929} & \scoredown{53}{-.786} & \scoredown{60}{-1.000} & -.786 & -.940 \\
     & GPT-5 mini & \scoreup{10}{+.050} & \scoredown{58}{-.950} & \scoredown{38}{-.400} & \scoredown{60}{-1.000} & \scoredown{38}{-.400} & \scoredown{50}{-.700} & \scoredown{30}{-.250} & \scoredown{38}{-.400} & \scoredown{33}{-.300} & \scoredown{13}{-.050} & \scoredown{27}{-.200} & \scoredown{35}{-.350} & -.250 & -.575 \\
     & GPT-5.5 & \scoreup{8}{+.030} & \scoredown{33}{-.303} & \scoredown{10}{-.030} & \scoredown{39}{-.424} & \scoredown{21}{-.121} & \scoredown{30}{-.242} & \scoredown{18}{-.091} & \scoredown{26}{-.182} & \scoredown{18}{-.091} & \scoredown{26}{-.182} & \scoredown{15}{-.061} & \scorezero{0.000} & -.061 & -.222 \\
     & Sonnet 5 & \scoredown{42}{-.500} & \scoredown{53}{-.769} & \scoredown{48}{-.643} & \scoredown{48}{-.643} & \scoredown{39}{-.429} & \scoredown{48}{-.643} & \scoredown{42}{-.500} & \scoredown{39}{-.429} & \scoredown{37}{-.385} & \scoredown{50}{-.692} & \scoredown{39}{-.429} & \scoredown{42}{-.500} & -.481 & -.613 \\
    \cmidrule(lr){1-16}
    Opus 4.8 & Gemini 3.5 FL & \scoreup{45}{+1.020} & \scoredown{44}{-.540} & \scoreup{45}{+1.500} & \scoredown{27}{-.200} & \scoreup{32}{+.520} & \scoredown{37}{-.380} & \scoreup{41}{+.820} & \scoreup{31}{+.480} & \scoreup{45}{+1.000} & \scoreup{17}{+.140} & \scoreup{24}{+.280} & \scoreup{25}{+.300} & +.857 & -.033 \\
     & Qwen 3.5 F & \scoredown{51}{-.714} & \scoredown{51}{-.714} & \scoredown{45}{-.571} & \scoredown{60}{-1.071} & \scoredown{42}{-.500} & \scoredown{60}{-1.000} & \scoredown{53}{-.786} & \scoredown{60}{-1.071} & \scoredown{28}{-.214} & \scoredown{48}{-.643} & \scoredown{42}{-.500} & \scoredown{53}{-.786} & -.548 & -.881 \\
     & GPT-5 mini & \scoredown{13}{-.050} & \scoredown{60}{-1.350} & \scoreup{10}{+.050} & \scoredown{58}{-.950} & \scoredown{33}{-.300} & \scoredown{52}{-.750} & \scoredown{23}{-.150} & \scoredown{23}{-.150} & \scoredown{33}{-.300} & \scoredown{13}{-.050} & \scoredown{38}{-.400} & \scoredown{19}{-.100} & -.192 & -.558 \\
     & GPT-5.5 & \scorezero{0.000} & \scoredown{35}{-.333} & \scoreup{14}{+.091} & \scoredown{39}{-.424} & \scoredown{18}{-.091} & \scoredown{36}{-.364} & \scoreup{14}{+.091} & \scoredown{18}{-.091} & \scoreup{11}{+.061} & \scoreup{11}{+.061} & \scoredown{30}{-.242} & \scoredown{28}{-.212} & -.015 & -.227 \\
     & Sonnet 5 & \scoredown{39}{-.429} & \scoredown{53}{-.769} & \scoredown{28}{-.214} & \scoredown{53}{-.786} & \scoredown{16}{-.071} & \scoredown{51}{-.714} & \scoredown{45}{-.571} & \scoredown{48}{-.643} & \scoredown{36}{-.357} & \scoredown{39}{-.429} & \scoredown{36}{-.357} & \scoredown{42}{-.500} & -.333 & -.640 \\
    \midrule
    \multicolumn{16}{l}{\textit{AI Original-score [8,10]\textsuperscript{\(\dagger\)}}} \\
    \cmidrule(lr){1-16}
    GPT-5.5 & Gemini 3.5 FL & \scoredown{35}{-.347} & \scoredown{59}{-.959} & \scoredown{30}{-.245} & \scoredown{60}{-1.102} & \scoredown{45}{-.571} & \scoredown{60}{-1.245} & \scoredown{38}{-.408} & \scoredown{47}{-.612} & \scoredown{31}{-.265} & \scoredown{42}{-.490} & \scoredown{38}{-.408} & \scoredown{44}{-.531} & -.374 & -.823 \\
     & Qwen 3.5 F & \scoredown{60}{-2.667} & \scoredown{60}{-3.167} & \scoredown{60}{-3.000} & \scoredown{60}{-2.833} & \scoredown{60}{-2.000} & \scoredown{60}{-2.500} & \scoredown{60}{-2.500} & \scoredown{60}{-2.500} & \scoredown{60}{-2.500} & \scoredown{60}{-1.667} & \scoredown{60}{-1.667} & \scoredown{60}{-2.333} & -2.389 & -2.500 \\
     & GPT-5 mini & \multicolumn{1}{c}{--} & \multicolumn{1}{c}{--} & \multicolumn{1}{c}{--} & \multicolumn{1}{c}{--} & \multicolumn{1}{c}{--} & \multicolumn{1}{c}{--} & \multicolumn{1}{c}{--} & \multicolumn{1}{c}{--} & \multicolumn{1}{c}{--} & \multicolumn{1}{c}{--} & \multicolumn{1}{c}{--} & \multicolumn{1}{c}{--} & \multicolumn{1}{c}{--} & \multicolumn{1}{c}{--} \\
     & GPT-5.5 & \scorezero{0.000} & \scoredown{60}{-1.000} & \scoredown{60}{-1.000} & \scoredown{60}{-2.000} & \scorezero{0.000} & \scoredown{60}{-2.000} & \scoredown{60}{-1.000} & \scorezero{0.000} & \scoredown{60}{-2.000} & \scoredown{60}{-1.000} & \scorezero{0.000} & \scorezero{0.000} & -.667 & -1.000 \\
     & Sonnet 5 & \scoredown{60}{-2.000} & \scoredown{60}{-2.000} & \scoredown{60}{-2.000} & \scoredown{60}{-2.000} & \scoredown{60}{-2.000} & \scoredown{60}{-2.000} & \scorezero{0.000} & \scoredown{60}{-2.000} & \multicolumn{1}{c}{--} & \scoredown{60}{-2.000} & \scoredown{60}{-2.000} & \scoredown{60}{-3.000} & -1.600 & -2.167 \\
    \cmidrule(lr){1-16}
    Opus 4.8 & Gemini 3.5 FL & \scoredown{30}{-.245} & \scoredown{60}{-1.204} & \scoredown{21}{-.122} & \scoredown{60}{-1.449} & \scoredown{32}{-.286} & \scoredown{60}{-1.184} & \scoredown{38}{-.408} & \scoredown{42}{-.490} & \scoredown{27}{-.204} & \scoredown{40}{-.449} & \scoredown{45}{-.551} & \scoredown{44}{-.531} & -.303 & -.884 \\
     & Qwen 3.5 F & \scoredown{60}{-1.500} & \scoredown{60}{-2.667} & \scoredown{60}{-1.000} & \scoredown{60}{-2.833} & \scoredown{60}{-1.833} & \scoredown{60}{-2.000} & \scoredown{60}{-2.167} & \scoredown{60}{-2.167} & \scoredown{60}{-2.500} & \scoredown{60}{-1.500} & \scoredown{60}{-2.500} & \scoredown{60}{-2.500} & -1.917 & -2.278 \\
     & GPT-5 mini & \multicolumn{1}{c}{--} & \multicolumn{1}{c}{--} & \multicolumn{1}{c}{--} & \multicolumn{1}{c}{--} & \multicolumn{1}{c}{--} & \multicolumn{1}{c}{--} & \multicolumn{1}{c}{--} & \multicolumn{1}{c}{--} & \multicolumn{1}{c}{--} & \multicolumn{1}{c}{--} & \multicolumn{1}{c}{--} & \multicolumn{1}{c}{--} & \multicolumn{1}{c}{--} & \multicolumn{1}{c}{--} \\
     & GPT-5.5 & \scoredown{60}{-1.000} & \scoredown{60}{-1.000} & \scorezero{0.000} & \scoredown{60}{-2.000} & \scorezero{0.000} & \scoredown{60}{-2.000} & \scoredown{60}{-1.000} & \scorezero{0.000} & \scorezero{0.000} & \scorezero{0.000} & \scoredown{60}{-1.000} & \scorezero{0.000} & -.500 & -.833 \\
     & Sonnet 5 & \scorezero{0.000} & \scoredown{60}{-2.000} & \scoredown{60}{-2.000} & \scoredown{60}{-2.000} & \scoredown{60}{-2.000} & \scoredown{60}{-2.000} & \scoredown{60}{-2.000} & \scoredown{60}{-2.000} & \scoredown{60}{-2.000} & \scoredown{60}{-2.000} & \scoredown{60}{-2.000} & \scoredown{60}{-2.000} & -1.667 & -2.000 \\
    \bottomrule
  \end{tabular}
  \end{adjustbox}
\end{table}

Each evaluation cell is assigned to $[1,3]$, $[4,5]$, $[6,7]$, or $[8,10]$ using the
prompt- and reviewer-matched AI original overall rating. The same paper can
occupy different ranges for different reviewer models, so the unique-paper
counts in Table~\ref{tab:score_band_counts} are not mutually exclusive across
columns.

\begin{table}[!htbp]
  \centering
  \caption{Unique papers contributing to each AI original-score range in the
  direction-specific response profiles.}
  \label{tab:score_band_counts}
  \small
  \begin{tabular}{lrrrr}
    \toprule
    Evaluation protocol & [1,3] & [4,5] & [6,7] & [8,10] \\
    \midrule
    Standard & 20 & 78 & 114 & 104 \\
    Strict & 73 & 104 & 86 & 49 \\
    \bottomrule
  \end{tabular}
\end{table}

Scores near the ends of a bounded scale have more room to move toward the
center than farther outward. Extreme-range movement may therefore partly
reflect scale bounds and regression to the mean rather than rhetorical
steering. These profiles are descriptive and should not be interpreted as
unbiased estimates of how baseline paper quality moderates rewriting effects.
Some cells for fixed reviewer and prompt combinations remain sparse at the extremes, so these
ranges are retained as descriptive diagnostics rather than independent
subgroup claims.

\subsubsection{Panel-Aligned Values for the Main-Text Displays}
\label{app:score_band_panel_values}

Tables~\ref{tab:score_band_panel_rating_values} and
\ref{tab:score_band_panel_threshold_values} report the direction-specific
rating and threshold values complementing
Table~\ref{tab:score_boundary_directional_separation} and the boundary
analysis in Section~\ref{subsec:reviewer_score_conditioning}. These tables
preserve the two rewrite models as separate panels. For each paper,
reviewer-model changes are first averaged within the relevant evaluation
protocol, rhetorical condition, and original-score range; the reported cells
then average those paper-level values.

\begingroup
\scriptsize
\setlength{\tabcolsep}{4.5pt}
\begin{longtable}{llrrrr}
\caption{Panel-aligned mean paper-level rating changes across original-score ranges. All five reviewers are averaged within paper.}\phantomsection\label{tab:score_band_panel_rating_values}\\
\toprule
Dimension & Direction & [1,3] & [4,5] & [6,7] & [8,10] \\
\midrule
\endfirsthead
\toprule
Dimension & Direction & [1,3] & [4,5] & [6,7] & [8,10] \\
\midrule
\endhead
\midrule
\endfoot
\bottomrule
\endlastfoot
\rowcolor{black!6}\multicolumn{6}{l}{\textit{Standard protocol; Rewrite model: GPT-5.5}}\\
Novelty & Negative & +0.633 & -0.122 & -0.111 & -0.674 \\
Novelty & Positive & +0.575 & +0.155 & +0.306 & -0.224 \\
Evidence & Negative & +0.667 & -0.134 & -0.074 & -0.605 \\
Evidence & Positive & +0.933 & +0.129 & +0.237 & -0.274 \\
Scope & Negative & +0.775 & -0.087 & -0.087 & -0.746 \\
Scope & Positive & +0.825 & +0.092 & +0.305 & -0.292 \\
Technical & Negative & +0.717 & +0.052 & +0.207 & -0.322 \\
Technical & Positive & +0.867 & +0.119 & +0.219 & -0.344 \\
Contribution & Negative & +0.758 & +0.144 & +0.234 & -0.300 \\
Contribution & Positive & +0.917 & +0.141 & +0.215 & -0.191 \\
Lexical & Negative & +0.733 & -0.003 & +0.168 & -0.399 \\
Lexical & Positive & +0.875 & +0.138 & +0.176 & -0.250 \\
\addlinespace
\rowcolor{black!6}\multicolumn{6}{l}{\textit{Standard protocol; Rewrite model: Opus 4.8}}\\
Novelty & Negative & +0.742 & -0.366 & -0.292 & -0.840 \\
Novelty & Positive & +0.733 & +0.174 & +0.369 & -0.227 \\
Evidence & Negative & +0.883 & -0.207 & -0.149 & -0.737 \\
Evidence & Positive & +1.167 & +0.281 & +0.591 & -0.101 \\
Scope & Negative & +1.083 & -0.016 & -0.211 & -0.726 \\
Scope & Positive & +1.175 & +0.238 & +0.388 & -0.172 \\
Technical & Negative & +0.767 & +0.138 & +0.131 & -0.281 \\
Technical & Positive & +1.092 & +0.303 & +0.408 & -0.209 \\
Contribution & Negative & +0.875 & +0.149 & +0.115 & -0.279 \\
Contribution & Positive & +0.933 & +0.182 & +0.492 & -0.119 \\
Lexical & Negative & +0.783 & +0.027 & +0.250 & -0.260 \\
Lexical & Positive & +0.683 & +0.082 & +0.124 & -0.275 \\
\addlinespace
\rowcolor{black!6}\multicolumn{6}{l}{\textit{Strict protocol; Rewrite model: GPT-5.5}}\\
Novelty & Negative & +0.821 & -0.499 & -0.465 & -1.128 \\
Novelty & Positive & +1.249 & -0.033 & +0.229 & -0.510 \\
Evidence & Negative & +0.634 & -0.543 & -0.534 & -1.269 \\
Evidence & Positive & +1.151 & -0.075 & +0.234 & -0.429 \\
Scope & Negative & +0.737 & -0.560 & -0.417 & -1.384 \\
Scope & Positive & +1.043 & -0.190 & +0.089 & -0.673 \\
Technical & Negative & +0.950 & -0.223 & -0.159 & -0.759 \\
Technical & Positive & +0.877 & -0.204 & -0.010 & -0.534 \\
Contribution & Negative & +0.882 & -0.250 & -0.049 & -0.616 \\
Contribution & Positive & +1.090 & -0.162 & +0.180 & -0.439 \\
Lexical & Negative & +0.896 & -0.292 & -0.195 & -0.651 \\
Lexical & Positive & +0.799 & -0.104 & -0.137 & -0.520 \\
\addlinespace
\rowcolor{black!6}\multicolumn{6}{l}{\textit{Strict protocol; Rewrite model: Opus 4.8}}\\
Novelty & Negative & +0.550 & -0.760 & -0.721 & -1.306 \\
Novelty & Positive & +1.012 & +0.040 & +0.278 & -0.340 \\
Evidence & Negative & +0.543 & -0.524 & -0.577 & -1.621 \\
Evidence & Positive & +1.524 & +0.293 & +0.687 & -0.194 \\
Scope & Negative & +0.596 & -0.503 & -0.591 & -1.310 \\
Scope & Positive & +0.878 & -0.058 & +0.078 & -0.391 \\
Technical & Negative & +0.912 & -0.184 & +0.053 & -0.626 \\
Technical & Positive & +1.191 & -0.026 & +0.209 & -0.561 \\
Contribution & Negative & +0.915 & -0.138 & -0.054 & -0.531 \\
Contribution & Positive & +1.161 & +0.034 & +0.380 & -0.345 \\
Lexical & Negative & +0.985 & -0.178 & -0.106 & -0.677 \\
Lexical & Positive & +0.757 & -0.287 & -0.096 & -0.677 \\
\addlinespace
\end{longtable}
\endgroup

\begingroup
\scriptsize
\setlength{\tabcolsep}{4.5pt}
\begin{longtable}{llrrrr}
\caption{Panel-aligned mean paper-level changes in weak-accept probability across original-score ranges, in percentage points. All five reviewers are averaged within paper.}\phantomsection\label{tab:score_band_panel_threshold_values}\\
\toprule
Dimension & Direction & [1,3] & [4,5] & [6,7] & [8,10] \\
\midrule
\endfirsthead
\toprule
Dimension & Direction & [1,3] & [4,5] & [6,7] & [8,10] \\
\midrule
\endhead
\midrule
\endfoot
\bottomrule
\endlastfoot
\rowcolor{black!6}\multicolumn{6}{l}{\textit{Standard protocol; Rewrite model: GPT-5.5}}\\
Novelty & Negative & +6.67 & +17.09 & -18.64 & -3.04 \\
Novelty & Positive & +2.50 & +34.29 & -9.50 & 0.00 \\
Evidence & Negative & +8.33 & +19.34 & -17.91 & 0.00 \\
Evidence & Positive & +8.33 & +29.59 & -14.33 & 0.00 \\
Scope & Negative & +9.17 & +20.19 & -19.81 & -1.92 \\
Scope & Positive & +7.50 & +26.92 & -8.11 & -0.48 \\
Technical & Negative & +5.00 & +23.82 & -8.55 & -0.32 \\
Technical & Positive & +1.67 & +23.82 & -13.08 & -0.96 \\
Contribution & Negative & +9.17 & +30.66 & -11.99 & -0.48 \\
Contribution & Positive & +8.33 & +28.85 & -13.30 & -0.32 \\
Lexical & Negative & 0.00 & +20.83 & -11.11 & 0.00 \\
Lexical & Positive & +9.17 & +22.12 & -12.35 & -0.48 \\
\addlinespace
\rowcolor{black!6}\multicolumn{6}{l}{\textit{Standard protocol; Rewrite model: Opus 4.8}}\\
Novelty & Negative & +9.17 & +11.86 & -34.14 & -4.01 \\
Novelty & Positive & +6.67 & +29.59 & -8.55 & -0.32 \\
Evidence & Negative & +5.00 & +11.65 & -24.85 & -1.60 \\
Evidence & Positive & +8.33 & +40.28 & -6.51 & 0.00 \\
Scope & Negative & +6.67 & +19.55 & -26.10 & -1.28 \\
Scope & Positive & +8.33 & +31.41 & -8.85 & -0.48 \\
Technical & Negative & +6.67 & +28.10 & -11.77 & -0.32 \\
Technical & Positive & +10.83 & +33.12 & -5.99 & 0.00 \\
Contribution & Negative & +14.17 & +28.95 & -11.77 & 0.00 \\
Contribution & Positive & +10.00 & +29.27 & -8.33 & 0.00 \\
Lexical & Negative & +8.33 & +25.75 & -12.43 & 0.00 \\
Lexical & Positive & +1.67 & +22.54 & -10.16 & -0.96 \\
\addlinespace
\rowcolor{black!6}\multicolumn{6}{l}{\textit{Strict protocol; Rewrite model: GPT-5.5}}\\
Novelty & Negative & +4.38 & +10.98 & -38.76 & -6.29 \\
Novelty & Positive & +16.53 & +21.11 & -15.89 & -5.78 \\
Evidence & Negative & +3.36 & +10.77 & -42.93 & -6.80 \\
Evidence & Positive & +12.01 & +25.24 & -20.35 & -3.74 \\
Scope & Negative & +4.57 & +9.01 & -38.18 & -13.27 \\
Scope & Positive & +6.89 & +13.73 & -21.32 & -2.04 \\
Technical & Negative & +8.42 & +15.34 & -25.97 & -3.06 \\
Technical & Positive & +6.37 & +17.37 & -27.91 & -3.06 \\
Contribution & Negative & +8.01 & +18.01 & -25.39 & -2.04 \\
Contribution & Positive & +10.59 & +18.27 & -22.87 & -5.61 \\
Lexical & Negative & +5.41 & +13.49 & -28.20 & -2.55 \\
Lexical & Positive & +4.38 & +19.39 & -26.84 & -2.04 \\
\addlinespace
\rowcolor{black!6}\multicolumn{6}{l}{\textit{Strict protocol; Rewrite model: Opus 4.8}}\\
Novelty & Negative & +3.01 & +6.73 & -51.16 & -5.78 \\
Novelty & Positive & +10.37 & +27.88 & -21.41 & -1.02 \\
Evidence & Negative & +4.11 & +7.85 & -46.41 & -18.88 \\
Evidence & Positive & +21.99 & +31.89 & -10.56 & 0.00 \\
Scope & Negative & +2.28 & +9.78 & -49.22 & -14.29 \\
Scope & Positive & +2.67 & +22.12 & -17.93 & -1.02 \\
Technical & Negative & +7.81 & +20.27 & -20.64 & -7.14 \\
Technical & Positive & +14.57 & +22.52 & -17.25 & -3.06 \\
Contribution & Negative & +4.84 & +18.89 & -21.32 & -0.68 \\
Contribution & Positive & +10.25 & +24.76 & -15.50 & -2.55 \\
Lexical & Negative & +10.89 & +20.11 & -21.80 & -3.06 \\
Lexical & Positive & +6.10 & +14.26 & -24.32 & -4.08 \\
\addlinespace
\end{longtable}
\endgroup

\subsubsection{Pooled Overall-Rating Inference}
\label{app:score_band_rating_details}

Table~\ref{tab:score_band_rating_inference_combined} gives the pooled original
and rewrite means, signed change, absolute movement, and paper-bootstrap
intervals for all four ranges, six rhetorical dimensions, both directions,
and both evaluation protocols. The two rewrite models and five reviewer
models are averaged within paper.

\begingroup
\scriptsize
\setlength{\tabcolsep}{2.0pt}
\begin{table*}[p]
\centering
\begin{adjustbox}{max width=\textwidth,center}
\begin{tabular}{lllrrrrl}
\toprule
Original range & Dimension & Direction & Original & Rewrite & Mean $\Delta$ & Mean $|\Delta|$ & 95\% CI \\
\midrule
\rowcolor{black!6}\multicolumn{8}{l}{\textit{Standard evaluation protocol}}\\
{[1,3]} & Novelty & Negative & 2.967 & 3.654 & +0.688 & 0.688 & [+0.267, +1.271] \\
{[1,3]} & Novelty & Positive & 2.967 & 3.621 & +0.654 & 0.654 & [+0.342, +0.979] \\
{[1,3]} & Evidence & Negative & 2.967 & 3.742 & +0.775 & 0.775 & [+0.383, +1.275] \\
{[1,3]} & Evidence & Positive & 2.967 & 4.017 & +1.050 & 1.100 & [+0.558, +1.617] \\
{[1,3]} & Scope & Negative & 2.967 & 3.896 & +0.929 & 0.929 & [+0.542, +1.337] \\
{[1,3]} & Scope & Positive & 2.967 & 3.962 & +0.996 & 0.996 & [+0.588, +1.525] \\
{[1,3]} & Technical & Negative & 2.967 & 3.708 & +0.742 & 0.742 & [+0.383, +1.133] \\
{[1,3]} & Technical & Positive & 2.967 & 3.946 & +0.979 & 0.979 & [+0.575, +1.417] \\
{[1,3]} & Contribution & Negative & 2.967 & 3.783 & +0.817 & 0.817 & [+0.392, +1.333] \\
{[1,3]} & Contribution & Positive & 2.967 & 3.892 & +0.925 & 0.975 & [+0.425, +1.500] \\
{[1,3]} & Lexical & Negative & 2.967 & 3.725 & +0.758 & 0.758 & [+0.408, +1.125] \\
{[1,3]} & Lexical & Positive & 2.967 & 3.746 & +0.779 & 0.779 & [+0.467, +1.125] \\
{[4,5]} & Novelty & Negative & 5.000 & 4.757 & -0.243 & 0.450 & [-0.379, -0.110] \\
{[4,5]} & Novelty & Positive & 5.000 & 5.162 & +0.162 & 0.481 & [+0.019, +0.300] \\
{[4,5]} & Evidence & Negative & 5.000 & 4.830 & -0.170 & 0.443 & [-0.315, -0.031] \\
{[4,5]} & Evidence & Positive & 5.000 & 5.205 & +0.205 & 0.489 & [+0.047, +0.343] \\
{[4,5]} & Scope & Negative & 5.000 & 4.949 & -0.051 & 0.422 & [-0.196, +0.085] \\
{[4,5]} & Scope & Positive & 5.000 & 5.163 & +0.163 & 0.433 & [+0.015, +0.301] \\
{[4,5]} & Technical & Negative & 5.000 & 5.094 & +0.094 & 0.373 & [-0.033, +0.221] \\
{[4,5]} & Technical & Positive & 5.000 & 5.211 & +0.211 & 0.399 & [+0.082, +0.349] \\
{[4,5]} & Contribution & Negative & 5.000 & 5.147 & +0.147 & 0.433 & [+0.009, +0.278] \\
{[4,5]} & Contribution & Positive & 5.000 & 5.160 & +0.160 & 0.468 & [+0.007, +0.310] \\
{[4,5]} & Lexical & Negative & 5.000 & 5.009 & +0.009 & 0.419 & [-0.127, +0.139] \\
{[4,5]} & Lexical & Positive & 5.000 & 5.110 & +0.110 & 0.356 & [-0.007, +0.225] \\
\bottomrule
\end{tabular}
\end{adjustbox}
\caption{Pooled overall-rating results across original-score ranges. Both rewrite models and all five reviewer models are averaged within paper; intervals bootstrap papers.}\phantomsection\label{tab:score_band_rating_inference_combined}
\end{table*}

\begin{table*}[p]
\ContinuedFloat
\centering
\begin{adjustbox}{max width=\textwidth,center}
\begin{tabular}{lllrrrrl}
\toprule
Original range & Dimension & Direction & Original & Rewrite & Mean $\Delta$ & Mean $|\Delta|$ & 95\% CI \\
\midrule
\rowcolor{black!6}\multicolumn{8}{l}{\textit{Standard evaluation protocol}}\\
{[6,7]} & Novelty & Negative & 6.000 & 5.795 & -0.205 & 0.338 & [-0.288, -0.126] \\
{[6,7]} & Novelty & Positive & 6.000 & 6.337 & +0.337 & 0.441 & [+0.246, +0.430] \\
{[6,7]} & Evidence & Negative & 6.000 & 5.889 & -0.111 & 0.315 & [-0.192, -0.034] \\
{[6,7]} & Evidence & Positive & 6.000 & 6.414 & +0.414 & 0.528 & [+0.313, +0.514] \\
{[6,7]} & Scope & Negative & 6.000 & 5.851 & -0.149 & 0.307 & [-0.238, -0.069] \\
{[6,7]} & Scope & Positive & 6.000 & 6.346 & +0.346 & 0.450 & [+0.252, +0.443] \\
{[6,7]} & Technical & Negative & 6.000 & 6.169 & +0.169 & 0.350 & [+0.080, +0.258] \\
{[6,7]} & Technical & Positive & 6.000 & 6.314 & +0.314 & 0.425 & [+0.222, +0.405] \\
{[6,7]} & Contribution & Negative & 6.000 & 6.174 & +0.174 & 0.332 & [+0.090, +0.260] \\
{[6,7]} & Contribution & Positive & 6.000 & 6.353 & +0.353 & 0.464 & [+0.254, +0.459] \\
{[6,7]} & Lexical & Negative & 6.000 & 6.209 & +0.209 & 0.379 & [+0.113, +0.307] \\
{[6,7]} & Lexical & Positive & 6.000 & 6.150 & +0.150 & 0.317 & [+0.072, +0.232] \\
{[8,10]} & Novelty & Negative & 8.000 & 7.243 & -0.757 & 0.757 & [-0.889, -0.632] \\
{[8,10]} & Novelty & Positive & 8.000 & 7.774 & -0.226 & 0.226 & [-0.301, -0.156] \\
{[8,10]} & Evidence & Negative & 8.000 & 7.329 & -0.671 & 0.671 & [-0.801, -0.544] \\
{[8,10]} & Evidence & Positive & 8.000 & 7.813 & -0.187 & 0.187 & [-0.248, -0.130] \\
{[8,10]} & Scope & Negative & 8.000 & 7.264 & -0.736 & 0.736 & [-0.858, -0.615] \\
{[8,10]} & Scope & Positive & 8.000 & 7.768 & -0.232 & 0.232 & [-0.309, -0.162] \\
{[8,10]} & Technical & Negative & 8.000 & 7.699 & -0.301 & 0.301 & [-0.385, -0.225] \\
{[8,10]} & Technical & Positive & 8.000 & 7.724 & -0.276 & 0.276 & [-0.351, -0.206] \\
{[8,10]} & Contribution & Negative & 8.000 & 7.710 & -0.290 & 0.290 & [-0.368, -0.217] \\
{[8,10]} & Contribution & Positive & 8.000 & 7.845 & -0.155 & 0.155 & [-0.210, -0.103] \\
{[8,10]} & Lexical & Negative & 8.000 & 7.671 & -0.329 & 0.329 & [-0.421, -0.242] \\
{[8,10]} & Lexical & Positive & 8.000 & 7.738 & -0.262 & 0.262 & [-0.339, -0.192] \\
\bottomrule
\end{tabular}
\end{adjustbox}
\caption[]{Pooled overall-rating results across original-score ranges. Both rewrite models and all five reviewer models are averaged within paper; intervals bootstrap papers.}
\end{table*}

\begin{table*}[p]
\ContinuedFloat
\centering
\begin{adjustbox}{max width=\textwidth,center}
\begin{tabular}{lllrrrrl}
\toprule
Original range & Dimension & Direction & Original & Rewrite & Mean $\Delta$ & Mean $|\Delta|$ & 95\% CI \\
\midrule
\rowcolor{black!6}\multicolumn{8}{l}{\textit{Strict evaluation protocol}}\\
{[1,3]} & Novelty & Negative & 3.000 & 3.685 & +0.685 & 0.685 & [+0.509, +0.875] \\
{[1,3]} & Novelty & Positive & 3.000 & 4.131 & +1.131 & 1.131 & [+0.899, +1.370] \\
{[1,3]} & Evidence & Negative & 3.000 & 3.589 & +0.589 & 0.589 & [+0.424, +0.769] \\
{[1,3]} & Evidence & Positive & 3.000 & 4.338 & +1.338 & 1.338 & [+1.121, +1.566] \\
{[1,3]} & Scope & Negative & 3.000 & 3.666 & +0.666 & 0.666 & [+0.508, +0.824] \\
{[1,3]} & Scope & Positive & 3.000 & 3.961 & +0.961 & 0.961 & [+0.783, +1.148] \\
{[1,3]} & Technical & Negative & 3.000 & 3.931 & +0.931 & 0.931 & [+0.729, +1.135] \\
{[1,3]} & Technical & Positive & 3.000 & 4.034 & +1.034 & 1.034 & [+0.839, +1.241] \\
{[1,3]} & Contribution & Negative & 3.000 & 3.899 & +0.899 & 0.899 & [+0.700, +1.104] \\
{[1,3]} & Contribution & Positive & 3.000 & 4.125 & +1.125 & 1.125 & [+0.927, +1.343] \\
{[1,3]} & Lexical & Negative & 3.000 & 3.940 & +0.940 & 0.940 & [+0.741, +1.155] \\
{[1,3]} & Lexical & Positive & 3.000 & 3.778 & +0.778 & 0.778 & [+0.586, +0.975] \\
{[4,5]} & Novelty & Negative & 5.000 & 4.378 & -0.622 & 0.715 & [-0.750, -0.496] \\
{[4,5]} & Novelty & Positive & 5.000 & 5.003 & +0.003 & 0.485 & [-0.134, +0.139] \\
{[4,5]} & Evidence & Negative & 5.000 & 4.466 & -0.534 & 0.619 & [-0.657, -0.414] \\
{[4,5]} & Evidence & Positive & 5.000 & 5.109 & +0.109 & 0.480 & [-0.015, +0.233] \\
{[4,5]} & Scope & Negative & 5.000 & 4.466 & -0.534 & 0.653 & [-0.659, -0.415] \\
{[4,5]} & Scope & Positive & 5.000 & 4.876 & -0.124 & 0.438 & [-0.244, -0.002] \\
{[4,5]} & Technical & Negative & 5.000 & 4.796 & -0.204 & 0.470 & [-0.333, -0.082] \\
{[4,5]} & Technical & Positive & 5.000 & 4.885 & -0.115 & 0.482 & [-0.242, +0.008] \\
{[4,5]} & Contribution & Negative & 5.000 & 4.807 & -0.193 & 0.473 & [-0.315, -0.072] \\
{[4,5]} & Contribution & Positive & 5.000 & 4.936 & -0.064 & 0.466 & [-0.196, +0.071] \\
{[4,5]} & Lexical & Negative & 5.000 & 4.765 & -0.235 & 0.542 & [-0.363, -0.111] \\
{[4,5]} & Lexical & Positive & 5.000 & 4.804 & -0.196 & 0.414 & [-0.300, -0.089] \\
\bottomrule
\end{tabular}
\end{adjustbox}
\caption[]{Pooled overall-rating results across original-score ranges. Both rewrite models and all five reviewer models are averaged within paper; intervals bootstrap papers.}
\end{table*}

\begin{table*}[p]
\ContinuedFloat
\centering
\begin{adjustbox}{max width=\textwidth,center}
\begin{tabular}{lllrrrrl}
\toprule
Original range & Dimension & Direction & Original & Rewrite & Mean $\Delta$ & Mean $|\Delta|$ & 95\% CI \\
\midrule
\rowcolor{black!6}\multicolumn{8}{l}{\textit{Strict evaluation protocol}}\\
{[6,7]} & Novelty & Negative & 6.000 & 5.410 & -0.590 & 0.712 & [-0.771, -0.419] \\
{[6,7]} & Novelty & Positive & 6.000 & 6.253 & +0.253 & 0.698 & [+0.054, +0.449] \\
{[6,7]} & Evidence & Negative & 6.000 & 5.445 & -0.555 & 0.622 & [-0.709, -0.406] \\
{[6,7]} & Evidence & Positive & 6.000 & 6.461 & +0.461 & 0.803 & [+0.259, +0.662] \\
{[6,7]} & Scope & Negative & 6.000 & 5.496 & -0.504 & 0.649 & [-0.663, -0.349] \\
{[6,7]} & Scope & Positive & 6.000 & 6.084 & +0.084 & 0.530 & [-0.079, +0.262] \\
{[6,7]} & Technical & Negative & 6.000 & 5.947 & -0.053 & 0.559 & [-0.219, +0.114] \\
{[6,7]} & Technical & Positive & 6.000 & 6.100 & +0.100 & 0.566 & [-0.061, +0.262] \\
{[6,7]} & Contribution & Negative & 6.000 & 5.949 & -0.051 & 0.423 & [-0.180, +0.078] \\
{[6,7]} & Contribution & Positive & 6.000 & 6.280 & +0.280 & 0.663 & [+0.102, +0.461] \\
{[6,7]} & Lexical & Negative & 6.000 & 5.850 & -0.150 & 0.428 & [-0.303, -0.003] \\
{[6,7]} & Lexical & Positive & 6.000 & 5.884 & -0.116 & 0.420 & [-0.253, +0.016] \\
{[8,10]} & Novelty & Negative & 8.000 & 6.783 & -1.217 & 1.217 & [-1.444, -0.980] \\
{[8,10]} & Novelty & Positive & 8.000 & 7.575 & -0.425 & 0.425 & [-0.617, -0.240] \\
{[8,10]} & Evidence & Negative & 8.000 & 6.555 & -1.445 & 1.445 & [-1.686, -1.202] \\
{[8,10]} & Evidence & Positive & 8.000 & 7.689 & -0.311 & 0.311 & [-0.454, -0.179] \\
{[8,10]} & Scope & Negative & 8.000 & 6.653 & -1.347 & 1.347 & [-1.607, -1.076] \\
{[8,10]} & Scope & Positive & 8.000 & 7.468 & -0.532 & 0.532 & [-0.724, -0.352] \\
{[8,10]} & Technical & Negative & 8.000 & 7.308 & -0.692 & 0.692 & [-0.937, -0.459] \\
{[8,10]} & Technical & Positive & 8.000 & 7.452 & -0.548 & 0.548 & [-0.730, -0.374] \\
{[8,10]} & Contribution & Negative & 8.000 & 7.427 & -0.573 & 0.573 & [-0.767, -0.383] \\
{[8,10]} & Contribution & Positive & 8.000 & 7.607 & -0.393 & 0.393 & [-0.582, -0.219] \\
{[8,10]} & Lexical & Negative & 8.000 & 7.336 & -0.664 & 0.664 & [-0.878, -0.451] \\
{[8,10]} & Lexical & Positive & 8.000 & 7.401 & -0.599 & 0.599 & [-0.796, -0.408] \\
\bottomrule
\end{tabular}
\end{adjustbox}
\caption[]{Pooled overall-rating results across original-score ranges. Both rewrite models and all five reviewer models are averaged within paper; intervals bootstrap papers.}
\end{table*}

\endgroup

\subsubsection{Pooled Rating-6 Transitions Near the Decision Boundary}
\label{app:score_band_threshold_details}

Table~\ref{tab:score_band_boundary_counts_combined} reports pooled transition
counts and rates for [4,5] and [6,7], the two ranges adjacent to the rating-6
boundary. Cells beginning in [4,5] can cross only upward, while cells
beginning in [6,7] can cross only downward. Extreme-range panel values remain
available in Table~\ref{tab:score_band_panel_threshold_values}, but their full
transition-count decomposition is omitted because those ranges are far from
the boundary and some fixed cells are sparse.

\begingroup
\tiny
\setlength{\tabcolsep}{1.2pt}
\begin{table*}[p]
\centering
\begin{adjustbox}{max width=\textwidth,center}
\begin{tabular}{lllrrrrrr}
\toprule
Range & Dimension & Direction & $n$ & Original $\geq6$ & Rewrite $\geq6$ & Change & Upward & Downward \\
\midrule
\rowcolor{black!6}\multicolumn{9}{l}{\textit{Standard evaluation protocol}}\\
{[1,3]} & Novelty & Negative & 64 & 0 (0.0\%) & 5 (7.8\%) & +7.81 pp & 5 (7.8\%) & 0 (0.0\%) \\
{[1,3]} & Novelty & Positive & 64 & 0 (0.0\%) & 3 (4.7\%) & +4.69 pp & 3 (4.7\%) & 0 (0.0\%) \\
{[1,3]} & Evidence & Negative & 64 & 0 (0.0\%) & 4 (6.2\%) & +6.25 pp & 4 (6.2\%) & 0 (0.0\%) \\
{[1,3]} & Evidence & Positive & 64 & 0 (0.0\%) & 6 (9.4\%) & +9.38 pp & 6 (9.4\%) & 0 (0.0\%) \\
{[1,3]} & Scope & Negative & 64 & 0 (0.0\%) & 5 (7.8\%) & +7.81 pp & 5 (7.8\%) & 0 (0.0\%) \\
{[1,3]} & Scope & Positive & 63 & 0 (0.0\%) & 5 (7.9\%) & +7.94 pp & 5 (7.9\%) & 0 (0.0\%) \\
{[1,3]} & Technical & Negative & 64 & 0 (0.0\%) & 3 (4.7\%) & +4.69 pp & 3 (4.7\%) & 0 (0.0\%) \\
{[1,3]} & Technical & Positive & 64 & 0 (0.0\%) & 5 (7.8\%) & +7.81 pp & 5 (7.8\%) & 0 (0.0\%) \\
{[1,3]} & Contribution & Negative & 64 & 0 (0.0\%) & 7 (10.9\%) & +10.94 pp & 7 (10.9\%) & 0 (0.0\%) \\
{[1,3]} & Contribution & Positive & 64 & 0 (0.0\%) & 7 (10.9\%) & +10.94 pp & 7 (10.9\%) & 0 (0.0\%) \\
{[1,3]} & Lexical & Negative & 64 & 0 (0.0\%) & 3 (4.7\%) & +4.69 pp & 3 (4.7\%) & 0 (0.0\%) \\
{[1,3]} & Lexical & Positive & 64 & 0 (0.0\%) & 4 (6.2\%) & +6.25 pp & 4 (6.2\%) & 0 (0.0\%) \\
{[4,5]} & Novelty & Negative & 247 & 0 (0.0\%) & 41 (16.6\%) & +16.60 pp & 41 (16.6\%) & 0 (0.0\%) \\
{[4,5]} & Novelty & Positive & 247 & 0 (0.0\%) & 83 (33.6\%) & +33.60 pp & 83 (33.6\%) & 0 (0.0\%) \\
{[4,5]} & Evidence & Negative & 246 & 0 (0.0\%) & 42 (17.1\%) & +17.07 pp & 42 (17.1\%) & 0 (0.0\%) \\
{[4,5]} & Evidence & Positive & 246 & 0 (0.0\%) & 88 (35.8\%) & +35.77 pp & 88 (35.8\%) & 0 (0.0\%) \\
{[4,5]} & Scope & Negative & 245 & 0 (0.0\%) & 53 (21.6\%) & +21.63 pp & 53 (21.6\%) & 0 (0.0\%) \\
{[4,5]} & Scope & Positive & 246 & 0 (0.0\%) & 73 (29.7\%) & +29.67 pp & 73 (29.7\%) & 0 (0.0\%) \\
{[4,5]} & Technical & Negative & 247 & 0 (0.0\%) & 67 (27.1\%) & +27.13 pp & 67 (27.1\%) & 0 (0.0\%) \\
{[4,5]} & Technical & Positive & 246 & 0 (0.0\%) & 75 (30.5\%) & +30.49 pp & 75 (30.5\%) & 0 (0.0\%) \\
{[4,5]} & Contribution & Negative & 245 & 0 (0.0\%) & 77 (31.4\%) & +31.43 pp & 77 (31.4\%) & 0 (0.0\%) \\
{[4,5]} & Contribution & Positive & 247 & 0 (0.0\%) & 76 (30.8\%) & +30.77 pp & 76 (30.8\%) & 0 (0.0\%) \\
{[4,5]} & Lexical & Negative & 246 & 0 (0.0\%) & 60 (24.4\%) & +24.39 pp & 60 (24.4\%) & 0 (0.0\%) \\
{[4,5]} & Lexical & Positive & 246 & 0 (0.0\%) & 58 (23.6\%) & +23.58 pp & 58 (23.6\%) & 0 (0.0\%) \\
\bottomrule
\end{tabular}
\end{adjustbox}
\caption{Pooled transition counts underlying weak-accept probability across original-score ranges for both rewrite models and all five reviewer models.}\phantomsection\label{tab:score_band_boundary_counts_combined}
\end{table*}

\begin{table*}[p]
\ContinuedFloat
\centering
\begin{adjustbox}{max width=\textwidth,center}
\begin{tabular}{lllrrrrrr}
\toprule
Range & Dimension & Direction & $n$ & Original $\geq6$ & Rewrite $\geq6$ & Change & Upward & Downward \\
\midrule
\rowcolor{black!6}\multicolumn{9}{l}{\textit{Standard evaluation protocol}}\\
{[6,7]} & Novelty & Negative & 489 & 489 (100.0\%) & 370 (75.7\%) & -24.34 pp & 0 (0.0\%) & 119 (24.3\%) \\
{[6,7]} & Novelty & Positive & 489 & 489 (100.0\%) & 442 (90.4\%) & -9.61 pp & 0 (0.0\%) & 47 (9.6\%) \\
{[6,7]} & Evidence & Negative & 490 & 490 (100.0\%) & 385 (78.6\%) & -21.43 pp & 0 (0.0\%) & 105 (21.4\%) \\
{[6,7]} & Evidence & Positive & 490 & 490 (100.0\%) & 441 (90.0\%) & -10.00 pp & 0 (0.0\%) & 49 (10.0\%) \\
{[6,7]} & Scope & Negative & 490 & 490 (100.0\%) & 379 (77.3\%) & -22.65 pp & 0 (0.0\%) & 111 (22.7\%) \\
{[6,7]} & Scope & Positive & 490 & 490 (100.0\%) & 447 (91.2\%) & -8.78 pp & 0 (0.0\%) & 43 (8.8\%) \\
{[6,7]} & Technical & Negative & 490 & 490 (100.0\%) & 439 (89.6\%) & -10.41 pp & 0 (0.0\%) & 51 (10.4\%) \\
{[6,7]} & Technical & Positive & 489 & 489 (100.0\%) & 440 (90.0\%) & -10.02 pp & 0 (0.0\%) & 49 (10.0\%) \\
{[6,7]} & Contribution & Negative & 488 & 488 (100.0\%) & 429 (87.9\%) & -12.09 pp & 0 (0.0\%) & 59 (12.1\%) \\
{[6,7]} & Contribution & Positive & 489 & 489 (100.0\%) & 433 (88.5\%) & -11.45 pp & 0 (0.0\%) & 56 (11.5\%) \\
{[6,7]} & Lexical & Negative & 490 & 490 (100.0\%) & 431 (88.0\%) & -12.04 pp & 0 (0.0\%) & 59 (12.0\%) \\
{[6,7]} & Lexical & Positive & 490 & 490 (100.0\%) & 432 (88.2\%) & -11.84 pp & 0 (0.0\%) & 58 (11.8\%) \\
{[8,10]} & Novelty & Negative & 398 & 398 (100.0\%) & 383 (96.2\%) & -3.77 pp & 0 (0.0\%) & 15 (3.8\%) \\
{[8,10]} & Novelty & Positive & 398 & 398 (100.0\%) & 397 (99.7\%) & -0.25 pp & 0 (0.0\%) & 1 (0.3\%) \\
{[8,10]} & Evidence & Negative & 398 & 398 (100.0\%) & 394 (99.0\%) & -1.01 pp & 0 (0.0\%) & 4 (1.0\%) \\
{[8,10]} & Evidence & Positive & 398 & 398 (100.0\%) & 398 (100.0\%) & 0.00 pp & 0 (0.0\%) & 0 (0.0\%) \\
{[8,10]} & Scope & Negative & 398 & 398 (100.0\%) & 391 (98.2\%) & -1.76 pp & 0 (0.0\%) & 7 (1.8\%) \\
{[8,10]} & Scope & Positive & 398 & 398 (100.0\%) & 396 (99.5\%) & -0.50 pp & 0 (0.0\%) & 2 (0.5\%) \\
{[8,10]} & Technical & Negative & 398 & 398 (100.0\%) & 396 (99.5\%) & -0.50 pp & 0 (0.0\%) & 2 (0.5\%) \\
{[8,10]} & Technical & Positive & 398 & 398 (100.0\%) & 396 (99.5\%) & -0.50 pp & 0 (0.0\%) & 2 (0.5\%) \\
{[8,10]} & Contribution & Negative & 398 & 398 (100.0\%) & 397 (99.7\%) & -0.25 pp & 0 (0.0\%) & 1 (0.3\%) \\
{[8,10]} & Contribution & Positive & 398 & 398 (100.0\%) & 397 (99.7\%) & -0.25 pp & 0 (0.0\%) & 1 (0.3\%) \\
{[8,10]} & Lexical & Negative & 398 & 398 (100.0\%) & 398 (100.0\%) & 0.00 pp & 0 (0.0\%) & 0 (0.0\%) \\
{[8,10]} & Lexical & Positive & 398 & 398 (100.0\%) & 395 (99.2\%) & -0.75 pp & 0 (0.0\%) & 3 (0.8\%) \\
\bottomrule
\end{tabular}
\end{adjustbox}
\caption[]{Pooled transition counts underlying weak-accept probability across original-score ranges for both rewrite models and all five reviewer models.}
\end{table*}

\begin{table*}[p]
\ContinuedFloat
\centering
\begin{adjustbox}{max width=\textwidth,center}
\begin{tabular}{lllrrrrrr}
\toprule
Range & Dimension & Direction & $n$ & Original $\geq6$ & Rewrite $\geq6$ & Change & Upward & Downward \\
\midrule
\rowcolor{black!6}\multicolumn{9}{l}{\textit{Strict evaluation protocol}}\\
{[1,3]} & Novelty & Negative & 352 & 0 (0.0\%) & 7 (2.0\%) & +1.99 pp & 7 (2.0\%) & 0 (0.0\%) \\
{[1,3]} & Novelty & Positive & 352 & 0 (0.0\%) & 30 (8.5\%) & +8.52 pp & 30 (8.5\%) & 0 (0.0\%) \\
{[1,3]} & Evidence & Negative & 352 & 0 (0.0\%) & 7 (2.0\%) & +1.99 pp & 7 (2.0\%) & 0 (0.0\%) \\
{[1,3]} & Evidence & Positive & 352 & 0 (0.0\%) & 41 (11.6\%) & +11.65 pp & 41 (11.6\%) & 0 (0.0\%) \\
{[1,3]} & Scope & Negative & 351 & 0 (0.0\%) & 8 (2.3\%) & +2.28 pp & 8 (2.3\%) & 0 (0.0\%) \\
{[1,3]} & Scope & Positive & 352 & 0 (0.0\%) & 13 (3.7\%) & +3.69 pp & 13 (3.7\%) & 0 (0.0\%) \\
{[1,3]} & Technical & Negative & 352 & 0 (0.0\%) & 17 (4.8\%) & +4.83 pp & 17 (4.8\%) & 0 (0.0\%) \\
{[1,3]} & Technical & Positive & 352 & 0 (0.0\%) & 24 (6.8\%) & +6.82 pp & 24 (6.8\%) & 0 (0.0\%) \\
{[1,3]} & Contribution & Negative & 351 & 0 (0.0\%) & 15 (4.3\%) & +4.27 pp & 15 (4.3\%) & 0 (0.0\%) \\
{[1,3]} & Contribution & Positive & 352 & 0 (0.0\%) & 26 (7.4\%) & +7.39 pp & 26 (7.4\%) & 0 (0.0\%) \\
{[1,3]} & Lexical & Negative & 352 & 0 (0.0\%) & 18 (5.1\%) & +5.11 pp & 18 (5.1\%) & 0 (0.0\%) \\
{[1,3]} & Lexical & Positive & 352 & 0 (0.0\%) & 10 (2.8\%) & +2.84 pp & 10 (2.8\%) & 0 (0.0\%) \\
{[4,5]} & Novelty & Negative & 467 & 0 (0.0\%) & 42 (9.0\%) & +8.99 pp & 42 (9.0\%) & 0 (0.0\%) \\
{[4,5]} & Novelty & Positive & 469 & 0 (0.0\%) & 107 (22.8\%) & +22.81 pp & 107 (22.8\%) & 0 (0.0\%) \\
{[4,5]} & Evidence & Negative & 469 & 0 (0.0\%) & 46 (9.8\%) & +9.81 pp & 46 (9.8\%) & 0 (0.0\%) \\
{[4,5]} & Evidence & Positive & 468 & 0 (0.0\%) & 131 (28.0\%) & +27.99 pp & 131 (28.0\%) & 0 (0.0\%) \\
{[4,5]} & Scope & Negative & 467 & 0 (0.0\%) & 44 (9.4\%) & +9.42 pp & 44 (9.4\%) & 0 (0.0\%) \\
{[4,5]} & Scope & Positive & 469 & 0 (0.0\%) & 84 (17.9\%) & +17.91 pp & 84 (17.9\%) & 0 (0.0\%) \\
{[4,5]} & Technical & Negative & 467 & 0 (0.0\%) & 80 (17.1\%) & +17.13 pp & 80 (17.1\%) & 0 (0.0\%) \\
{[4,5]} & Technical & Positive & 468 & 0 (0.0\%) & 94 (20.1\%) & +20.09 pp & 94 (20.1\%) & 0 (0.0\%) \\
{[4,5]} & Contribution & Negative & 469 & 0 (0.0\%) & 88 (18.8\%) & +18.76 pp & 88 (18.8\%) & 0 (0.0\%) \\
{[4,5]} & Contribution & Positive & 468 & 0 (0.0\%) & 100 (21.4\%) & +21.37 pp & 100 (21.4\%) & 0 (0.0\%) \\
{[4,5]} & Lexical & Negative & 470 & 0 (0.0\%) & 77 (16.4\%) & +16.38 pp & 77 (16.4\%) & 0 (0.0\%) \\
{[4,5]} & Lexical & Positive & 468 & 0 (0.0\%) & 73 (15.6\%) & +15.60 pp & 73 (15.6\%) & 0 (0.0\%) \\
\bottomrule
\end{tabular}
\end{adjustbox}
\caption[]{Pooled transition counts underlying weak-accept probability across original-score ranges for both rewrite models and all five reviewer models.}
\end{table*}

\begin{table*}[p]
\ContinuedFloat
\centering
\begin{adjustbox}{max width=\textwidth,center}
\begin{tabular}{lllrrrrrr}
\toprule
Range & Dimension & Direction & $n$ & Original $\geq6$ & Rewrite $\geq6$ & Change & Upward & Downward \\
\midrule
\rowcolor{black!6}\multicolumn{9}{l}{\textit{Strict evaluation protocol}}\\
{[6,7]} & Novelty & Negative & 260 & 260 (100.0\%) & 139 (53.5\%) & -46.54 pp & 0 (0.0\%) & 121 (46.5\%) \\
{[6,7]} & Novelty & Positive & 262 & 262 (100.0\%) & 209 (79.8\%) & -20.23 pp & 0 (0.0\%) & 53 (20.2\%) \\
{[6,7]} & Evidence & Negative & 262 & 262 (100.0\%) & 142 (54.2\%) & -45.80 pp & 0 (0.0\%) & 120 (45.8\%) \\
{[6,7]} & Evidence & Positive & 262 & 262 (100.0\%) & 214 (81.7\%) & -18.32 pp & 0 (0.0\%) & 48 (18.3\%) \\
{[6,7]} & Scope & Negative & 262 & 262 (100.0\%) & 144 (55.0\%) & -45.04 pp & 0 (0.0\%) & 118 (45.0\%) \\
{[6,7]} & Scope & Positive & 262 & 262 (100.0\%) & 206 (78.6\%) & -21.37 pp & 0 (0.0\%) & 56 (21.4\%) \\
{[6,7]} & Technical & Negative & 262 & 262 (100.0\%) & 193 (73.7\%) & -26.34 pp & 0 (0.0\%) & 69 (26.3\%) \\
{[6,7]} & Technical & Positive & 262 & 262 (100.0\%) & 198 (75.6\%) & -24.43 pp & 0 (0.0\%) & 64 (24.4\%) \\
{[6,7]} & Contribution & Negative & 261 & 261 (100.0\%) & 195 (74.7\%) & -25.29 pp & 0 (0.0\%) & 66 (25.3\%) \\
{[6,7]} & Contribution & Positive & 261 & 261 (100.0\%) & 206 (78.9\%) & -21.07 pp & 0 (0.0\%) & 55 (21.1\%) \\
{[6,7]} & Lexical & Negative & 262 & 262 (100.0\%) & 192 (73.3\%) & -26.72 pp & 0 (0.0\%) & 70 (26.7\%) \\
{[6,7]} & Lexical & Positive & 262 & 262 (100.0\%) & 190 (72.5\%) & -27.48 pp & 0 (0.0\%) & 72 (27.5\%) \\
{[8,10]} & Novelty & Negative & 116 & 116 (100.0\%) & 105 (90.5\%) & -9.48 pp & 0 (0.0\%) & 11 (9.5\%) \\
{[8,10]} & Novelty & Positive & 116 & 116 (100.0\%) & 110 (94.8\%) & -5.17 pp & 0 (0.0\%) & 6 (5.2\%) \\
{[8,10]} & Evidence & Negative & 116 & 116 (100.0\%) & 99 (85.3\%) & -14.66 pp & 0 (0.0\%) & 17 (14.7\%) \\
{[8,10]} & Evidence & Positive & 116 & 116 (100.0\%) & 112 (96.6\%) & -3.45 pp & 0 (0.0\%) & 4 (3.4\%) \\
{[8,10]} & Scope & Negative & 116 & 116 (100.0\%) & 100 (86.2\%) & -13.79 pp & 0 (0.0\%) & 16 (13.8\%) \\
{[8,10]} & Scope & Positive & 116 & 116 (100.0\%) & 113 (97.4\%) & -2.59 pp & 0 (0.0\%) & 3 (2.6\%) \\
{[8,10]} & Technical & Negative & 116 & 116 (100.0\%) & 108 (93.1\%) & -6.90 pp & 0 (0.0\%) & 8 (6.9\%) \\
{[8,10]} & Technical & Positive & 116 & 116 (100.0\%) & 110 (94.8\%) & -5.17 pp & 0 (0.0\%) & 6 (5.2\%) \\
{[8,10]} & Contribution & Negative & 116 & 116 (100.0\%) & 113 (97.4\%) & -2.59 pp & 0 (0.0\%) & 3 (2.6\%) \\
{[8,10]} & Contribution & Positive & 115 & 115 (100.0\%) & 108 (93.9\%) & -6.09 pp & 0 (0.0\%) & 7 (6.1\%) \\
{[8,10]} & Lexical & Negative & 116 & 116 (100.0\%) & 110 (94.8\%) & -5.17 pp & 0 (0.0\%) & 6 (5.2\%) \\
{[8,10]} & Lexical & Positive & 116 & 116 (100.0\%) & 110 (94.8\%) & -5.17 pp & 0 (0.0\%) & 6 (5.2\%) \\
\bottomrule
\end{tabular}
\end{adjustbox}
\caption[]{Pooled transition counts underlying weak-accept probability across original-score ranges for both rewrite models and all five reviewer models.}
\end{table*}

\endgroup

\newpage
\subsubsection{Full Reviewer-Model Decomposition}
\label{app:score_band_reviewer_details}

Tables~\ref{tab:score_band_reviewer_rating_matrix} and
\ref{tab:score_band_reviewer_threshold_matrix} hold both the rewrite model and
reviewer model fixed. Each cell reports original, rewrite (change). This
complete decomposition is retained after the pooled inference and boundary
summaries so that reviewer-specific baselines and responses remain available
without interrupting the primary score-range results.

\begingroup
\tiny
\setlength{\tabcolsep}{0.5pt}
\begin{table*}[p]
\centering
\begin{adjustbox}{max width=\textwidth,center}
\begin{tabular}{lllccccc}
\toprule
Original range & Dimension & Direction & \shortstack{Gemini 3.5\\FL} & \shortstack{Qwen 3.5\\F} & GPT-5 mini & GPT-5.5 & Sonnet 5 \\
\midrule
\rowcolor{black!6}\multicolumn{8}{l}{\textit{Standard protocol; Rewrite model: GPT-5.5}}\\
{[1,3]} & Novelty & Negative & N/A & 2.00$\to$6.50 (+4.50) & 3.00$\to$5.50 (+2.50) & 3.00$\to$3.50 (+0.50) & 3.00$\to$3.12 (+0.12) \\
{[1,3]} & Novelty & Positive & N/A & 2.00$\to$4.00 (+2.00) & 3.00$\to$3.00 (0.00) & 3.00$\to$3.58 (+0.58) & 3.00$\to$3.50 (+0.50) \\
{[1,3]} & Evidence & Negative & N/A & 2.00$\to$6.00 (+4.00) & 3.00$\to$4.50 (+1.50) & 3.00$\to$3.50 (+0.50) & 3.00$\to$3.38 (+0.38) \\
{[1,3]} & Evidence & Positive & N/A & 2.00$\to$6.50 (+4.50) & 3.00$\to$6.00 (+3.00) & 3.00$\to$3.67 (+0.67) & 3.00$\to$3.50 (+0.50) \\
{[1,3]} & Scope & Negative & N/A & 2.00$\to$5.50 (+3.50) & 3.00$\to$5.50 (+2.50) & 3.00$\to$3.42 (+0.42) & 3.00$\to$3.38 (+0.38) \\
{[1,3]} & Scope & Positive & N/A & 2.00$\to$6.50 (+4.50) & 3.00$\to$5.00 (+2.00) & 3.00$\to$3.75 (+0.75) & 3.00$\to$3.38 (+0.38) \\
\bottomrule
\end{tabular}
\end{adjustbox}
\caption{Overall ratings across original-score ranges by fixed rewrite model and reviewer model. Each cell reports original to rewrite (change).}\phantomsection\label{tab:score_band_reviewer_rating_matrix}
\end{table*}

\begin{table*}[p]
\ContinuedFloat
\centering
\begin{adjustbox}{max width=\textwidth,center}
\begin{tabular}{lllccccc}
\toprule
Original range & Dimension & Direction & \shortstack{Gemini 3.5\\FL} & \shortstack{Qwen 3.5\\F} & GPT-5 mini & GPT-5.5 & Sonnet 5 \\
\midrule
\rowcolor{black!6}\multicolumn{8}{l}{\textit{Standard protocol; Rewrite model: Opus 4.8}}\\
{[1,3]} & Novelty & Negative & N/A & 2.00$\to$7.00 (+5.00) & 3.00$\to$4.00 (+1.00) & 3.00$\to$3.42 (+0.42) & 3.00$\to$3.38 (+0.38) \\
{[1,3]} & Novelty & Positive & N/A & 2.00$\to$5.50 (+3.50) & 3.00$\to$5.50 (+2.50) & 3.00$\to$3.50 (+0.50) & 3.00$\to$3.38 (+0.38) \\
{[1,3]} & Evidence & Negative & N/A & 2.00$\to$6.50 (+4.50) & 3.00$\to$5.00 (+2.00) & 3.00$\to$3.67 (+0.67) & 3.00$\to$3.25 (+0.25) \\
{[1,3]} & Evidence & Positive & N/A & 2.00$\to$7.00 (+5.00) & 3.00$\to$5.50 (+2.50) & 3.00$\to$3.67 (+0.67) & 3.00$\to$3.62 (+0.62) \\
{[1,3]} & Scope & Negative & N/A & 2.00$\to$5.50 (+3.50) & 3.00$\to$5.50 (+2.50) & 3.00$\to$3.83 (+0.83) & 3.00$\to$3.62 (+0.62) \\
{[1,3]} & Scope & Positive & N/A & 2.00$\to$7.00 (+5.00) & 3.00$\to$5.50 (+2.50) & 3.00$\to$3.83 (+0.83) & 3.00$\to$3.67 (+0.67) \\
\bottomrule
\end{tabular}
\end{adjustbox}
\caption[]{Overall ratings across original-score ranges by fixed rewrite model and reviewer model. Each cell reports original to rewrite (change).}
\end{table*}

\begin{table*}[p]
\ContinuedFloat
\centering
\begin{adjustbox}{max width=\textwidth,center}
\begin{tabular}{lllccccc}
\toprule
Original range & Dimension & Direction & \shortstack{Gemini 3.5\\FL} & \shortstack{Qwen 3.5\\F} & GPT-5 mini & GPT-5.5 & Sonnet 5 \\
\midrule
\rowcolor{black!6}\multicolumn{8}{l}{\textit{Strict protocol; Rewrite model: GPT-5.5}}\\
{[1,3]} & Novelty & Negative & 3.00$\to$3.91 (+0.91) & 3.00$\to$3.79 (+0.79) & 3.00$\to$3.69 (+0.69) & 3.00$\to$3.46 (+0.46) & 3.00$\to$3.14 (+0.14) \\
{[1,3]} & Novelty & Positive & 3.00$\to$4.55 (+1.55) & 3.00$\to$4.75 (+1.75) & 3.00$\to$4.17 (+1.17) & 3.00$\to$3.46 (+0.46) & 3.00$\to$3.43 (+0.43) \\
{[1,3]} & Evidence & Negative & 3.00$\to$4.00 (+1.00) & 3.00$\to$3.86 (+0.86) & 3.00$\to$3.37 (+0.37) & 3.00$\to$3.37 (+0.37) & 3.00$\to$3.14 (+0.14) \\
{[1,3]} & Evidence & Positive & 3.00$\to$5.09 (+2.09) & 3.00$\to$4.21 (+1.21) & 3.00$\to$4.11 (+1.11) & 3.00$\to$3.80 (+0.80) & 3.00$\to$3.14 (+0.14) \\
{[1,3]} & Scope & Negative & 3.00$\to$4.18 (+1.18) & 3.00$\to$3.79 (+0.79) & 3.00$\to$3.56 (+0.56) & 3.00$\to$3.54 (+0.54) & 3.00$\to$3.05 (+0.05) \\
{[1,3]} & Scope & Positive & 3.00$\to$4.36 (+1.36) & 3.00$\to$4.39 (+1.39) & 3.00$\to$3.98 (+0.98) & 3.00$\to$3.59 (+0.59) & 3.00$\to$3.33 (+0.33) \\
\bottomrule
\end{tabular}
\end{adjustbox}
\caption[]{Overall ratings across original-score ranges by fixed rewrite model and reviewer model. Each cell reports original to rewrite (change).}
\end{table*}

\begin{table*}[p]
\ContinuedFloat
\centering
\begin{adjustbox}{max width=\textwidth,center}
\begin{tabular}{lllccccc}
\toprule
Original range & Dimension & Direction & \shortstack{Gemini 3.5\\FL} & \shortstack{Qwen 3.5\\F} & GPT-5 mini & GPT-5.5 & Sonnet 5 \\
\midrule
\rowcolor{black!6}\multicolumn{8}{l}{\textit{Strict protocol; Rewrite model: Opus 4.8}}\\
{[1,3]} & Novelty & Negative & 3.00$\to$3.73 (+0.73) & 3.00$\to$3.89 (+0.89) & 3.00$\to$3.26 (+0.26) & 3.00$\to$3.54 (+0.54) & 3.00$\to$3.05 (+0.05) \\
{[1,3]} & Novelty & Positive & 3.00$\to$4.73 (+1.73) & 3.00$\to$4.04 (+1.04) & 3.00$\to$4.00 (+1.00) & 3.00$\to$3.39 (+0.39) & 3.00$\to$3.38 (+0.38) \\
{[1,3]} & Evidence & Negative & 3.00$\to$3.45 (+0.45) & 3.00$\to$3.96 (+0.96) & 3.00$\to$3.43 (+0.43) & 3.00$\to$3.29 (+0.29) & 3.00$\to$3.00 (0.00) \\
{[1,3]} & Evidence & Positive & 3.00$\to$5.36 (+2.36) & 3.00$\to$4.64 (+1.64) & 3.00$\to$4.48 (+1.48) & 3.00$\to$3.95 (+0.95) & 3.00$\to$3.48 (+0.48) \\
{[1,3]} & Scope & Negative & 3.00$\to$3.18 (+0.18) & 3.00$\to$3.86 (+0.86) & 3.00$\to$3.50 (+0.50) & 3.00$\to$3.49 (+0.49) & 3.00$\to$3.10 (+0.10) \\
{[1,3]} & Scope & Positive & 3.00$\to$4.45 (+1.45) & 3.00$\to$4.18 (+1.18) & 3.00$\to$3.78 (+0.78) & 3.00$\to$3.51 (+0.51) & 3.00$\to$3.43 (+0.43) \\
\bottomrule
\end{tabular}
\end{adjustbox}
\caption[]{Overall ratings across original-score ranges by fixed rewrite model and reviewer model. Each cell reports original to rewrite (change).}
\end{table*}

\endgroup

\begingroup
\tiny
\setlength{\tabcolsep}{0.5pt}
\begin{table*}[p]
\centering
\begin{adjustbox}{max width=\textwidth,center}
\begin{tabular}{lllccccc}
\toprule
Original range & Dimension & Direction & \shortstack{Gemini 3.5\\FL} & \shortstack{Qwen 3.5\\F} & GPT-5 mini & GPT-5.5 & Sonnet 5 \\
\midrule
\rowcolor{black!6}\multicolumn{8}{l}{\textit{Standard protocol; Rewrite model: GPT-5.5}}\\
{[1,3]} & Novelty & Negative & N/A & 0$\to$50 (+50.0 pp) & 0$\to$50 (+50.0 pp) & 0$\to$0 (0.0 pp) & 0$\to$0 (0.0 pp) \\
{[1,3]} & Novelty & Positive & N/A & 0$\to$0 (0.0 pp) & 0$\to$0 (0.0 pp) & 0$\to$8 (+8.3 pp) & 0$\to$0 (0.0 pp) \\
{[1,3]} & Evidence & Negative & N/A & 0$\to$100 (+100.0 pp) & 0$\to$50 (+50.0 pp) & 0$\to$0 (0.0 pp) & 0$\to$0 (0.0 pp) \\
{[1,3]} & Evidence & Positive & N/A & 0$\to$50 (+50.0 pp) & 0$\to$100 (+100.0 pp) & 0$\to$0 (0.0 pp) & 0$\to$0 (0.0 pp) \\
{[1,3]} & Scope & Negative & N/A & 0$\to$50 (+50.0 pp) & 0$\to$50 (+50.0 pp) & 0$\to$8 (+8.3 pp) & 0$\to$0 (0.0 pp) \\
{[1,3]} & Scope & Positive & N/A & 0$\to$50 (+50.0 pp) & 0$\to$0 (0.0 pp) & 0$\to$8 (+8.3 pp) & 0$\to$0 (0.0 pp) \\
\bottomrule
\end{tabular}
\end{adjustbox}
\caption{Weak-accept probabilities across original-score ranges by fixed rewrite model and reviewer model. Each cell reports original to rewrite (change in percentage points).}\phantomsection\label{tab:score_band_reviewer_threshold_matrix}
\end{table*}

\begin{table*}[p]
\ContinuedFloat
\centering
\begin{adjustbox}{max width=\textwidth,center}
\begin{tabular}{lllccccc}
\toprule
Original range & Dimension & Direction & \shortstack{Gemini 3.5\\FL} & \shortstack{Qwen 3.5\\F} & GPT-5 mini & GPT-5.5 & Sonnet 5 \\
\midrule
\rowcolor{black!6}\multicolumn{8}{l}{\textit{Standard protocol; Rewrite model: Opus 4.8}}\\
{[1,3]} & Novelty & Negative & N/A & 0$\to$100 (+100.0 pp) & 0$\to$0 (0.0 pp) & 0$\to$8 (+8.3 pp) & 0$\to$0 (0.0 pp) \\
{[1,3]} & Novelty & Positive & N/A & 0$\to$50 (+50.0 pp) & 0$\to$50 (+50.0 pp) & 0$\to$0 (0.0 pp) & 0$\to$0 (0.0 pp) \\
{[1,3]} & Evidence & Negative & N/A & 0$\to$50 (+50.0 pp) & 0$\to$0 (0.0 pp) & 0$\to$0 (0.0 pp) & 0$\to$0 (0.0 pp) \\
{[1,3]} & Evidence & Positive & N/A & 0$\to$100 (+100.0 pp) & 0$\to$50 (+50.0 pp) & 0$\to$0 (0.0 pp) & 0$\to$0 (0.0 pp) \\
{[1,3]} & Scope & Negative & N/A & 0$\to$50 (+50.0 pp) & 0$\to$50 (+50.0 pp) & 0$\to$0 (0.0 pp) & 0$\to$0 (0.0 pp) \\
{[1,3]} & Scope & Positive & N/A & 0$\to$100 (+100.0 pp) & 0$\to$50 (+50.0 pp) & 0$\to$0 (0.0 pp) & 0$\to$0 (0.0 pp) \\
\bottomrule
\end{tabular}
\end{adjustbox}
\caption[]{Weak-accept probabilities across original-score ranges by fixed rewrite model and reviewer model. Each cell reports original to rewrite (change in percentage points).}
\end{table*}

\begin{table*}[p]
\ContinuedFloat
\centering
\begin{adjustbox}{max width=\textwidth,center}
\begin{tabular}{lllccccc}
\toprule
Original range & Dimension & Direction & \shortstack{Gemini 3.5\\FL} & \shortstack{Qwen 3.5\\F} & GPT-5 mini & GPT-5.5 & Sonnet 5 \\
\midrule
\rowcolor{black!6}\multicolumn{8}{l}{\textit{Strict protocol; Rewrite model: GPT-5.5}}\\
{[1,3]} & Novelty & Negative & 0$\to$18 (+18.2 pp) & 0$\to$0 (0.0 pp) & 0$\to$2 (+1.9 pp) & 0$\to$2 (+2.4 pp) & 0$\to$0 (0.0 pp) \\
{[1,3]} & Novelty & Positive & 0$\to$27 (+27.3 pp) & 0$\to$18 (+17.9 pp) & 0$\to$17 (+16.7 pp) & 0$\to$2 (+2.4 pp) & 0$\to$0 (0.0 pp) \\
{[1,3]} & Evidence & Negative & 0$\to$27 (+27.3 pp) & 0$\to$0 (0.0 pp) & 0$\to$0 (0.0 pp) & 0$\to$2 (+2.4 pp) & 0$\to$0 (0.0 pp) \\
{[1,3]} & Evidence & Positive & 0$\to$45 (+45.5 pp) & 0$\to$7 (+7.1 pp) & 0$\to$11 (+11.1 pp) & 0$\to$7 (+7.3 pp) & 0$\to$0 (0.0 pp) \\
{[1,3]} & Scope & Negative & 0$\to$9 (+9.1 pp) & 0$\to$0 (0.0 pp) & 0$\to$4 (+3.7 pp) & 0$\to$5 (+4.9 pp) & 0$\to$0 (0.0 pp) \\
{[1,3]} & Scope & Positive & 0$\to$27 (+27.3 pp) & 0$\to$4 (+3.6 pp) & 0$\to$6 (+5.6 pp) & 0$\to$5 (+4.9 pp) & 0$\to$0 (0.0 pp) \\
\bottomrule
\end{tabular}
\end{adjustbox}
\caption[]{Weak-accept probabilities across original-score ranges by fixed rewrite model and reviewer model. Each cell reports original to rewrite (change in percentage points).}
\end{table*}

\begin{table*}[p]
\ContinuedFloat
\centering
\begin{adjustbox}{max width=\textwidth,center}
\begin{tabular}{lllccccc}
\toprule
Original range & Dimension & Direction & \shortstack{Gemini 3.5\\FL} & \shortstack{Qwen 3.5\\F} & GPT-5 mini & GPT-5.5 & Sonnet 5 \\
\midrule
\rowcolor{black!6}\multicolumn{8}{l}{\textit{Strict protocol; Rewrite model: Opus 4.8}}\\
{[1,3]} & Novelty & Negative & 0$\to$18 (+18.2 pp) & 0$\to$4 (+3.6 pp) & 0$\to$0 (0.0 pp) & 0$\to$0 (0.0 pp) & 0$\to$0 (0.0 pp) \\
{[1,3]} & Novelty & Positive & 0$\to$27 (+27.3 pp) & 0$\to$4 (+3.6 pp) & 0$\to$11 (+11.1 pp) & 0$\to$5 (+4.9 pp) & 0$\to$0 (0.0 pp) \\
{[1,3]} & Evidence & Negative & 0$\to$9 (+9.1 pp) & 0$\to$4 (+3.6 pp) & 0$\to$2 (+1.9 pp) & 0$\to$0 (0.0 pp) & 0$\to$0 (0.0 pp) \\
{[1,3]} & Evidence & Positive & 0$\to$55 (+54.5 pp) & 0$\to$21 (+21.4 pp) & 0$\to$19 (+18.5 pp) & 0$\to$7 (+7.3 pp) & 0$\to$0 (0.0 pp) \\
{[1,3]} & Scope & Negative & 0$\to$0 (0.0 pp) & 0$\to$0 (0.0 pp) & 0$\to$2 (+1.9 pp) & 0$\to$5 (+4.9 pp) & 0$\to$0 (0.0 pp) \\
{[1,3]} & Scope & Positive & 0$\to$18 (+18.2 pp) & 0$\to$4 (+3.6 pp) & 0$\to$0 (0.0 pp) & 0$\to$2 (+2.4 pp) & 0$\to$0 (0.0 pp) \\
\bottomrule
\end{tabular}
\end{adjustbox}
\caption[]{Weak-accept probabilities across original-score ranges by fixed rewrite model and reviewer model. Each cell reports original to rewrite (change in percentage points).}
\end{table*}

\endgroup

\subsubsection{Detailed Secondary-Score Profiles Across the Rating Scale}
\label{app:score_band_secondary}

Table~\ref{tab:score_band_secondary_combined} combines the corresponding
presentation, contribution, and soundness responses. Each cell reports
original, rewrite (change). These are secondary diagnostics because the range
is defined by original \emph{overall rating}, not by the original secondary
score itself.

\begingroup
\tiny
\setlength{\tabcolsep}{1.5pt}
\begin{table*}[p]
\centering
\begin{adjustbox}{max width=\textwidth,center}
\begin{tabular}{lllccc}
\toprule
Original range & Dimension & Direction & Presentation & Contribution & Soundness \\
\midrule
\rowcolor{black!6}\multicolumn{6}{l}{\textit{Standard evaluation protocol}}\\
{[1,3]} & Novelty & Negative & 2.650, 2.821 (+0.171) & 2.083, 2.200 (+0.117) & 2.017, 2.154 (+0.138) \\
{[1,3]} & Novelty & Positive & 2.650, 2.883 (+0.233) & 2.083, 2.221 (+0.138) & 2.017, 2.183 (+0.167) \\
{[1,3]} & Evidence & Negative & 2.650, 2.900 (+0.250) & 2.083, 2.212 (+0.129) & 2.017, 2.112 (+0.096) \\
{[1,3]} & Evidence & Positive & 2.650, 2.842 (+0.192) & 2.083, 2.250 (+0.167) & 2.017, 2.192 (+0.175) \\
{[1,3]} & Scope & Negative & 2.650, 2.925 (+0.275) & 2.083, 2.163 (+0.079) & 2.017, 2.204 (+0.188) \\
{[1,3]} & Scope & Positive & 2.650, 2.871 (+0.221) & 2.083, 2.229 (+0.146) & 2.017, 2.163 (+0.146) \\
{[1,3]} & Technical & Negative & 2.650, 2.783 (+0.133) & 2.083, 2.237 (+0.154) & 2.017, 2.112 (+0.096) \\
{[1,3]} & Technical & Positive & 2.650, 2.921 (+0.271) & 2.083, 2.362 (+0.279) & 2.017, 2.192 (+0.175) \\
{[1,3]} & Contribution & Negative & 2.650, 2.892 (+0.242) & 2.083, 2.300 (+0.217) & 2.017, 2.221 (+0.204) \\
{[1,3]} & Contribution & Positive & 2.650, 2.833 (+0.183) & 2.083, 2.292 (+0.208) & 2.017, 2.196 (+0.179) \\
{[1,3]} & Lexical & Negative & 2.650, 2.792 (+0.142) & 2.083, 2.237 (+0.154) & 2.017, 2.125 (+0.108) \\
{[1,3]} & Lexical & Positive & 2.650, 2.696 (+0.046) & 2.083, 2.146 (+0.062) & 2.017, 2.150 (+0.133) \\
{[4,5]} & Novelty & Negative & 2.954, 2.901 (-0.053) & 2.681, 2.544 (-0.136) & 2.517, 2.452 (-0.065) \\
{[4,5]} & Novelty & Positive & 2.952, 2.996 (+0.044) & 2.681, 2.762 (+0.081) & 2.519, 2.681 (+0.162) \\
{[4,5]} & Evidence & Negative & 2.954, 2.973 (+0.019) & 2.681, 2.505 (-0.176) & 2.517, 2.502 (-0.015) \\
{[4,5]} & Evidence & Positive & 2.958, 2.954 (-0.005) & 2.681, 2.730 (+0.050) & 2.513, 2.670 (+0.157) \\
{[4,5]} & Scope & Negative & 2.954, 3.030 (+0.076) & 2.681, 2.525 (-0.155) & 2.517, 2.708 (+0.191) \\
{[4,5]} & Scope & Positive & 2.954, 2.983 (+0.029) & 2.681, 2.676 (-0.004) & 2.517, 2.635 (+0.118) \\
{[4,5]} & Technical & Negative & 2.954, 2.952 (-0.002) & 2.681, 2.710 (+0.029) & 2.517, 2.610 (+0.093) \\
{[4,5]} & Technical & Positive & 2.954, 3.017 (+0.063) & 2.681, 2.739 (+0.059) & 2.517, 2.646 (+0.129) \\
{[4,5]} & Contribution & Negative & 2.954, 2.981 (+0.027) & 2.681, 2.688 (+0.007) & 2.517, 2.652 (+0.135) \\
{[4,5]} & Contribution & Positive & 2.952, 3.010 (+0.058) & 2.681, 2.691 (+0.010) & 2.519, 2.661 (+0.142) \\
{[4,5]} & Lexical & Negative & 2.954, 2.964 (+0.010) & 2.681, 2.678 (-0.003) & 2.517, 2.587 (+0.070) \\
{[4,5]} & Lexical & Positive & 2.958, 2.881 (-0.077) & 2.681, 2.708 (+0.028) & 2.513, 2.582 (+0.069) \\
\bottomrule
\end{tabular}
\end{adjustbox}
\caption{Pooled secondary-score responses across original overall-rating ranges for both rewrite models and all five reviewer models. Each cell reports original, rewrite (change).}\phantomsection\label{tab:score_band_secondary_combined}
\end{table*}

\begin{table*}[p]
\ContinuedFloat
\centering
\begin{adjustbox}{max width=\textwidth,center}
\begin{tabular}{lllccc}
\toprule
Original range & Dimension & Direction & Presentation & Contribution & Soundness \\
\midrule
\rowcolor{black!6}\multicolumn{6}{l}{\textit{Standard evaluation protocol}}\\
{[6,7]} & Novelty & Negative & 3.158, 3.135 (-0.023) & 3.012, 2.970 (-0.043) & 3.046, 2.954 (-0.092) \\
{[6,7]} & Novelty & Positive & 3.158, 3.227 (+0.069) & 3.012, 3.153 (+0.141) & 3.046, 3.092 (+0.046) \\
{[6,7]} & Evidence & Negative & 3.158, 3.168 (+0.010) & 3.012, 2.981 (-0.031) & 3.046, 3.024 (-0.022) \\
{[6,7]} & Evidence & Positive & 3.158, 3.220 (+0.062) & 3.012, 3.181 (+0.168) & 3.046, 3.132 (+0.086) \\
{[6,7]} & Scope & Negative & 3.158, 3.204 (+0.046) & 3.012, 2.964 (-0.048) & 3.046, 3.048 (+0.001) \\
{[6,7]} & Scope & Positive & 3.158, 3.217 (+0.059) & 3.012, 3.139 (+0.126) & 3.046, 3.121 (+0.075) \\
{[6,7]} & Technical & Negative & 3.158, 3.170 (+0.012) & 3.012, 3.090 (+0.078) & 3.046, 3.073 (+0.027) \\
{[6,7]} & Technical & Positive & 3.158, 3.236 (+0.079) & 3.012, 3.128 (+0.115) & 3.046, 3.123 (+0.077) \\
{[6,7]} & Contribution & Negative & 3.158, 3.180 (+0.022) & 3.012, 3.070 (+0.057) & 3.046, 3.095 (+0.049) \\
{[6,7]} & Contribution & Positive & 3.158, 3.239 (+0.082) & 3.012, 3.153 (+0.140) & 3.046, 3.125 (+0.079) \\
{[6,7]} & Lexical & Negative & 3.158, 3.190 (+0.032) & 3.012, 3.110 (+0.097) & 3.046, 3.085 (+0.039) \\
{[6,7]} & Lexical & Positive & 3.158, 3.145 (-0.013) & 3.012, 3.080 (+0.068) & 3.046, 3.078 (+0.032) \\
{[8,10]} & Novelty & Negative & 3.867, 3.827 (-0.039) & 3.902, 3.566 (-0.336) & 3.793, 3.561 (-0.232) \\
{[8,10]} & Novelty & Positive & 3.867, 3.888 (+0.021) & 3.902, 3.868 (-0.034) & 3.793, 3.811 (+0.017) \\
{[8,10]} & Evidence & Negative & 3.867, 3.853 (-0.014) & 3.902, 3.580 (-0.322) & 3.793, 3.584 (-0.210) \\
{[8,10]} & Evidence & Positive & 3.867, 3.880 (+0.013) & 3.902, 3.864 (-0.038) & 3.793, 3.843 (+0.050) \\
{[8,10]} & Scope & Negative & 3.867, 3.836 (-0.030) & 3.902, 3.541 (-0.361) & 3.793, 3.599 (-0.194) \\
{[8,10]} & Scope & Positive & 3.867, 3.860 (-0.007) & 3.902, 3.836 (-0.066) & 3.793, 3.814 (+0.021) \\
{[8,10]} & Technical & Negative & 3.867, 3.827 (-0.039) & 3.902, 3.811 (-0.091) & 3.793, 3.745 (-0.048) \\
{[8,10]} & Technical & Positive & 3.867, 3.888 (+0.022) & 3.902, 3.850 (-0.052) & 3.793, 3.823 (+0.029) \\
{[8,10]} & Contribution & Negative & 3.867, 3.856 (-0.010) & 3.902, 3.796 (-0.106) & 3.793, 3.784 (-0.009) \\
{[8,10]} & Contribution & Positive & 3.867, 3.885 (+0.019) & 3.902, 3.875 (-0.027) & 3.793, 3.835 (+0.042) \\
{[8,10]} & Lexical & Negative & 3.867, 3.843 (-0.023) & 3.902, 3.797 (-0.105) & 3.793, 3.731 (-0.063) \\
{[8,10]} & Lexical & Positive & 3.867, 3.786 (-0.080) & 3.902, 3.835 (-0.067) & 3.793, 3.800 (+0.007) \\
\bottomrule
\end{tabular}
\end{adjustbox}
\caption[]{Pooled secondary-score responses across original overall-rating ranges for both rewrite models and all five reviewer models. Each cell reports original, rewrite (change).}
\end{table*}

\begin{table*}[p]
\ContinuedFloat
\centering
\begin{adjustbox}{max width=\textwidth,center}
\begin{tabular}{lllccc}
\toprule
Original range & Dimension & Direction & Presentation & Contribution & Soundness \\
\midrule
\rowcolor{black!6}\multicolumn{6}{l}{\textit{Strict evaluation protocol}}\\
{[1,3]} & Novelty & Negative & 3.045, 3.074 (+0.029) & 2.146, 2.252 (+0.106) & 2.267, 2.365 (+0.098) \\
{[1,3]} & Novelty & Positive & 3.045, 3.070 (+0.025) & 2.146, 2.441 (+0.295) & 2.267, 2.530 (+0.263) \\
{[1,3]} & Evidence & Negative & 3.045, 3.090 (+0.045) & 2.146, 2.241 (+0.095) & 2.267, 2.407 (+0.139) \\
{[1,3]} & Evidence & Positive & 3.045, 3.080 (+0.035) & 2.146, 2.503 (+0.357) & 2.267, 2.615 (+0.348) \\
{[1,3]} & Scope & Negative & 3.045, 3.102 (+0.056) & 2.146, 2.267 (+0.121) & 2.267, 2.497 (+0.230) \\
{[1,3]} & Scope & Positive & 3.045, 3.073 (+0.028) & 2.146, 2.370 (+0.224) & 2.267, 2.526 (+0.259) \\
{[1,3]} & Technical & Negative & 3.045, 3.059 (+0.013) & 2.146, 2.340 (+0.195) & 2.267, 2.462 (+0.195) \\
{[1,3]} & Technical & Positive & 3.045, 3.089 (+0.044) & 2.146, 2.391 (+0.245) & 2.267, 2.492 (+0.224) \\
{[1,3]} & Contribution & Negative & 3.045, 3.090 (+0.045) & 2.146, 2.344 (+0.199) & 2.267, 2.508 (+0.241) \\
{[1,3]} & Contribution & Positive & 3.045, 3.096 (+0.051) & 2.146, 2.438 (+0.292) & 2.267, 2.532 (+0.265) \\
{[1,3]} & Lexical & Negative & 3.045, 3.043 (-0.002) & 2.146, 2.375 (+0.229) & 2.267, 2.463 (+0.196) \\
{[1,3]} & Lexical & Positive & 3.045, 3.009 (-0.036) & 2.146, 2.325 (+0.179) & 2.267, 2.399 (+0.132) \\
{[4,5]} & Novelty & Negative & 3.150, 3.149 (-0.001) & 2.712, 2.450 (-0.262) & 2.877, 2.736 (-0.140) \\
{[4,5]} & Novelty & Positive & 3.150, 3.242 (+0.092) & 2.711, 2.753 (+0.041) & 2.876, 2.981 (+0.106) \\
{[4,5]} & Evidence & Negative & 3.150, 3.151 (+0.001) & 2.711, 2.477 (-0.233) & 2.876, 2.772 (-0.103) \\
{[4,5]} & Evidence & Positive & 3.150, 3.242 (+0.092) & 2.712, 2.765 (+0.054) & 2.877, 2.974 (+0.097) \\
{[4,5]} & Scope & Negative & 3.150, 3.214 (+0.064) & 2.711, 2.488 (-0.226) & 2.876, 2.874 (-0.001) \\
{[4,5]} & Scope & Positive & 3.150, 3.212 (+0.062) & 2.711, 2.676 (-0.034) & 2.876, 2.947 (+0.071) \\
{[4,5]} & Technical & Negative & 3.150, 3.179 (+0.029) & 2.711, 2.625 (-0.086) & 2.876, 2.857 (-0.019) \\
{[4,5]} & Technical & Positive & 3.150, 3.221 (+0.072) & 2.711, 2.672 (-0.039) & 2.876, 2.939 (+0.063) \\
{[4,5]} & Contribution & Negative & 3.149, 3.185 (+0.036) & 2.710, 2.625 (-0.085) & 2.875, 2.926 (+0.051) \\
{[4,5]} & Contribution & Positive & 3.150, 3.224 (+0.075) & 2.712, 2.714 (+0.002) & 2.877, 2.939 (+0.062) \\
{[4,5]} & Lexical & Negative & 3.149, 3.182 (+0.032) & 2.710, 2.651 (-0.059) & 2.875, 2.905 (+0.030) \\
{[4,5]} & Lexical & Positive & 3.150, 3.115 (-0.034) & 2.712, 2.692 (-0.020) & 2.877, 2.890 (+0.013) \\
\bottomrule
\end{tabular}
\end{adjustbox}
\caption[]{Pooled secondary-score responses across original overall-rating ranges for both rewrite models and all five reviewer models. Each cell reports original, rewrite (change).}
\end{table*}

\begin{table*}[p]
\ContinuedFloat
\centering
\begin{adjustbox}{max width=\textwidth,center}
\begin{tabular}{lllccc}
\toprule
Original range & Dimension & Direction & Presentation & Contribution & Soundness \\
\midrule
\rowcolor{black!6}\multicolumn{6}{l}{\textit{Strict evaluation protocol}}\\
{[6,7]} & Novelty & Negative & 3.521, 3.489 (-0.032) & 3.007, 2.847 (-0.160) & 3.077, 3.015 (-0.062) \\
{[6,7]} & Novelty & Positive & 3.521, 3.583 (+0.062) & 3.007, 3.179 (+0.172) & 3.077, 3.191 (+0.114) \\
{[6,7]} & Evidence & Negative & 3.521, 3.501 (-0.020) & 3.007, 2.810 (-0.197) & 3.077, 2.974 (-0.102) \\
{[6,7]} & Evidence & Positive & 3.521, 3.562 (+0.041) & 3.007, 3.276 (+0.269) & 3.077, 3.323 (+0.246) \\
{[6,7]} & Scope & Negative & 3.521, 3.534 (+0.013) & 3.007, 2.862 (-0.144) & 3.077, 3.063 (-0.014) \\
{[6,7]} & Scope & Positive & 3.521, 3.515 (-0.007) & 3.007, 3.103 (+0.096) & 3.077, 3.160 (+0.084) \\
{[6,7]} & Technical & Negative & 3.521, 3.512 (-0.010) & 3.007, 3.073 (+0.066) & 3.077, 3.109 (+0.033) \\
{[6,7]} & Technical & Positive & 3.521, 3.570 (+0.049) & 3.007, 3.129 (+0.122) & 3.077, 3.206 (+0.130) \\
{[6,7]} & Contribution & Negative & 3.521, 3.534 (+0.013) & 3.007, 3.031 (+0.025) & 3.077, 3.125 (+0.048) \\
{[6,7]} & Contribution & Positive & 3.521, 3.567 (+0.046) & 3.007, 3.194 (+0.187) & 3.077, 3.236 (+0.159) \\
{[6,7]} & Lexical & Negative & 3.521, 3.466 (-0.055) & 3.007, 3.012 (+0.005) & 3.077, 3.085 (+0.008) \\
{[6,7]} & Lexical & Positive & 3.521, 3.513 (-0.009) & 3.007, 3.015 (+0.008) & 3.077, 3.090 (+0.013) \\
{[8,10]} & Novelty & Negative & 3.990, 3.933 (-0.057) & 3.901, 3.461 (-0.440) & 3.835, 3.429 (-0.406) \\
{[8,10]} & Novelty & Positive & 3.990, 3.948 (-0.042) & 3.901, 3.808 (-0.094) & 3.835, 3.774 (-0.061) \\
{[8,10]} & Evidence & Negative & 3.990, 3.916 (-0.074) & 3.901, 3.413 (-0.488) & 3.835, 3.403 (-0.432) \\
{[8,10]} & Evidence & Positive & 3.990, 3.964 (-0.026) & 3.901, 3.853 (-0.048) & 3.835, 3.816 (-0.019) \\
{[8,10]} & Scope & Negative & 3.990, 3.933 (-0.057) & 3.901, 3.385 (-0.516) & 3.835, 3.442 (-0.393) \\
{[8,10]} & Scope & Positive & 3.990, 3.963 (-0.026) & 3.901, 3.737 (-0.164) & 3.835, 3.713 (-0.122) \\
{[8,10]} & Technical & Negative & 3.990, 3.950 (-0.040) & 3.901, 3.734 (-0.168) & 3.835, 3.633 (-0.202) \\
{[8,10]} & Technical & Positive & 3.990, 3.929 (-0.060) & 3.901, 3.714 (-0.187) & 3.835, 3.741 (-0.094) \\
{[8,10]} & Contribution & Negative & 3.990, 3.930 (-0.060) & 3.901, 3.763 (-0.139) & 3.835, 3.718 (-0.117) \\
{[8,10]} & Contribution & Positive & 3.990, 3.954 (-0.036) & 3.901, 3.835 (-0.066) & 3.835, 3.825 (-0.010) \\
{[8,10]} & Lexical & Negative & 3.990, 3.951 (-0.039) & 3.901, 3.675 (-0.226) & 3.835, 3.679 (-0.156) \\
{[8,10]} & Lexical & Positive & 3.990, 3.958 (-0.031) & 3.901, 3.765 (-0.136) & 3.835, 3.764 (-0.071) \\
\bottomrule
\end{tabular}
\end{adjustbox}
\caption[]{Pooled secondary-score responses across original overall-rating ranges for both rewrite models and all five reviewer models. Each cell reports original, rewrite (change).}
\end{table*}

\endgroup

\newpage
\section{Detailed Results for Multi-Stage Rhetorical Rewriting}
\label{app:advanced_adaptation}

This appendix expands Section~\ref{sec:targeted_adaptation}. The joint rewrite
applies one combined 6-dimension prompt to the complete manuscript, and the
recursive experiment reapplies that same prompt across successive rounds. The
reviewer-guided experiment compares the prompt-matched original with the Final
reviewer-guided manuscript. Its no-review reference is an independently
generated two-pass endpoint produced without an intermediate review; the two
pipelines are therefore not treated as matched causal treatments.

\subsection{Joint Rewrite by Reviewer Model}
\label{app:combo_composition_details}

Table~\ref{tab:combo_fixed_composition} provides the nonredundant expansion
of the main-text joint rewrite results with both the rewrite model and reviewer
model fixed. It also retains the mean of the six corresponding
single-dimension positive rewrites as a secondary reference. Because the
joint rewrite and single-dimension manuscripts were generated in different
batches, their difference is a descriptive cross-batch comparison rather than
a formal interaction or synergy estimate.
Table~\ref{tab:joint_rewrite_score_range_details} expands
Table~\ref{tab:joint_rewrite_score_ranges} with original and rewrite means,
matched counts, and all four AI original-OA ranges.

\begingroup
\scriptsize
\setlength{\tabcolsep}{3pt}
\begin{longtable}{llrrrrll}
\caption{Joint rewrite results with rewriter, review prompt, and reviewer
fixed. The matched endpoint effect uses the prompt-matched original; the final
column compares joint rewrite with the paper-matched mean of the six positive
single-dimension rewrites. Intervals are 95\% paper-bootstrap intervals and
failed evaluations are retained as missing.}
\phantomsection\label{tab:combo_fixed_composition}\\
\toprule
Prompt & Reviewer & $n$ & Original & Joint & $\Delta$ [95\% CI] &
$\Delta_{\geq6}$ (pp) & Joint$-$single Positive [95\% CI] \\
\midrule
\endfirsthead
\multicolumn{8}{c}{\tablename\ \thetable\ (continued)}\\
\toprule
Prompt & Reviewer & $n$ & Original & Joint & $\Delta$ [95\% CI] &
$\Delta_{\geq6}$ (pp) & Joint$-$single Positive [95\% CI] \\
\midrule
\endhead
\midrule\multicolumn{8}{r}{Continued on next page}\\\endfoot
\bottomrule\endlastfoot
\rowcolor{black!6}\multicolumn{8}{l}{\textit{GPT-5.5 rewriter}}\\
Standard & Gemini 3.5 FL & 120 & 7.717 & 7.775 & $+0.058$ [-0.069,+0.185] & $+0.83$ & $-0.025$ [-0.132,+0.081] \\
 & Qwen 3.5 F & 120 & 6.875 & 6.950 & $+0.075$ [-0.189,+0.339] & $+2.50$ & $-0.201$ [-0.392,-0.008] \\
 & GPT-5 mini & 120 & 6.292 & 6.408 & $+0.117$ [-0.056,+0.289] & $+3.33$ & $-0.050$ [-0.185,+0.090] \\
 & GPT-5.5 & 120 & 5.383 & 5.375 & $-0.008$ [-0.127,+0.110] & $+0.83$ & $-0.100$ [-0.190,-0.014] \\
 & Sonnet 5 & 118 & 5.203 & 5.068 & $-0.136$ [-0.268,-0.003] & $-5.08$ & $+0.056$ [-0.056,+0.165] \\
\addlinespace
Strict & Gemini 3.5 FL & 120 & 6.458 & 6.450 & $-0.008$ [-0.219,+0.202] & $+4.17$ & $-0.322$ [-0.485,-0.160] \\
 & Qwen 3.5 F & 120 & 4.800 & 4.842 & $+0.042$ [-0.206,+0.289] & $+4.17$ & $+0.013$ [-0.171,+0.204] \\
 & GPT-5 mini & 120 & 4.267 & 4.542 & $+0.275$ [+0.068,+0.482] & $+2.50$ & $-0.018$ [-0.210,+0.168] \\
 & GPT-5.5 & 120 & 4.642 & 4.667 & $+0.025$ [-0.170,+0.220] & $-2.50$ & $-0.142$ [-0.272,-0.014] \\
 & Sonnet 5 & 119 & 4.437 & 4.328 & $-0.109$ [-0.260,+0.042] & $-0.84$ & $+0.046$ [-0.083,+0.174] \\
\addlinespace
\rowcolor{black!6}\multicolumn{8}{l}{\textit{Opus 4.8 rewriter}}\\
Standard & Gemini 3.5 FL & 120 & 7.717 & 7.950 & $+0.233$ [+0.114,+0.352] & $+1.67$ & $+0.081$ [+0.025,+0.136] \\
 & Qwen 3.5 F & 120 & 6.875 & 7.333 & $+0.458$ [+0.190,+0.726] & $+5.83$ & $+0.104$ [-0.074,+0.292] \\
 & GPT-5 mini & 120 & 6.292 & 6.758 & $+0.467$ [+0.284,+0.650] & $+6.67$ & $+0.136$ [-0.003,+0.278] \\
 & GPT-5.5 & 120 & 5.383 & 5.475 & $+0.092$ [-0.072,+0.256] & $+10.83$ & $-0.074$ [-0.178,+0.029] \\
 & Sonnet 5 & 119 & 5.202 & 5.395 & $+0.193$ [+0.063,+0.323] & $+7.56$ & $+0.243$ [+0.133,+0.357] \\
\addlinespace
Strict & Gemini 3.5 FL & 120 & 6.458 & 7.158 & $+0.700$ [+0.452,+0.948] & $+8.33$ & $+0.225$ [+0.037,+0.408] \\
 & Qwen 3.5 F & 119 & 4.790 & 5.126 & $+0.336$ [+0.076,+0.596] & $+13.45$ & $+0.147$ [-0.050,+0.349] \\
 & GPT-5 mini & 120 & 4.267 & 5.100 & $+0.833$ [+0.629,+1.038] & $+20.00$ & $+0.467$ [+0.325,+0.614] \\
 & GPT-5.5 & 120 & 4.642 & 4.900 & $+0.258$ [+0.070,+0.447] & $+7.50$ & $+0.053$ [-0.086,+0.192] \\
 & Sonnet 5 & 118 & 4.432 & 4.619 & $+0.186$ [+0.021,+0.351] & $+6.78$ & $+0.220$ [+0.105,+0.336] \\
\end{longtable}
\endgroup

\begin{table}[!htbp]
  \centering
  \caption{\textbf{Joint rewrite scores across original-score ranges.}
  Each cell reports original/rewrite mean OA, followed by the number of matched
  papers in parentheses. Bins are defined separately from each reviewer and
  prompt's original OA. Sparse [8,10] cells are retained for completeness.}
  \label{tab:joint_rewrite_score_range_details}
  \scriptsize
  \begin{tabular}{llrrrr}
    \toprule
    \textbf{Review prompt} & \textbf{Reviewer}
      & \multicolumn{1}{c}{\textbf{[1,3]}}
      & \multicolumn{1}{c}{\textbf{[4,5]}}
      & \multicolumn{1}{c}{\textbf{[6,7]}}
      & \multicolumn{1}{c}{\textbf{[8,10]}} \\
    \midrule
    \multicolumn{6}{l}{\textbf{GPT-5.5 rewriter}} \\
    \cmidrule(lr){1-6}
    Standard & Gemini 3.5 FL & N/A & 5.00/7.00 (2) & 6.00/6.93 (14) & 8.00/7.90 (104) \\
     & Qwen 3.5 F & 2.00/4.50 (2) & 5.00/5.91 (11) & 6.00/6.96 (45) & 8.00/7.21 (62) \\
     & GPT-5 mini & 3.00/4.00 (2) & 5.00/5.64 (11) & 6.00/6.31 (81) & 8.00/7.23 (26) \\
     & GPT-5.5 & 3.00/3.33 (12) & 5.00/5.08 (48) & 6.00/5.95 (55) & 8.00/6.80 (5) \\
     & Sonnet 5 & 3.00/3.50 (16) & 5.00/4.78 (50) & 6.00/5.82 (50) & 8.00/6.00 (2) \\
    \addlinespace
    Strict & Gemini 3.5 FL & 3.00/4.27 (11) & 5.00/5.50 (10) & 6.00/6.20 (50) & 8.00/7.39 (49) \\
     & Qwen 3.5 F & 3.00/4.21 (28) & 5.00/4.97 (72) & 6.00/5.00 (14) & 8.00/5.83 (6) \\
     & GPT-5 mini & 3.00/3.89 (54) & 5.00/4.87 (46) & 6.00/5.55 (20) & N/A \\
     & GPT-5.5 & 3.00/3.71 (41) & 5.00/4.61 (44) & 6.00/5.73 (33) & 8.00/8.00 (2) \\
     & Sonnet 5 & 3.00/3.29 (42) & 5.00/4.73 (62) & 6.00/5.57 (14) & 8.00/6.00 (1) \\
    \midrule
    \multicolumn{6}{l}{\textbf{Opus 4.8 rewriter}} \\
    \cmidrule(lr){1-6}
    Standard & Gemini 3.5 FL & N/A & 5.00/7.00 (2) & 6.00/7.71 (14) & 8.00/8.00 (104) \\
     & Qwen 3.5 F & 2.00/8.00 (2) & 5.00/6.55 (11) & 6.00/7.11 (45) & 8.00/7.61 (62) \\
     & GPT-5 mini & 3.00/5.50 (2) & 5.00/6.00 (11) & 6.00/6.59 (81) & 8.00/7.69 (26) \\
     & GPT-5.5 & 3.00/3.83 (12) & 5.00/5.25 (48) & 6.00/5.91 (55) & 8.00/6.80 (5) \\
     & Sonnet 5 & 3.00/3.62 (16) & 5.00/5.22 (51) & 6.00/6.04 (50) & 8.00/8.00 (2) \\
    \addlinespace
    Strict & Gemini 3.5 FL & 3.00/5.36 (11) & 5.00/6.60 (10) & 6.00/7.04 (50) & 8.00/7.80 (49) \\
     & Qwen 3.5 F & 3.00/4.61 (28) & 5.00/5.15 (72) & 6.00/5.77 (13) & 8.00/5.83 (6) \\
     & GPT-5 mini & 3.00/4.61 (54) & 5.00/5.33 (46) & 6.00/5.90 (20) & N/A \\
     & GPT-5.5 & 3.00/3.71 (41) & 5.00/5.05 (44) & 6.00/6.00 (33) & 8.00/8.00 (2) \\
     & Sonnet 5 & 3.00/3.67 (42) & 5.00/5.00 (61) & 6.00/5.71 (14) & 8.00/6.00 (1) \\
    \bottomrule
  \end{tabular}
\end{table}

\newpage

\subsection{Final Reviewer-Guided Endpoints by Reviewer Model}
\label{app:review_final_endpoints}

Table~\ref{tab:joint_guided_summary} reports the compact,
reviewer-specific endpoint and no-review two-generation workflow contrasts
under standard and strict in the main text.
Table~\ref{tab:review_fixed_evaluator_summary} gives the corresponding
original and Final levels for all 20 configurations, together with paired
changes, threshold shifts, and 95\% intervals for both the endpoint and the
descriptive no-review workflow contrast.

\begingroup
\scriptsize
\setlength{\tabcolsep}{3pt}
\begin{longtable}{llrrrrll}
\caption{Final reviewer-guided endpoints with rewriter, review prompt, and
reviewer fixed. $\Delta$ compares Final with the prompt-matched original;
$\Delta_{\mathrm{RN}}$ compares Final with the independently generated
no-review two-pass endpoint. Intervals are 95\% paper-bootstrap intervals and
failed evaluations are retained as missing.}
\phantomsection\label{tab:review_fixed_evaluator_summary}\\
\toprule
Prompt & Reviewer & $n$ & Original & Final & $\Delta$ [95\% CI] &
$\Delta_{\geq6}$ (pp) & $\Delta_{\mathrm{RN}}$ [95\% CI] \\
\midrule
\endfirsthead
\multicolumn{8}{c}{\tablename\ \thetable\ (continued)}\\
\toprule
Prompt & Reviewer & $n$ & Original & Final & $\Delta$ [95\% CI] &
$\Delta_{\geq6}$ (pp) & $\Delta_{\mathrm{RN}}$ [95\% CI] \\
\midrule
\endhead
\midrule\multicolumn{8}{r}{Continued on next page}\\\endfoot
\bottomrule\endlastfoot
\rowcolor{black!6}\multicolumn{8}{l}{\textit{GPT-5.5 rewriter}}\\
Standard & Gemini 3.5 FL & 119 & 7.714 & 7.697 & $-0.017$ [-0.153,+0.119] & $+1.68$ & $-0.067$ [-0.214,+0.080] \\
 & Qwen 3.5 F & 119 & 6.882 & 6.941 & $+0.059$ [-0.179,+0.297] & $+4.20$ & $-0.101$ [-0.336,+0.135] \\
 & GPT-5 mini & 119 & 6.294 & 6.513 & $+0.218$ [+0.032,+0.405] & $+5.04$ & $+0.092$ [-0.100,+0.284] \\
 & GPT-5.5 & 119 & 5.387 & 5.319 & $-0.067$ [-0.192,+0.057] & $-6.72$ & $-0.025$ [-0.166,+0.116] \\
 & Sonnet 5 & 118 & 5.203 & 5.085 & $-0.119$ [-0.241,+0.003] & $-5.08$ & $-0.076$ [-0.207,+0.055] \\
\addlinespace
Strict & Gemini 3.5 FL & 119 & 6.471 & 6.370 & $-0.101$ [-0.295,+0.093] & $-1.68$ & $-0.235$ [-0.477,+0.007] \\
 & Qwen 3.5 F & 119 & 4.798 & 4.807 & $+0.008$ [-0.211,+0.228] & $+2.52$ & $+0.017$ [-0.218,+0.252] \\
 & GPT-5 mini & 119 & 4.277 & 4.613 & $+0.336$ [+0.151,+0.521] & $0.00$ & $+0.076$ [-0.131,+0.282] \\
 & GPT-5.5 & 119 & 4.639 & 4.613 & $-0.025$ [-0.207,+0.157] & $-7.56$ & $-0.134$ [-0.308,+0.039] \\
 & Sonnet 5 & 118 & 4.449 & 4.390 & $-0.059$ [-0.222,+0.103] & $-2.54$ & $+0.068$ [-0.091,+0.227] \\
\addlinespace
\rowcolor{black!6}\multicolumn{8}{l}{\textit{Opus 4.8 rewriter}}\\
Standard & Gemini 3.5 FL & 120 & 7.717 & 7.883 & $+0.167$ [+0.069,+0.264] & $+1.67$ & $-0.067$ [-0.148,+0.014] \\
 & Qwen 3.5 F & 120 & 6.875 & 7.400 & $+0.525$ [+0.278,+0.772] & $+7.50$ & $-0.118$ [-0.318,+0.083] \\
 & GPT-5 mini & 120 & 6.292 & 6.983 & $+0.692$ [+0.488,+0.896] & $+9.17$ & $-0.050$ [-0.233,+0.132] \\
 & GPT-5.5 & 120 & 5.383 & 5.758 & $+0.375$ [+0.236,+0.514] & $+20.83$ & $+0.101$ [-0.001,+0.202] \\
 & Sonnet 5 & 117 & 5.222 & 5.444 & $+0.222$ [+0.092,+0.352] & $+13.68$ & $+0.095$ [-0.042,+0.232] \\
\addlinespace
Strict & Gemini 3.5 FL & 120 & 6.458 & 7.058 & $+0.600$ [+0.355,+0.845] & $+8.33$ & $-0.277$ [-0.521,-0.034] \\
 & Qwen 3.5 F & 120 & 4.800 & 5.283 & $+0.483$ [+0.205,+0.762] & $+21.67$ & $-0.067$ [-0.353,+0.218] \\
 & GPT-5 mini & 120 & 4.267 & 5.183 & $+0.917$ [+0.708,+1.126] & $+23.33$ & $-0.134$ [-0.301,+0.033] \\
 & GPT-5.5 & 120 & 4.642 & 5.125 & $+0.483$ [+0.316,+0.651] & $+15.00$ & $+0.059$ [-0.110,+0.228] \\
 & Sonnet 5 & 119 & 4.437 & 4.739 & $+0.303$ [+0.128,+0.477] & $+6.72$ & $+0.085$ [-0.071,+0.240] \\
\end{longtable}
\endgroup

\newpage

\subsection{Complete Recursive Joint-Rewrite Trajectories}
\label{app:combo_trajectory_details}

Tables~\ref{tab:combo_rating_trajectory} and
\ref{tab:combo_threshold_trajectory} report the reviewer-pooled trajectories
under standard and strict. Tables~\ref{tab:combo_fixed_rating_changes} and
\ref{tab:combo_fixed_threshold_trajectory} then decompose
Figure~\ref{fig:combo_trajectory} by the reviewer model. Each entry is a
cumulative change from the matched original for the indicated round, so the
tables can be compared directly with the original-normalized trajectories in
the figure.

\begin{table}[!htbp]
\centering
\caption{Overall-rating trajectories across recursive joint-rewrite rounds after
averaging all five reviewers within paper. The final column reports Round 3
minus the displayed Original mean; missing evaluations are averaged over available
reviewers.}
\label{tab:combo_rating_trajectory}
\small
\begin{tabular}{llrrrrr}
\toprule
Rewrite model & Review prompt & Original & Round 1 & Round 2 & Round 3 & Round 3 change \\
\midrule
GPT-5.5 & Standard & 6.299 & 6.320 & 6.343 & 6.375 & $+0.076$ \\
 & Strict & 4.922 & 4.965 & 4.999 & 5.078 & $+0.157$ \\
\addlinespace
Opus 4.8 & Standard & 6.295 & 6.586 & 6.703 & 6.733 & $+0.438$ \\
 & Strict & 4.921 & 5.388 & 5.541 & 5.552 & $+0.631$ \\
\bottomrule
\end{tabular}
\end{table}

\begin{table}[!htbp]
\centering
\caption{Weak-accept-probability trajectories across recursive joint-rewrite rounds after
averaging all five reviewers within paper. The final column reports Round 3
minus the matched original; missing evaluations are averaged over available
reviewers.}
\label{tab:combo_threshold_trajectory}
\small
\begin{tabular}{llrrrrr}
\toprule
Rewrite model & Review prompt & Original & Round 1 & Round 2 & Round 3 & Round 3 change \\
\midrule
GPT-5.5 & Standard & 74.25\% & 74.79\% & 75.00\% & 75.46\% & $+1.21$ pp \\
 & Strict & 31.50\% & 33.04\% & 31.42\% & 34.58\% & $+3.08$ pp \\
\addlinespace
Opus 4.8 & Standard & 74.08\% & 80.67\% & 81.68\% & 80.84\% & $+6.76$ pp \\
 & Strict & 31.50\% & 42.88\% & 47.39\% & 48.40\% & $+16.90$ pp \\
\bottomrule
\end{tabular}
\end{table}

\begingroup
\scriptsize
\begin{longtable}{llrrr}
\caption{Reviewer-model decomposition of cumulative overall-rating changes from
the matched original across recursive joint-rewrite rounds. Failed
evaluations are retained as missing.}
\phantomsection\label{tab:combo_fixed_rating_changes}\\
\toprule
Prompt & Reviewer model & Round 1 & Round 2 & Round 3 \\
\midrule
\endfirsthead
\toprule
Prompt & Reviewer model & Round 1 & Round 2 & Round 3 \\
\midrule
\endhead
\midrule\endfoot
\bottomrule\endlastfoot
\rowcolor{black!6}\multicolumn{5}{l}{\textit{GPT-5.5 rewriter}}\\
Standard & Gemini 3.5 FL & $+0.058$ & $+0.050$ & $+0.092$ \\
 & Qwen 3.5 F & $+0.075$ & $+0.158$ & $+0.308$ \\
 & GPT-5 mini & $+0.117$ & $+0.125$ & $+0.142$ \\
 & GPT-5.5 & $-0.008$ & $-0.042$ & $-0.058$ \\
 & Sonnet 5 & $-0.136$ & $-0.042$ & $-0.084$ \\
\addlinespace
Strict & Gemini 3.5 FL & $-0.008$ & $+0.142$ & $+0.300$ \\
 & Qwen 3.5 F & $+0.042$ & $-0.008$ & $+0.200$ \\
 & GPT-5 mini & $+0.275$ & $+0.258$ & $+0.358$ \\
 & GPT-5.5 & $+0.025$ & $+0.108$ & $+0.025$ \\
 & Sonnet 5 & $-0.109$ & $-0.110$ & $-0.119$ \\
\addlinespace
\rowcolor{black!6}\multicolumn{5}{l}{\textit{Opus 4.8 rewriter}}\\
Standard & Gemini 3.5 FL & $+0.233$ & $+0.235$ & $+0.269$ \\
 & Qwen 3.5 F & $+0.458$ & $+0.647$ & $+0.622$ \\
 & GPT-5 mini & $+0.467$ & $+0.765$ & $+0.941$ \\
 & GPT-5.5 & $+0.092$ & $+0.277$ & $+0.252$ \\
 & Sonnet 5 & $+0.193$ & $+0.127$ & $+0.151$ \\
\addlinespace
Strict & Gemini 3.5 FL & $+0.700$ & $+0.882$ & $+1.092$ \\
 & Qwen 3.5 F & $+0.336$ & $+0.546$ & $+0.487$ \\
 & GPT-5 mini & $+0.833$ & $+1.034$ & $+0.924$ \\
 & GPT-5.5 & $+0.258$ & $+0.429$ & $+0.471$ \\
 & Sonnet 5 & $+0.186$ & $+0.227$ & $+0.186$ \\
\end{longtable}
\endgroup

\newpage
\begingroup
\scriptsize
\begin{longtable}{llrrr}
\caption{Reviewer-model decomposition of cumulative changes in weak-accept probability from
the matched original across recursive joint-rewrite rounds. Failed
evaluations are retained as missing.}
\phantomsection\label{tab:combo_fixed_threshold_trajectory}\\
\toprule
Prompt & Reviewer model & Round 1 & Round 2 & Round 3 \\
\midrule
\endfirsthead
\toprule
Prompt & Reviewer model & Round 1 & Round 2 & Round 3 \\
\midrule
\endhead
\midrule\endfoot
\bottomrule\endlastfoot
\rowcolor{black!6}\multicolumn{5}{l}{\textit{GPT-5.5 rewriter}}\\
Standard & Gemini 3.5 FL & $+0.83$ pp & $0.00$ pp & $+0.83$ pp \\
 & Qwen 3.5 F & $+2.50$ pp & $+4.17$ pp & $+4.17$ pp \\
 & GPT-5 mini & $+3.33$ pp & $+4.17$ pp & $+4.17$ pp \\
 & GPT-5.5 & $+0.83$ pp & $+2.50$ pp & $-0.83$ pp \\
 & Sonnet 5 & $-5.08$ pp & $-5.83$ pp & $-1.68$ pp \\
\addlinespace
Strict & Gemini 3.5 FL & $+4.17$ pp & $+2.50$ pp & $+10.00$ pp \\
 & Qwen 3.5 F & $+4.17$ pp & $+2.50$ pp & $+10.00$ pp \\
 & GPT-5 mini & $+2.50$ pp & $-2.50$ pp & $+4.17$ pp \\
 & GPT-5.5 & $-2.50$ pp & $+2.50$ pp & $-4.17$ pp \\
 & Sonnet 5 & $-0.84$ pp & $-5.93$ pp & $-5.08$ pp \\
\addlinespace
\rowcolor{black!6}\multicolumn{5}{l}{\textit{Opus 4.8 rewriter}}\\
Standard & Gemini 3.5 FL & $+1.67$ pp & $+1.68$ pp & $+1.68$ pp \\
 & Qwen 3.5 F & $+5.83$ pp & $+7.56$ pp & $+5.04$ pp \\
 & GPT-5 mini & $+6.67$ pp & $+9.24$ pp & $+10.08$ pp \\
 & GPT-5.5 & $+10.83$ pp & $+15.97$ pp & $+13.45$ pp \\
 & Sonnet 5 & $+7.56$ pp & $+4.24$ pp & $+5.04$ pp \\
\addlinespace
Strict & Gemini 3.5 FL & $+8.33$ pp & $+10.92$ pp & $+13.45$ pp \\
 & Qwen 3.5 F & $+13.45$ pp & $+21.01$ pp & $+21.85$ pp \\
 & GPT-5 mini & $+20.00$ pp & $+31.09$ pp & $+28.57$ pp \\
 & GPT-5.5 & $+7.50$ pp & $+10.92$ pp & $+11.76$ pp \\
 & Sonnet 5 & $+6.78$ pp & $+5.88$ pp & $+8.47$ pp \\
\end{longtable}
\endgroup

\newpage
\section{Detailed Reviewer and Secondary-Score Results}
\label{app:score_scale_results}

This appendix expands the reviewer and secondary-score analyses in
Section~\ref{sec:reviewer_conditions}. The complete prompt-level
calibration comparison is retained in
Table~\ref{tab:prompt_score_comparison}; the additional results below focus on
reviewer differences and cross-reviewer consistency.

\subsection{Five-Reviewer Comparison}
\label{app:reviewer_response_details}

Figure~\ref{fig:dimension_response_distributions} reports the full
five-reviewer response distributions for Gemini~3.5 FL, Qwen~3.5 F,
GPT-5 mini, GPT-5.5, and Sonnet~5, displayed in that order.
Table~\ref{tab:reviewer_response_comparison} reports the exact mean changes
for all five reviewers and the range of the ten pairwise cross-reviewer rank
correlations.

\begingroup
\tiny
\setlength{\tabcolsep}{0.8pt}
\begin{longtable}{llcccccc}
\caption{Reviewer models preserve broad paper ordering but differ in OA changes after rewriting. Cells report mean $\Delta$OA (across-paper SD) over the 12 single-dimension rewrites. The final column is the range of the ten pairwise Spearman correlations of Rewritten OA.}\phantomsection\label{tab:reviewer_response_comparison}\\
\toprule
Rewriter & Prompt & \shortstack{Gemini 3.5\\FL} & \shortstack{Qwen 3.5\\F} & GPT-5 mini & GPT-5.5 & Sonnet 5 & Pairwise $\rho$ range \\
\midrule
\endfirsthead
\toprule
Rewriter & Prompt & \shortstack{Gemini 3.5\\FL} & \shortstack{Qwen 3.5\\F} & GPT-5 mini & GPT-5.5 & Sonnet 5 & Pairwise $\rho$ range \\
\midrule
\endhead
\midrule
\endfoot
\bottomrule
\endlastfoot
GPT-5.5 & Standard & -0.010 (0.482) & +0.109 (1.011) & +0.047 (0.763) & +0.052 (0.559) & -0.280 (0.553) & [0.662, 0.808] \\
GPT-5.5 & Strict & +0.045 (0.885) & -0.103 (1.035) & +0.155 (0.867) & +0.109 (0.658) & -0.260 (0.624) & [0.655, 0.777] \\
Opus 4.8 & Standard & -0.017 (0.437) & +0.101 (1.032) & +0.112 (0.739) & +0.082 (0.563) & -0.173 (0.546) & [0.673, 0.774] \\
Opus 4.8 & Strict & +0.086 (0.884) & -0.005 (0.960) & +0.168 (0.841) & +0.135 (0.660) & -0.178 (0.604) & [0.619, 0.730] \\
\end{longtable}
\endgroup

\subsection{Secondary-Score Associations and Cross-Reviewer Consistency}
\label{app:secondary_score_association}

Table~\ref{tab:secondary_score_oa_association_detailed} expands the
paper-first analysis in Section~\ref{subsec:secondary_oa_coupling}. For each
fixed rewriter, reviewer, and prompt, the 12 single-dimension rewrites are
averaged within paper before estimating the mean paired change and its
association with $\Delta\mathrm{OA}$. The intervals resample papers and
therefore preserve the paper as the unit of analysis.

\begingroup
\scriptsize
\begin{longtable}{llllll}
\caption{Paper-first secondary-score changes and their association with
$\Delta$OA. Each row reports the mean paired change and the Spearman
correlation between that secondary-score change and $\Delta$OA, with 95\%
paper-bootstrap intervals in brackets. The 12 single-dimension rewrites are
averaged within paper before estimation.}
\phantomsection\label{tab:secondary_score_oa_association_detailed}\\
\toprule
Rewriter & Review prompt & Reviewer & Outcome
  & Mean $\Delta$ [95\% CI] & $\rho_{\Delta\mathrm{OA}}$ [95\% CI] \\
\midrule
\endfirsthead
\multicolumn{6}{c}{\tablename\ \thetable\ (continued)}\\
\toprule
Rewriter & Review prompt & Reviewer & Outcome
  & Mean $\Delta$ [95\% CI] & $\rho_{\Delta\mathrm{OA}}$ [95\% CI] \\
\midrule
\endhead
\midrule
\multicolumn{6}{r}{Continued on next page}\\
\endfoot
\bottomrule
\endlastfoot
GPT-5.5 & Standard & Gemini 3.5 FL & Soundness & +.027 [-.022,+.079] & .416 [.172,.642] \\
 &  &  & Presentation & +.019 [-.001,+.047] & .135 [-.162,.389] \\
 &  &  & Contribution & -.010 [-.056,+.037] & .651 [.451,.829] \\
\addlinespace
 &  & Qwen 3.5 F & Soundness & +.042 [-.034,+.124] & .500 [.339,.638] \\
 &  &  & Presentation & +.006 [-.063,+.078] & .275 [.101,.434] \\
 &  &  & Contribution & -.020 [-.095,+.057] & .747 [.604,.864] \\
\addlinespace
 &  & GPT-5 mini & Soundness & +.048 [+.009,+.091] & .261 [.090,.419] \\
 &  &  & Presentation & +.087 [+.052,+.124] & -.045 [-.238,.148] \\
 &  &  & Contribution & +.007 [-.046,+.061] & .500 [.326,.653] \\
\addlinespace
 &  & GPT-5.5 & Soundness & +.067 [+.019,+.120] & .295 [.116,.461] \\
 &  &  & Presentation & +.037 [+.004,+.072] & .196 [-.004,.379] \\
 &  &  & Contribution & +.024 [-.026,+.074] & .298 [.114,.461] \\
\addlinespace
 &  & Sonnet 5 & Soundness & -.041 [-.096,+.013] & .124 [-.065,.308] \\
 &  &  & Presentation & +.040 [-.013,+.097] & .018 [-.192,.227] \\
 &  &  & Contribution & -.101 [-.154,-.047] & -.025 [-.225,.175] \\
\addlinespace
 & Strict & Gemini 3.5 FL & Soundness & +.064 [-.001,+.130] & .631 [.469,.770] \\
 &  &  & Presentation & +.076 [+.026,+.130] & .164 [-.030,.337] \\
 &  &  & Contribution & +.063 [-.003,+.131] & .795 [.685,.890] \\
\addlinespace
 &  & Qwen 3.5 F & Soundness & +.008 [-.079,+.097] & .388 [.227,.532] \\
 &  &  & Presentation & +.024 [-.065,+.115] & .167 [-.021,.338] \\
 &  &  & Contribution & -.039 [-.113,+.037] & .715 [.586,.813] \\
\addlinespace
 &  & GPT-5 mini & Soundness & +.135 [+.067,+.202] & .460 [.279,.626] \\
 &  &  & Presentation & +.007 [-.017,+.026] & -.052 [-.202,.101] \\
 &  &  & Contribution & +.069 [+.004,+.132] & .446 [.257,.610] \\
\addlinespace
 &  & GPT-5.5 & Soundness & +.104 [+.050,+.161] & .373 [.186,.545] \\
 &  &  & Presentation & +.010 [-.001,+.024] & .038 [-.156,.233] \\
 &  &  & Contribution & -.005 [-.062,+.054] & .439 [.270,.586] \\
\addlinespace
 &  & Sonnet 5 & Soundness & -.064 [-.115,-.014] & .349 [.160,.535] \\
 &  &  & Presentation & -.008 [-.047,+.032] & .061 [-.112,.232] \\
 &  &  & Contribution & -.072 [-.134,-.011] & .187 [.021,.351] \\
\midrule
Opus 4.8 & Standard & Gemini 3.5 FL & Soundness & +.010 [-.039,+.062] & .504 [.286,.692] \\
 &  &  & Presentation & +.009 [-.013,+.034] & .101 [-.153,.334] \\
 &  &  & Contribution & +.006 [-.039,+.053] & .633 [.436,.806] \\
\addlinespace
 &  & Qwen 3.5 F & Soundness & +.038 [-.038,+.115] & .458 [.279,.612] \\
 &  &  & Presentation & -.060 [-.130,+.015] & .295 [.110,.462] \\
 &  &  & Contribution & +.003 [-.074,+.083] & .733 [.588,.855] \\
\addlinespace
 &  & GPT-5 mini & Soundness & +.031 [-.008,+.074] & .214 [.045,.379] \\
 &  &  & Presentation & +.097 [+.059,+.138] & -.056 [-.238,.125] \\
 &  &  & Contribution & +.035 [-.015,+.084] & .635 [.478,.764] \\
\addlinespace
 &  & GPT-5.5 & Soundness & +.049 [+.003,+.099] & .349 [.175,.504] \\
 &  &  & Presentation & +.041 [+.010,+.079] & .180 [-.017,.366] \\
 &  &  & Contribution & +.045 [-.006,+.098] & .388 [.214,.535] \\
\addlinespace
 &  & Sonnet 5 & Soundness & -.017 [-.070,+.038] & -.011 [-.205,.186] \\
 &  &  & Presentation & +.025 [-.026,+.080] & .074 [-.116,.265] \\
 &  &  & Contribution & -.060 [-.114,-.008] & -.108 [-.298,.091] \\
\addlinespace
 & Strict & Gemini 3.5 FL & Soundness & +.087 [+.018,+.155] & .587 [.411,.741] \\
 &  &  & Presentation & +.071 [+.018,+.131] & .178 [-.015,.358] \\
 &  &  & Contribution & +.089 [+.022,+.154] & .791 [.667,.888] \\
\addlinespace
 &  & Qwen 3.5 F & Soundness & +.017 [-.065,+.096] & .361 [.191,.520] \\
 &  &  & Presentation & +.019 [-.068,+.108] & .107 [-.065,.272] \\
 &  &  & Contribution & +.007 [-.069,+.080] & .680 [.535,.795] \\
\addlinespace
 &  & GPT-5 mini & Soundness & +.087 [+.018,+.156] & .479 [.307,.638] \\
 &  &  & Presentation & +.013 [-.008,+.031] & .003 [-.167,.171] \\
 &  &  & Contribution & +.067 [+.007,+.128] & .477 [.299,.637] \\
\addlinespace
 &  & GPT-5.5 & Soundness & +.094 [+.038,+.153] & .346 [.159,.515] \\
 &  &  & Presentation & +.007 [-.010,+.022] & .087 [-.093,.265] \\
 &  &  & Contribution & +.024 [-.033,+.083] & .308 [.121,.479] \\
\addlinespace
 &  & Sonnet 5 & Soundness & -.046 [-.096,+.001] & .363 [.165,.551] \\
 &  &  & Presentation & -.017 [-.058,+.024] & .063 [-.117,.234] \\
 &  &  & Contribution & -.038 [-.093,+.019] & .159 [-.010,.321] \\
\end{longtable}
\endgroup

\newpage
Table~\ref{tab:secondary_cross_reviewer_consistency} asks a different
question: whether two reviewers agree on \emph{which papers} exhibit the
largest rewrite responses. Agreement is weak for OA and for all three
secondary scores, including reviewer pairs that produce similarly signed
aggregate changes.

\begingroup
\scriptsize
\setlength{\tabcolsep}{3.5pt}
\begin{longtable}{llcccc}
\caption{Cross-reviewer agreement on paper-level mean rewrite responses.
  Entries are pairwise Spearman correlations between reviewers' paper-level
  changes after averaging the 12 single-dimension rewrites within paper.
  Values near zero indicate that the reviewers do not agree on which papers
  change most, even when their mean changes have the same sign.}
\phantomsection\label{tab:secondary_cross_reviewer_consistency}\\
\toprule
Review prompt & Reviewer pair
  & $\rho(\Delta\mathrm{OA})$
  & $\rho(\Delta\mathrm{Sound.})$
  & $\rho(\Delta\mathrm{Pres.})$
  & $\rho(\Delta\mathrm{Contrib.})$ \\
\midrule
\endfirsthead
\multicolumn{6}{c}{\tablename\ \thetable\ (continued)}\\
\toprule
Review prompt & Reviewer pair
  & $\rho(\Delta\mathrm{OA})$
  & $\rho(\Delta\mathrm{Sound.})$
  & $\rho(\Delta\mathrm{Pres.})$
  & $\rho(\Delta\mathrm{Contrib.})$ \\
\midrule
\endhead
\midrule
\multicolumn{6}{r}{Continued on next page}\\
\endfoot
\bottomrule
\endlastfoot
    \multicolumn{6}{l}{\textbf{GPT-5.5 rewriter}} \\
    \cmidrule(lr){1-6}
    Standard & Gemini 3.5 FL / Qwen 3.5 F & .066 & .041 & -.135 & -.116 \\
     & Gemini 3.5 FL / GPT-5 mini & .056 & .015 & .108 & .098 \\
     & Gemini 3.5 FL / GPT-5.5 & -.103 & -.190 & -.096 & .033 \\
     & Gemini 3.5 FL / Sonnet 5 & .078 & .195 & .144 & .017 \\
     & Qwen 3.5 F / GPT-5 mini & .108 & -.057 & -.102 & -.153 \\
     & Qwen 3.5 F / GPT-5.5 & .026 & .181 & .193 & .076 \\
     & Qwen 3.5 F / Sonnet 5 & .103 & -.020 & .012 & .008 \\
     & GPT-5 mini / GPT-5.5 & .112 & .119 & .236 & .079 \\
     & GPT-5 mini / Sonnet 5 & .033 & .003 & .142 & -.079 \\
     & GPT-5.5 / Sonnet 5 & .160 & -.013 & .053 & .044 \\
    \addlinespace
    Strict & Gemini 3.5 FL / Qwen 3.5 F & .099 & .086 & .108 & .164 \\
     & Gemini 3.5 FL / GPT-5 mini & -.039 & .017 & .033 & .001 \\
     & Gemini 3.5 FL / GPT-5.5 & .148 & .051 & -.046 & -.019 \\
     & Gemini 3.5 FL / Sonnet 5 & -.010 & -.086 & .067 & .188 \\
     & Qwen 3.5 F / GPT-5 mini & .096 & .088 & .096 & .014 \\
     & Qwen 3.5 F / GPT-5.5 & .057 & -.200 & -.027 & -.126 \\
     & Qwen 3.5 F / Sonnet 5 & .058 & -.167 & .006 & .010 \\
     & GPT-5 mini / GPT-5.5 & .120 & -.187 & .209 & -.020 \\
     & GPT-5 mini / Sonnet 5 & .127 & .000 & -.020 & -.059 \\
     & GPT-5.5 / Sonnet 5 & .166 & .214 & .069 & .158 \\
    \midrule
    \multicolumn{6}{l}{\textbf{Opus 4.8 rewriter}} \\
    \cmidrule(lr){1-6}
    Standard & Gemini 3.5 FL / Qwen 3.5 F & .086 & .035 & -.099 & -.074 \\
     & Gemini 3.5 FL / GPT-5 mini & .027 & .085 & .062 & .065 \\
     & Gemini 3.5 FL / GPT-5.5 & -.106 & -.107 & -.064 & .123 \\
     & Gemini 3.5 FL / Sonnet 5 & .166 & .155 & .215 & .002 \\
     & Qwen 3.5 F / GPT-5 mini & .098 & -.113 & -.125 & -.177 \\
     & Qwen 3.5 F / GPT-5.5 & .083 & .284 & .095 & .055 \\
     & Qwen 3.5 F / Sonnet 5 & .114 & -.142 & -.045 & -.009 \\
     & GPT-5 mini / GPT-5.5 & .067 & -.011 & .172 & -.008 \\
     & GPT-5 mini / Sonnet 5 & .128 & -.043 & .135 & -.138 \\
     & GPT-5.5 / Sonnet 5 & .205 & -.095 & -.005 & -.024 \\
    \addlinespace
    Strict & Gemini 3.5 FL / Qwen 3.5 F & .108 & .044 & .060 & .166 \\
     & Gemini 3.5 FL / GPT-5 mini & -.100 & -.027 & .027 & -.069 \\
     & Gemini 3.5 FL / GPT-5.5 & .107 & -.031 & -.039 & -.024 \\
     & Gemini 3.5 FL / Sonnet 5 & -.029 & -.053 & .194 & .181 \\
     & Qwen 3.5 F / GPT-5 mini & .111 & .111 & .059 & -.020 \\
     & Qwen 3.5 F / GPT-5.5 & .153 & -.176 & -.109 & -.124 \\
     & Qwen 3.5 F / Sonnet 5 & .044 & -.159 & -.019 & -.034 \\
     & GPT-5 mini / GPT-5.5 & .032 & -.065 & .241 & .025 \\
     & GPT-5 mini / Sonnet 5 & .177 & -.057 & .081 & -.013 \\
     & GPT-5.5 / Sonnet 5 & .068 & .241 & .136 & .093 \\
\end{longtable}
\endgroup

As an additional record-level analysis, we retain the individual rewrite records but
remove each paper's mean response before calculating the association between
OA and secondary-score changes. Across the 20 combinations of rewriters,
reviewers, and prompts, the resulting correlations range from $.210$ to $.691$ for
soundness, $.020$ to $.239$ for presentation, and $.171$ to $.841$ for
contribution. The within-paper analysis therefore preserves the main
contrast: soundness and contribution changes co-move with OA more strongly
than presentation changes.

\newpage
\section{Stability Under Repeated Review Sampling}
\label{app:reviewer_heterogeneity}

Because repeated evaluation is costly, particularly for larger reviewer models, the primary protocol uses one independent review per evaluation cell. To assess whether this design provides sufficiently stable aggregate estimates, we conduct an auxiliary audit with three independent reviews of the same PDFs. We recompute each estimand using one designated review and using the mean of all three reviews, and then compare the resulting paired contrasts.

\begin{table}[!htbp]
  \centering
  \caption{Agreement between single-review and three-review-mean GPT-5 mini estimands.}
  \label{tab:single_vs_three_summary}
  \small
  \begin{tabular}{lrrrr}
    \toprule
    Estimand family & Pearson & Spearman & Mean $|\Delta|$ & Max $|\Delta|$ \\
    \midrule
    Twelve baseline-paired dimension-direction effects & 0.995 & 0.993 & 0.040 & 0.098 \\
    Six contrasts between Positive and Negative conditions & 0.996 & 0.943 & 0.025 & 0.044 \\
    \bottomrule
  \end{tabular}
\end{table}

As shown in Table~\ref{tab:single_vs_three_summary}, averaging three reviews produces estimates that are very similar to those obtained from a single designated review. Across the 12 baseline-paired effects defined by dimension and direction, the two sets of estimates have Pearson $r=0.995$ and Spearman $\rho=0.993$, with a mean absolute difference of 0.040 rating points and a maximum difference of 0.098 points. Across the six contrasts between the positive and negative conditions, Pearson $r=0.996$ and Spearman $\rho=0.943$, while the mean and maximum absolute differences are only 0.025 and 0.044 points. Thus, both the magnitudes and the directional patterns of the aggregate effects remain nearly unchanged when three reviews are averaged instead of using a single review.

This auxiliary audit suggests that the primary findings are not driven by idiosyncratic variation in one GPT-5 mini review draw. We therefore use one designated review per evaluation cell in the main analyses, allowing broader coverage across reviewer models, evaluation protocols, and rewriting conditions. Complete API resource accounting for the reported experiments is provided in Appendix~\ref{app:protocol}, Table~\ref{tab:api_costs}.

\newpage
\section{Complete Rewrite and Review Prompt Templates}
\label{app:prompts}

This appendix presents the prompts in a reader-oriented format. Each rewrite
request concatenates one dimension-and-direction instruction from
Appendix~\ref{app:rewrite_direction_prompts} with the shared full-paper
requirements in Appendix~\ref{app:rewrite_requirements}. The review prompts
preserve the complete evaluation and structured-output instructions supplied
with each PDF. The advanced-experiment templates make explicit that the same
combined 6-dimension joint rewrite prompt is reused across rounds and that
only the reviewer-guided Final rewrite receives inserted reviewer feedback.

\subsection{Direction-Specific Rewrite Prompts}
\label{app:rewrite_direction_prompts}

Each box contains both interventions for one dimension. The positive and
negative instructions alter only the targeted rhetorical property.

\promptfigure{Claim and Novelty Stance}{prompts/rewrite_claim_novelty_stance.txt}

\promptfigure{Scope and Generalization}{prompts/rewrite_scope_generalization.txt}

\promptfigure{Quantitative Evidence Framing}{prompts/rewrite_evidence_quantitative_framing.txt}

\promptfigure{Contribution Structure and Salience}{prompts/rewrite_contribution_structure.txt}

\promptfigure{Technical Register and Formalism}{prompts/rewrite_technical_register_formalism.txt}

\promptfigure{Lexical and Syntactic Complexity}{prompts/rewrite_lexical_complexity.txt}

\subsection{Shared Full-Paper Rewrite Requirements}
\label{app:rewrite_requirements}

The following requirements are appended to every direction-specific instruction.

\promptfigure{Shared Full-Paper Rewrite Requirements}{prompts/shared_full_paper_rewrite_requirements.txt}

\subsection{Joint and Reviewer-Guided Rewrite Prompts}
\label{app:advanced_rewrite_prompts}

The joint rewrite applies the combined 6-dimension prompt below to the
original source and appends the shared full-paper requirements above. In the
recursive experiment, the same instruction is subsequently applied,
unchanged, to the immediately preceding source; there is no separate
continuation prompt.

\promptfigure{Joint 6-Dimension Positive Rewrite}{prompts/combo_positive.txt}

The reviewer-guided Intermediate manuscript uses the same joint rewrite prompt
and shared requirements. The rewriting backbone model then reviews the compiled Intermediate PDF
under standard. For the Final rewrite, the validated Intermediate Review is
inserted into the wrapper below, and
\texttt{<COORDINATED\_SIX\_DIMENSION\_PROMPT\_AND\_SHARED\_CONTRACT>} is replaced
verbatim by the preceding joint rewrite prompt plus shared requirements.

\promptfigure{Reviewer-Feedback Wrapper for the Final Rewrite}{prompts/review_feedback_wrapper.txt}

\subsection{Review Prompts}

Standard and strict use the same review form and response fields while altering
the decision rule used to translate scientific evidence into an overall
rating. These are the two evaluation protocols used in the reported analyses.

\promptfigure{Standard Review}{prompts/review_standard.txt}

\promptfigure{Strict Review}{prompts/review_strict.txt}

\end{document}